\documentclass[10pt,twocolumn,letterpaper]{article}

\usepackage{cvpr}
\usepackage{times}
\usepackage{epsfig}
\usepackage{graphicx}
\usepackage{amsmath}
\usepackage{amssymb}

\usepackage{color}
\usepackage{soul}
\usepackage{bm}
\usepackage{subfigure}
\usepackage{multirow}
\usepackage{algorithm}
\usepackage[update,prepend]{epstopdf} 
\usepackage[numbers,sort]{natbib} 

\def \spcfigcaption {-5mm}
\def \spcsection {-5mm}
\def \spcpara {0.5mm}

\usepackage[pagebackref=true,breaklinks=true,letterpaper=true,colorlinks,bookmarks=false]{hyperref}

\graphicspath{{./figures/}{./figures/average_train_from_10/}{./figures/pose_distribution/}{./figures/features/}{./figures/clusters/}{./figures/pose_output/}{./figures/baseline_results_camera_ready/}{./figures/average_comparison_graphs/}{./figures/psm_vs_arm/}{./figures/propagation/}{./figures/youtube_graphs/}{./figures/example_poses_youtube/}}

  \cvprfinalcopy 

\def\cvprPaperID{273} 

\ifcvprfinal\fi
\begin{document}

\title{Personalizing Human Video Pose Estimation}

\author{James Charles\\
University of Leeds\\
{\tt\small j.charles@leeds.ac.uk}
\and
Tomas Pfister\\
University of Oxford\\
{\tt\small tp@robots.ox.ac.uk}
\and
Derek Magee\\
University of Leeds\\
{\tt\small d.r.magee@leeds.ac.uk}
\and
David Hogg\\
University of Leeds\\
{\tt\small d.c.hogg@leeds.ac.uk}
\and
Andrew Zisserman\\
University of Oxford\\
{\tt\small az@robots.ox.ac.uk}
}

\def\eg{\emph{e.g}.}
\def\Eg{\emph{E.g}.}
\def\etal{\emph{et al}.}
\def\ie{\emph{i.e}.}
\def\etc{\emph{etc}.}

\hyphenation{ConvNet}
 \hyphenation{ConvNets}

\makeatletter
\def\@maketitle{%
  \newpage
  \null
  \vskip .305in%
  \begin{center}%
    {\Large \bf{\@title} \par}%
    \vspace{24pt}%
    {\large
      \lineskip .3em%
\vspace{-10pt}
      \begin{tabular}[t]{c}%
        \@author
      \end{tabular}\par}%
    \vskip 0.5em%
    \vspace{-4pt}
  \end{center}}
\makeatother

\maketitle

\begin{abstract}
\vspace{-2mm}
We propose a \emph{personalized ConvNet pose estimator} that
automatically adapts itself to the uniqueness of a person's appearance
to improve pose estimation in long videos.

We make the following contributions:
(i)~we show that given a few \emph{high-precision} pose annotations,
e.g.\ from a generic ConvNet pose estimator, additional annotations 
can be generated throughout the video using a combination of
image-based matching for temporally distant frames, and dense optical
flow for temporally local frames;
(ii)~we develop an occlusion aware self-evaluation model that is
able to automatically select the high-quality and
reject the erroneous additional annotations; and
(iii)~we demonstrate that these high-quality annotations can be used
to \emph{fine-tune} a ConvNet pose estimator and thereby personalize
it to lock on to key discriminative features of the
person's appearance.
The outcome is a {\em substantial}
improvement in the pose estimates for the target video using the
personalized ConvNet compared to the original generic ConvNet.

Our method outperforms the state of the art (including
top ConvNet methods) by a large margin on three standard benchmarks, as
well as on a new challenging YouTube video dataset.  Furthermore, we
show that training from the automatically generated annotations can be
used to improve the performance of a generic ConvNet on other
benchmarks.
\end{abstract}

\vspace{\spcsection}
\vspace{-0.5mm}
\section{Introduction}
\vspace{-1mm}
\def \imgscale {0.5}
\def \imgscaletwo {0.3}

\begin{figure*}
\vspace{-4mm}
\hspace{0.7cm}\begin{minipage}{0.64\linewidth}
\begin{center}
\includegraphics[scale=\imgscale]{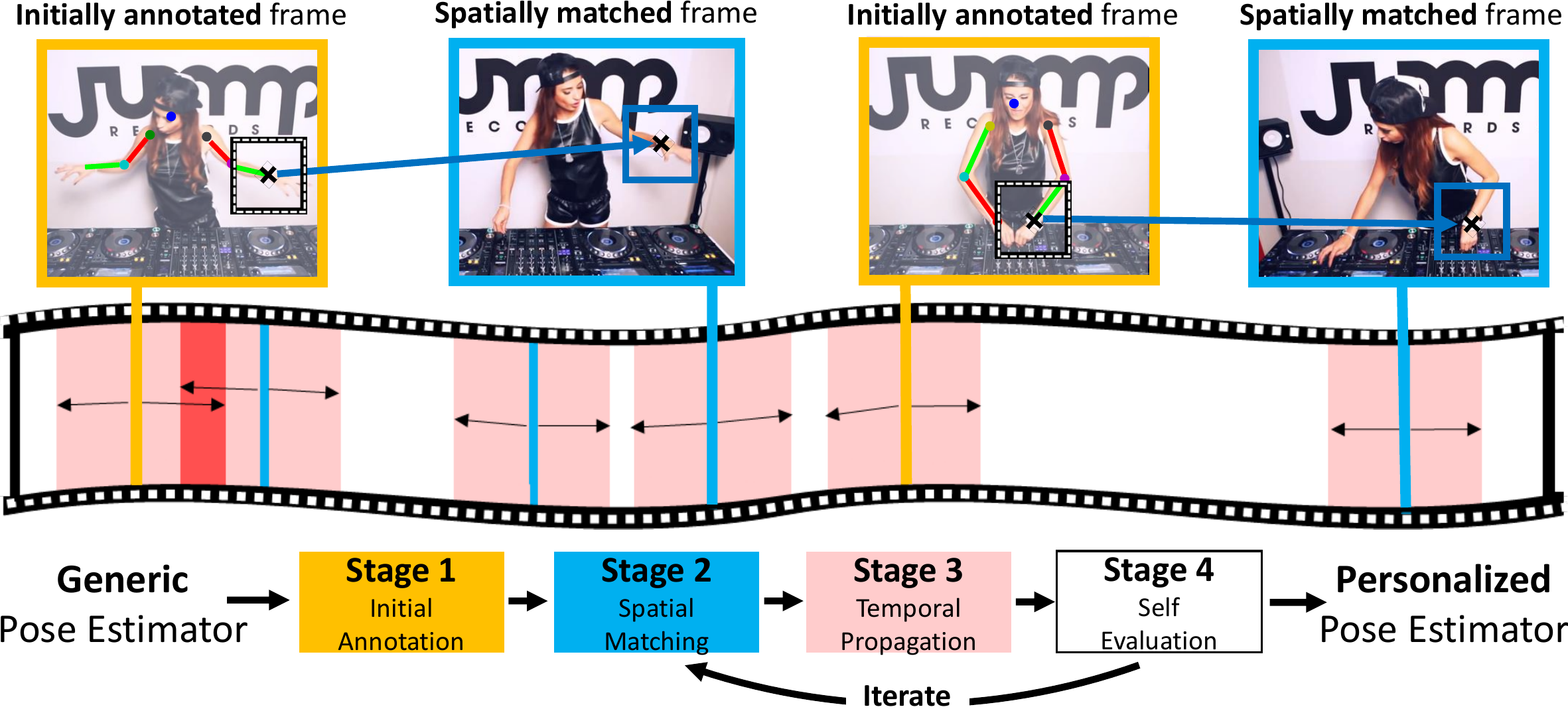}
\end{center}
\end{minipage}
\begin{minipage}{0.35\linewidth}
\begin{center}
\includegraphics[scale=\imgscaletwo]{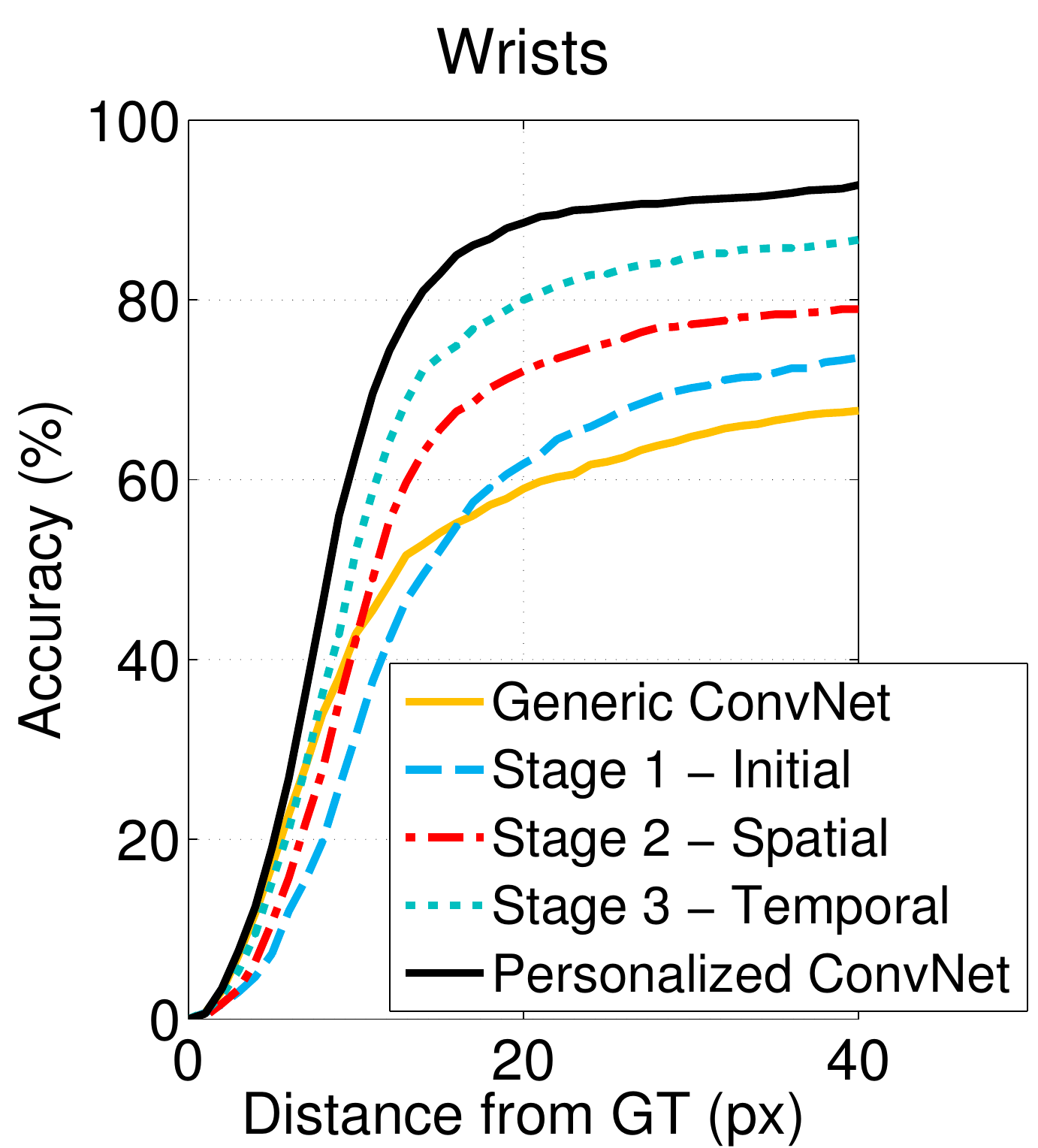}
\vspace{1cm}
\end{center}
\end{minipage}
\vspace{-2mm}
\caption{\small {\bf Personalized video pose estimation.} 
Left: Overview. A few video frames are annotated with confident pose estimates
from one or more generic pose estimators in stage 1. Pose annotations
are spread throughout the video in two more stages: Stage 2 uses spatial
matching (illustrated with the wrist joint), stage 3 propagates
annotation temporally. Stage 4 self-evaluates the new annotations to 
discard errors. These stages are iterated, and resulting annotations
used to train a personalized pose estimator. Right: the improvement in wrist
accuracy after each stage on the YouTube Pose Subset dataset. Starting 
from a generic ConvNet
based pose estimator~\cite{Pfister15}, new annotations are generated over
five iterations, and used to fine-tune the ConvNet. Note,
the large improvement gains obtained by personalizing.}
\label{fig:overview}
\vspace{\spcfigcaption}
\vspace{2mm}
\end{figure*}

Recent advances in 2D human pose estimation exploit complex appearance
models~\cite{johnson10,Bourdev09,Roussos13,Pishchulin13a} and more
recently convolutional neural networks
(ConvNets)~\cite{krizhevsky12,pfister14,ciresan11,tompson12,jain14,toshev14,chen14,Jammalamadaka15,Pfister15}.
However, even the state of the art ConvNets often produce absurdly
erroneous predictions in videos -- particularly for unusual poses,
challenging illumination or viewing conditions, self-occlusions or
unusual shapes (\eg\ when wearing baggy clothing, or unusual body
proportions).  This is due to the lack of large quantities of
annotated data, which is critical for training ConvNets; and, as we
show below, the models failing to exploit person-specific information.

To address these issues, this paper proposes an occlusion-aware method for
automatically learning reliable, \emph{person-specific} pose estimators in long videos.  Using the fact that
people tend not to change appearance over the course of a video
(same clothes, same body shape), we show that the large quantity of
data in the video can be exploited to `personalize' a pose
estimator, thereby improving performance for unusual poses.  The key
idea is to `spread' a small number of high quality automatic pose
annotations throughout the video using spatial image matching
techniques {\em and} temporal propagation (see
Fig~\ref{fig:overview}), and use this new annotation to fine-tune
a generic ConvNet pose estimator. We demonstrate that such personalization
yields significant improvements in detection performance over the original
generic pose estimation method.

Our idea stems from the observation that current pose estimation
methods fail to exploit person-specific information, such as jewelery,
clothing, tattoos \etc\ For example, if one learns a person is
wearing an item of clothing which is easily tracked, such as a
necklace, then this information can be used to help localize the head
and shoulders.  Similarly, for a distinctive pattern or color on an
item of clothing, or a tattoo/watch on a wrist.  The personalization
algorithm essentially `locks on' to these person-specific features,
and exploits them to more accurately determine the pose.

Operationalizing personalization requires a novel set of methods: 
(i)~spatial matching for body parts;
(ii)~temporal propagation based on dense optical flow; and
(iii)~an occlusion-aware pose model
for self-evaluation to verify or excise erroneous annotations.  
We evaluate the personalization algorithm on both long and short videos from YouTube, sign language TV broadcasts and cooking videos, and show that
our method significantly outperforms state of the art generic pose estimators.

More generally, the approach provides a `production system' for
effortlessly and automatically generating copious quantities of high
quality annotated pose data, for example starting from
the
abundant repository of long video sequences containing the same person
on YouTube (\eg\ comedy and cooking shows; DJs; single player sports
such as golf, aerobics, gymnastics; training videos \etc).
These
annotations can be used for large scale training of a generic ConvNet
pose estimator, thereby overcoming current limitations due to limited
and restricted training regimes which rely upon manual annotation.
The code, models and data are available at {\small \url{https://www.robots.ox.ac.uk/~vgg/research/personalization}}.

\vspace{-3mm}
\paragraph{Related work.} 
The fact that human appearance tends to stay unchanged through videos has been used in the past to aid pose estimation~\cite{Ramanan05,Shen14,Amin14}.
Ramanan~\etal~\cite{Ramanan05} train discriminative body part detectors by first detecting `easy' poses (such as a `scissors' walking pose) and then using the appearance learnt from these poses to track the remaining video with a pictorial structure model. 
Shen~\etal~\cite{Shen14} iteratively re-train a pose estimator from confident detections, and also include temporal constraints.
In the same spirit as~\cite{Ramanan05}, we too initialize from high-precision poses and re-train a discriminative model (a ConvNet pose estimator) -- but rather than training a personalized part detector from these poses alone, we first \emph{spread} the initial annotation throughout the whole video~\cite{Buchanan06} using image matching~\cite{Sullivan02} and optical flow~\cite{Zuffi13,Charles14} to generate far more annotated frames. Since long video sequences contain an abundance of data, we can simply delete poor pose annotations (rather than trying to correct them); and even if some of the remaining annotations are incorrect, our ConvNet is able to deal with the label noise. Prior work~\cite{Jammalamadaka12} has used evaluator algorithms to remove entire erroneous pose estimates, whereas here we evaluate individual body part annotations.
Furthermore, in a similar manner to poselets~\cite{Bourdev09,Pishchulin13,Gkioxari13,Sapp13}, our matching framework captures dependencies between both connected and non-connected body parts.

More generally, the idea of starting from general classifiers, and then `personalizing' them has been used in other areas of computer vision, such as pedestrian detection~\cite{Htike14} or object tracking~\cite{Kalal12}. Kalal~\etal~\cite{Kalal12} proposed a tracking-learning-detection paradigm for learning a set of object templates online, \eg\ for vehicles.  
Typically, in this type of approach, object models are initialized from a single frame~\cite{Duffner13, Kalal2010}, and then matched to the next frame before being re-learnt.  
Supancic and Ramanan~\cite{Supancic13} improve this by proposing to revisit tracked frames and re-learn a model to correct previous errors.  
In a similar way, our approach utilizes the whole video to learn body part appearance, but differs to previous work in that we match to \emph{all} frames in a video sequence in one step, and not just frame to frame.

Self-occlusion is a challenging problem for pose estimation, with some methods addressing the issue by incorporating a state in their body model to signal body part occlusion~\cite{Cho13,Kumar09}, or by including an explicit occlusion part template~\cite{Girshick11b}.
Alternatively one can opt to use multiple body models, each one handling a different type of part occlusion \cite{Chen15,Wang08}.
Fine-scale occlusion reasoning (pixel level) is also possible when the depth order is known~\cite{Buehler11} or unknown~\cite{Cho13,Ladicky13}.
Another approach is to train better discriminative part detectors which learn the occlusion patterns from large datasets~\cite{Ghiasi14}.
In our case we use an occlusion-aware pose model trained \emph{only} to signal body part occlusion.  
In this setting, occlusion inference and pose estimation are decoupled, resulting in fast inference while also handling occluded parts correctly.


\section{Personalizing Pose Estimation}
We start with an overview of the personalization algorithm before going into details below.  
Fig~\ref{fig:overview} shows an overview of the process, which has six distinct stages:

\noindent {\bf 1.\ Initial pose annotation.}
Initial pose annotations for the video are obtained with generic pose 
estimators for a few frames (yellow regions in Fig~\ref{fig:overview}).  
By design, these annotations have high precision, but low recall (\ie\ only cover a very small proportion of the frames in the video).

\noindent {\bf 2.\ Spatial matching.}
Image \emph{patches} from the remaining frames are matched (blue regions in Fig~\ref{fig:overview}) to image patches of body joints in frames with annotations (from stage~1).  
This forms a correspondence between annotated frames and matched frames, allowing the body joint annotations to be transferred to quite temporally distant frames in the video.

\noindent {\bf 3.\ Temporal propagation.}
Pose annotations are spread from the annotated frames to \emph{temporally near-by} (neighboring) frames,  
by propagating current annotation temporally along tracks using dense optical flow (pink regions in Fig~\ref{fig:overview}).

\noindent {\bf 4.\ Annotation evaluation.}
In this stage, an evaluation measure discards annotations from the
previous stages that are deemed to be poor.  
Multiple evaluation measures are employed, the two principal ones are: 
(i)~{\em consistency of overlapping annotations} where regions in the video with multiple overlapping annotations (red regions in
Fig~\ref{fig:overview}) coming from different `sources' (\eg\ propagated from different initial annotations) are tested to see whether the annotations agree -- this provides a very natural way to evaluate annotation correctness; and 
(ii)~an {\em occlusion-aware puppet model} that renders a layout according to the predicted pose, and measures consistency with the image, similar to~\cite{Buehler11, Charles11, Zuffi13}.

\noindent {\bf 5.\ Iterating.} 
To maximize frame annotation coverage, stages 2--4 are iterated, with
the evaluator used to discard incorrect propagation histories, and
propagate further those that are verified. Fig~\ref{fig:resultspropagation} demonstrates increased coverage and accuracy as our system iterates. 

\def \imgscale {0.15}
\def \imgscalepose {0.2}
\begin{figure}[t!]
\centering
\begin{tabular}{c*{3}{@{\hspace{2pt}}c}}
\includegraphics[scale=\imgscale, trim = 60mm 10mm 10mm 10mm,clip]{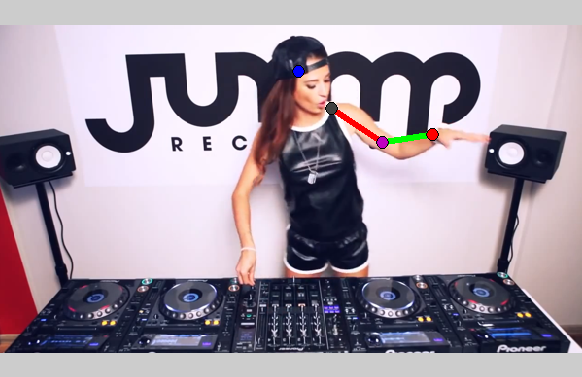}&
\includegraphics[scale=\imgscale, trim = 40mm 10mm 30mm 10mm,clip]{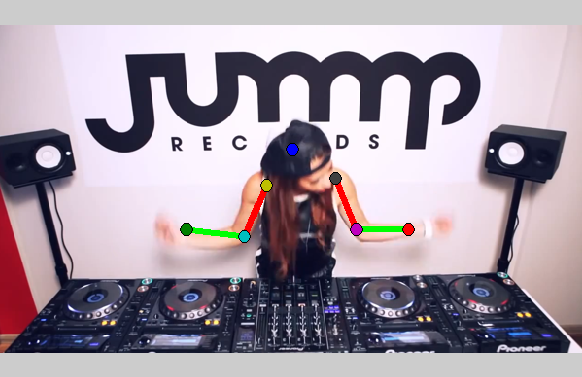}&
\includegraphics[scale=\imgscale, trim = 00mm 10mm 70mm 10mm,clip]{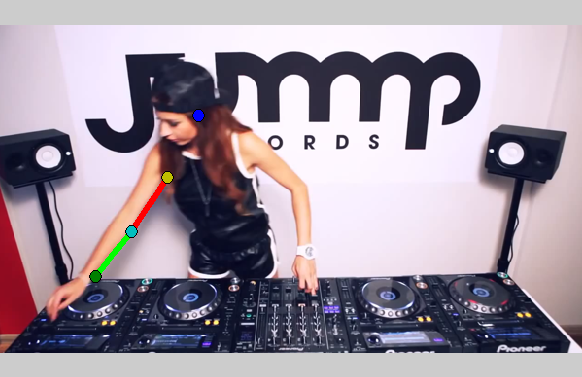}&
\includegraphics[scale=\imgscale, trim = 30mm 10mm 40mm 10mm,clip]{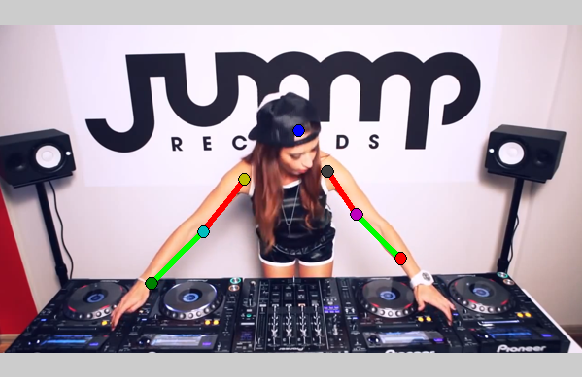}\\
\multicolumn{2}{c}{(a)} & \multicolumn{2}{c}{(b)}\\
\includegraphics[scale=\imgscale, trim = 30mm 10mm 40mm 10mm,clip]{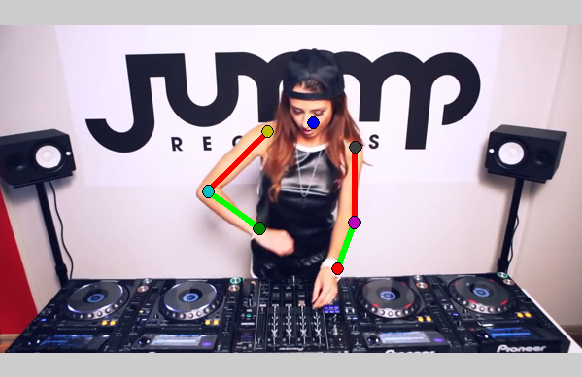}&
\includegraphics[scale=\imgscale, trim = 40mm 10mm 30mm 10mm,clip]{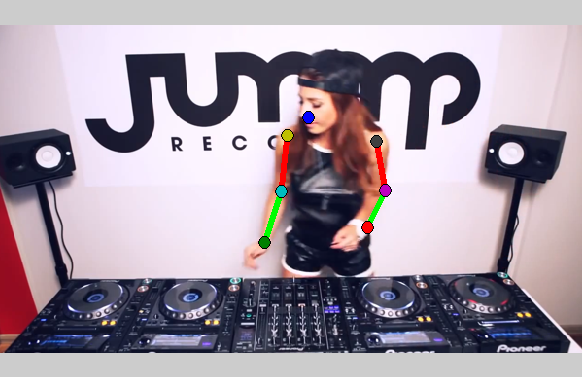}&
\includegraphics[scale=\imgscale, trim = 40mm 10mm 30mm 10mm,clip]{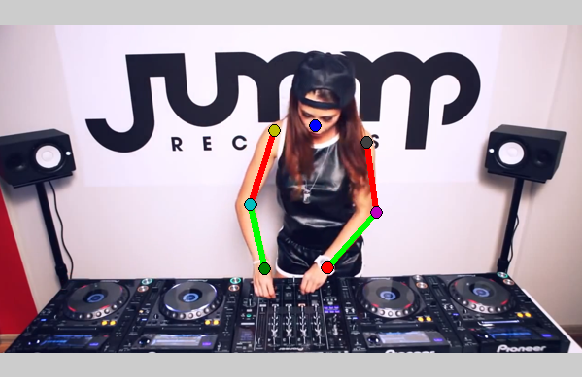}&
\includegraphics[scale=\imgscale, trim = 30mm 10mm 40mm 10mm,clip]{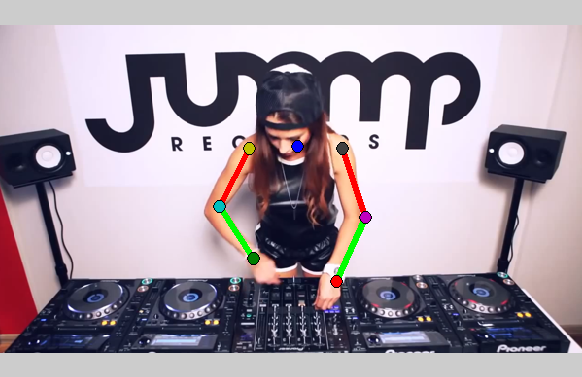}\\
\multicolumn{4}{c}{(c)}
\end{tabular}
\vspace{-0.2cm}
\caption{{\bf Initial pose estimates with generic pose estimators.} 
Two arm pose specific model estimate examples for (a)
a bent arm, and (b) a straight arm  --
note that the models need not fire on both left and
right arms. (c) Poses showing joint detections with high confidence output
from the generic ConvNet  -- in practice not all joints
in a pose will have high confidence and therefore not all will be used
during initialization.}
\label{fig:generaloutput}
\vspace{\spcfigcaption}
\end{figure}

\def \imgscale {0.29}
\begin{figure*}[t!]
\begin{center}
\begin{tabular}{c*{4}{@{\hspace{2pt}}c}}
\hspace{-0.25cm}\includegraphics[scale=\imgscale]{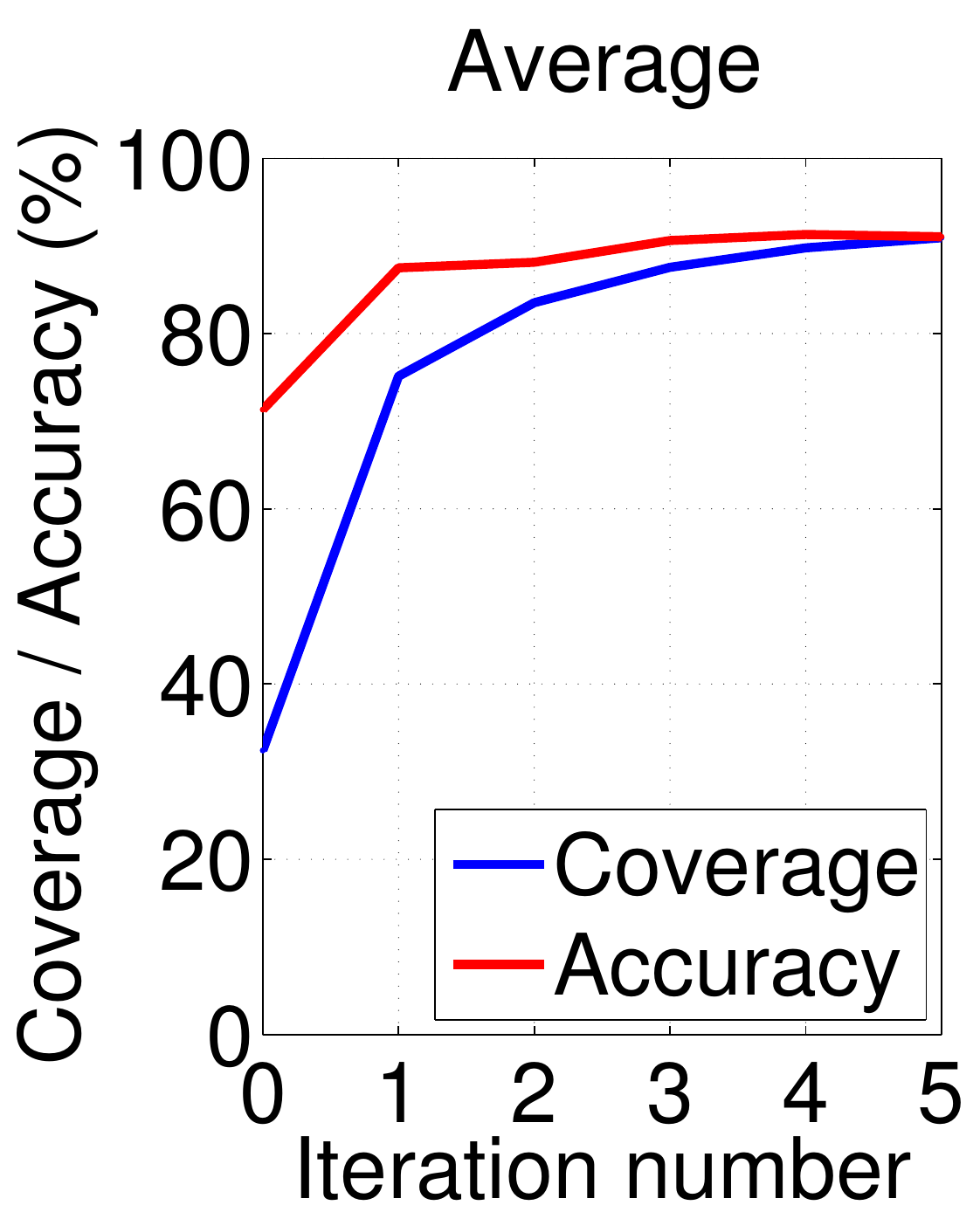}&
\includegraphics[scale=\imgscale]{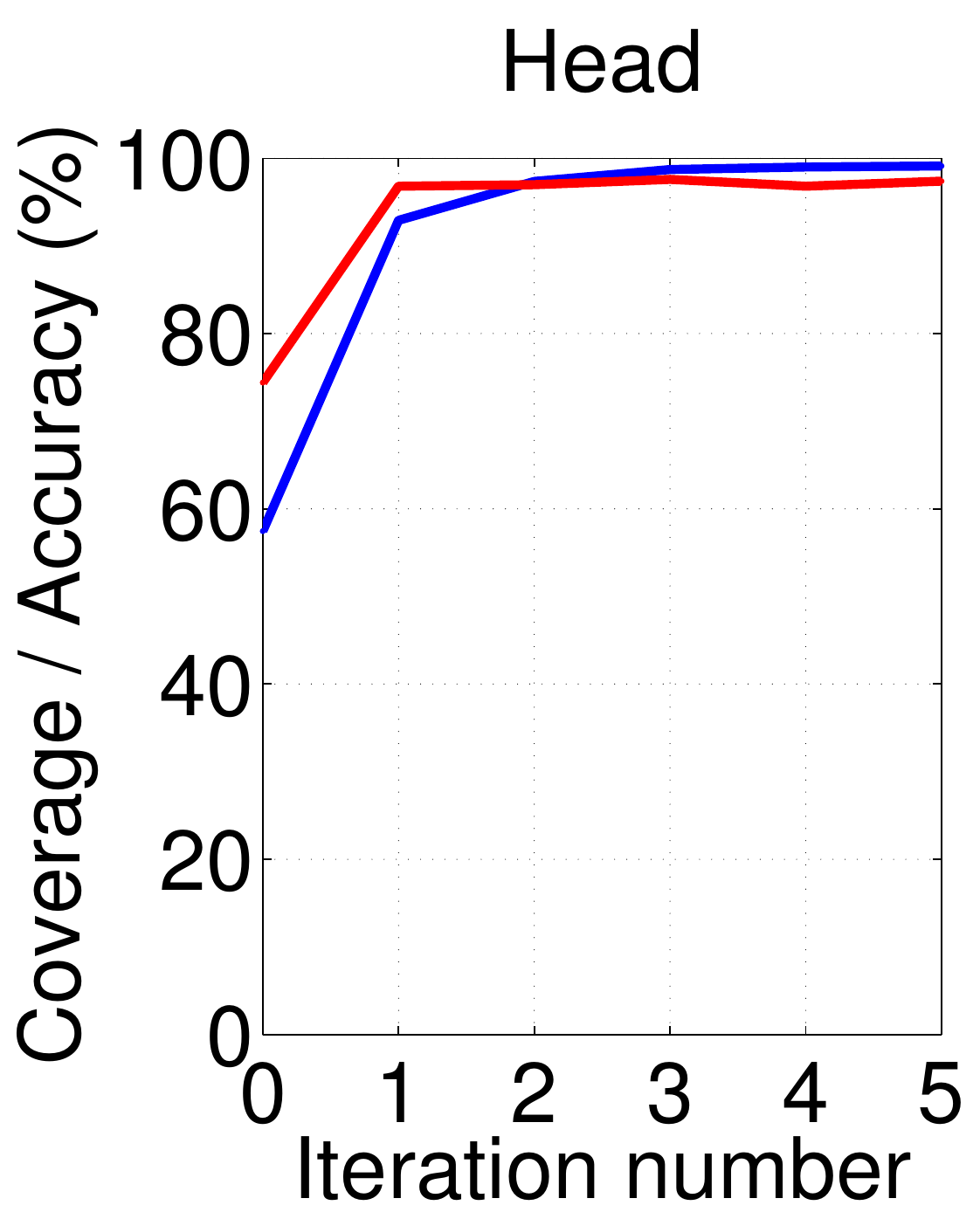} &
\includegraphics[scale=\imgscale]{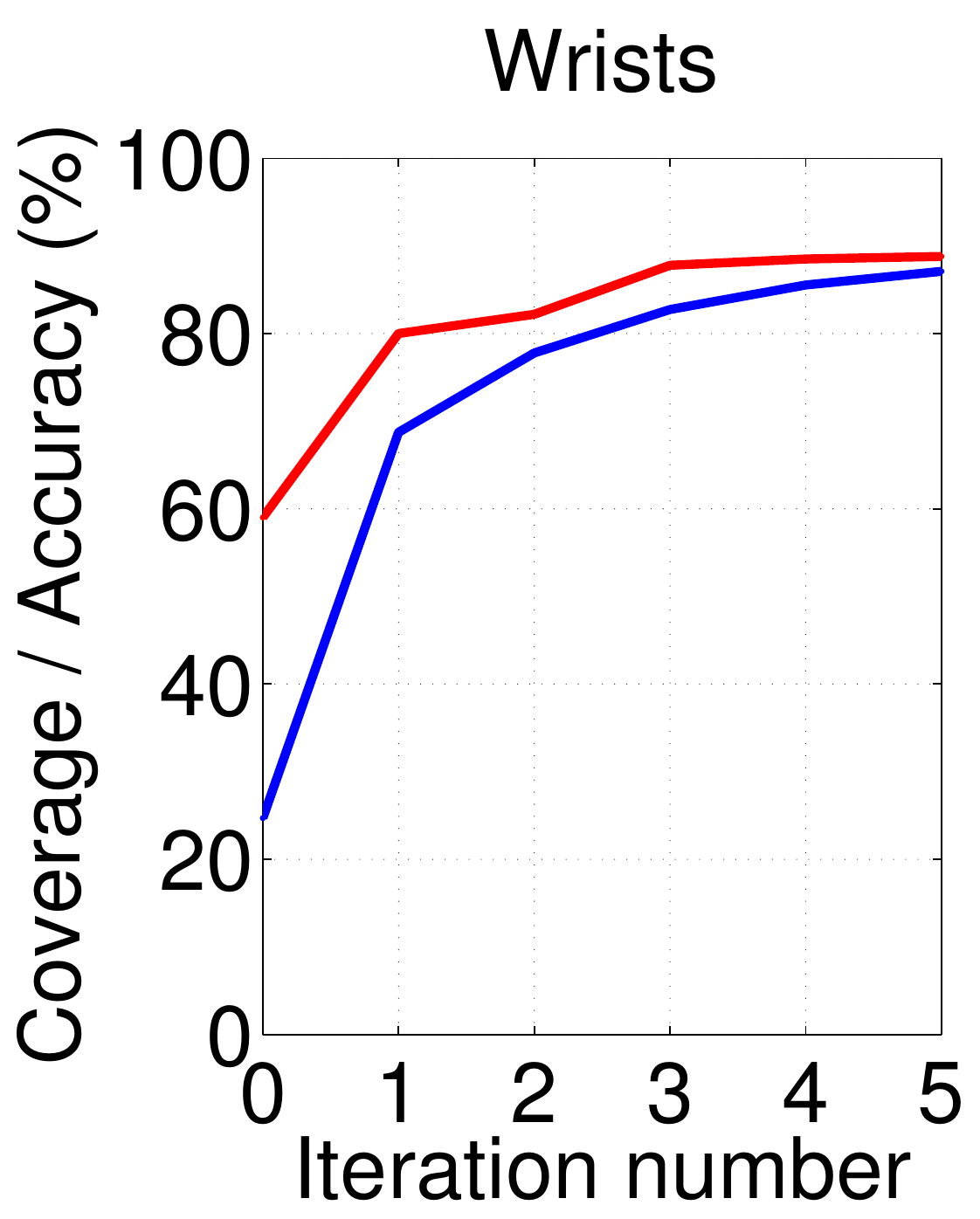}&
\includegraphics[scale=\imgscale]{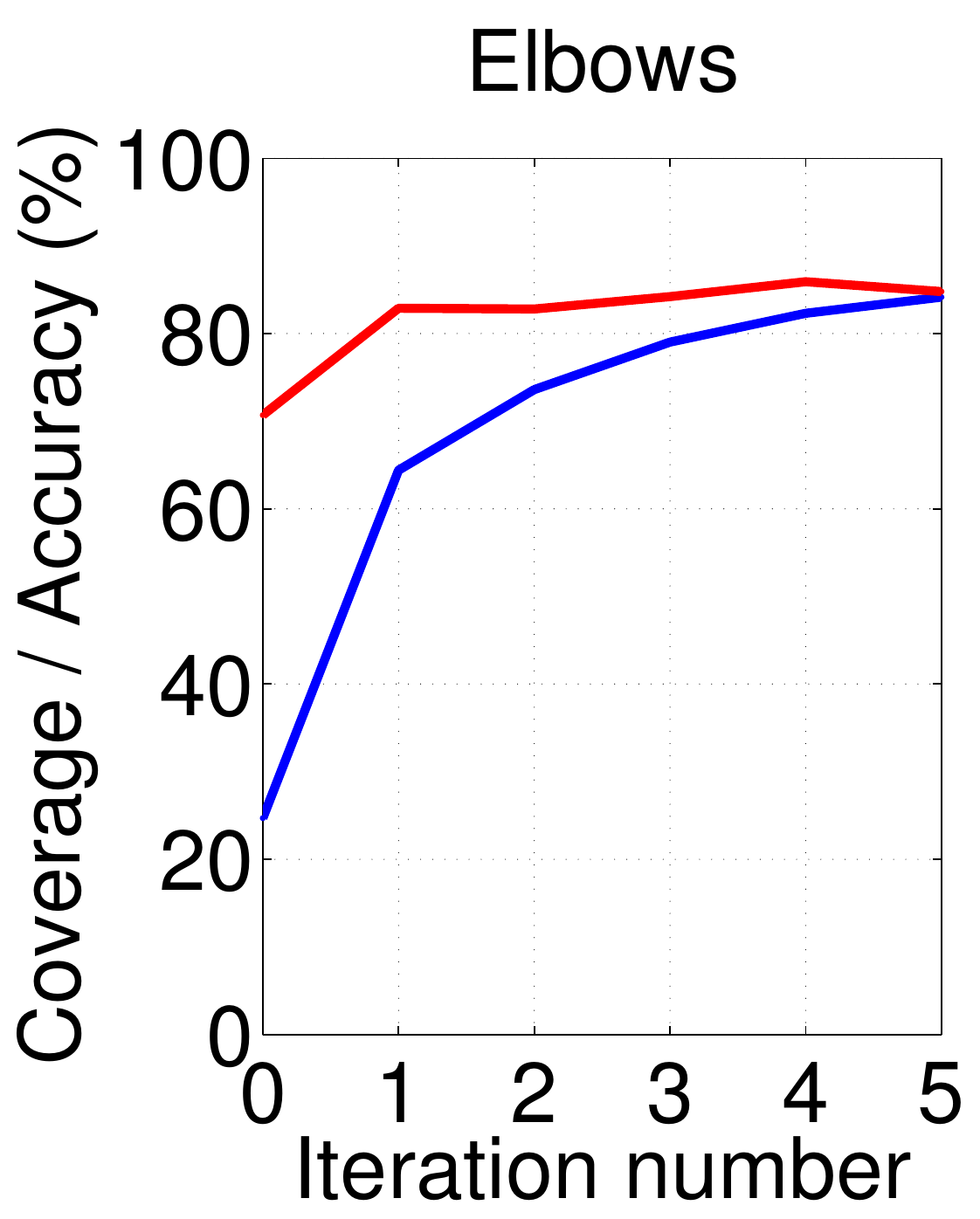}&
\includegraphics[scale=\imgscale]{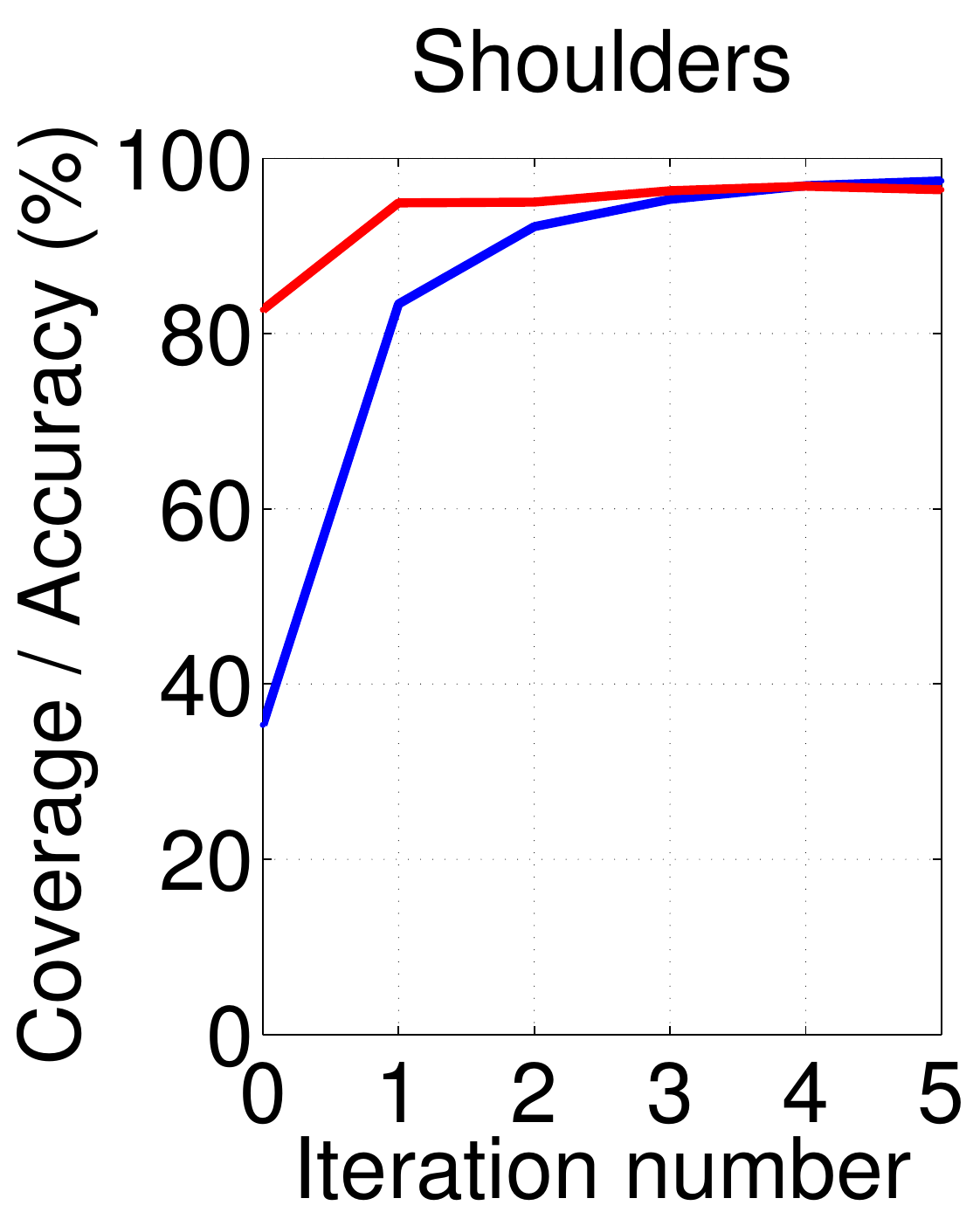}
\end{tabular}
\end{center}
\vspace{-0.4cm}
\caption{{\bf Annotation accuracy and coverage when iterating.}
Accuracy of annotation and coverage (\% of frames with annotation) across the video increases as the system iterates. 
Body joints with less appearance variation, such as the shoulders, have consistent accuracy and coverage rapidly approaches 100\%.
Most notable gains in accuracy are for joints with high appearance variation such as the wrists, improving by 9\% from iteration 1. 
Accuracy is measured as the percentage of estimated annotations within $d = 20$ pixels from ground truth (approx wrist width  15 pixels). 
Results are averaged over videos with ground truth from the Youtube Pose Subset dataset. }
\label{fig:resultspropagation}

\end{figure*}

\noindent {\bf 6.\ Personalizing a ConvNet.}
A generic ConvNet pose estimator is personalized by fine-tuning with the annotations acting as training data for the input video. 
This ConvNet can be applied to frames which the annotation process hasn't reached. 
The caveat here is that the reasoning about self-occlusions in the annotation is lost, since the ConvNet pose estimator's predictions are not occlusion-aware.

\noindent We next describe each of these stages in detail.

\subsection{Generic pose estimator} 
\label{sec:generic_estimator}
In the first stage of the algorithm we obtain high-precision initial
pose annotations for a small number of frames.  These
high-precision pose estimates are obtained
by two approaches: first, by using very high confidence pose
estimates from a ConvNet pose estimator~\cite{Pfister15} (we have
determined empirically that the high confidence joint predictions
(greater than 80\% confidence) are quite accurate). The second approach
is poselet 
like~\cite{Bourdev09,Pishchulin13,Gkioxari13,Sapp13,Jammalamadaka15} 
and involves detecting specific limb poses using 
a Yang and Ramanan~\cite{Yang11} pose detector.
This is done by training the detector to only fire on a small number
of poses~\cite{Ramanan05,Delaitre12}, such as those with no complex
self-occlusions or extreme foreshortening of limbs.

In the case of the Yang and Ramanan~\cite{Yang11} approach,
we modify their release code to detect poses for the left and right arm separately.
This effectively squares the total number of detectable 
upper-body poses, compared to learning separate models for
each pose involving both arms. 15 arm models are trained, enabling us to detect up to 225 different poses with high precision. In general, this model captures more poses with arms above the head than the ConvNet.

Example high-precision detections from the arm pose-specific models,
and the high confidence ConvNet pose estimates are
shown in Fig~\ref{fig:generaloutput}.  We next discuss how we
propagate these pose annotations spatially and temporally.

\def \imgscale {0.33}
\begin{figure*}
\begin{center}
\includegraphics[scale=\imgscale, trim = 0mm 0mm 0mm 70mm, clip]{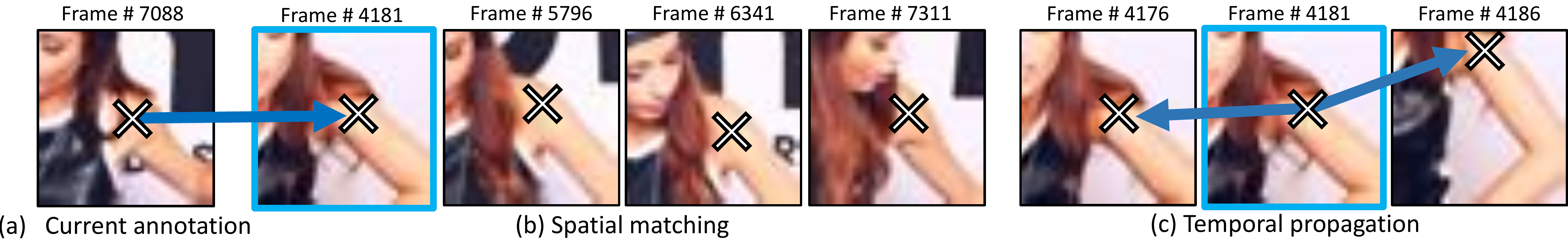}
\end{center}
\vspace{-0.5cm}
\caption{{\bf 
Spreading the annotations spatially and temporally through the video.} 
(a)~Example patch of annotated left shoulder joint (white cross) from the
generic pose estimator.  (b)~Patches matched spatially
to the patch in~(a). Blue arrow illustrates an example propagated annotation, and white crosses show locations in other frames where this annotation has also been propagated.  Note, in this stage annotations can propagate to
temporally very distant frames within the video.  (c)~Temporal
propagation of annotations to neighboring frames.}
\label{fig:matching}
\vspace{0mm}
\vspace{\spcfigcaption}
\end{figure*}

\def \imgscale {0.3}
\begin{figure}[t]
\begin{center}
\includegraphics[scale=\imgscale,trim=0mm 7mm 0mm 0mm,clip]{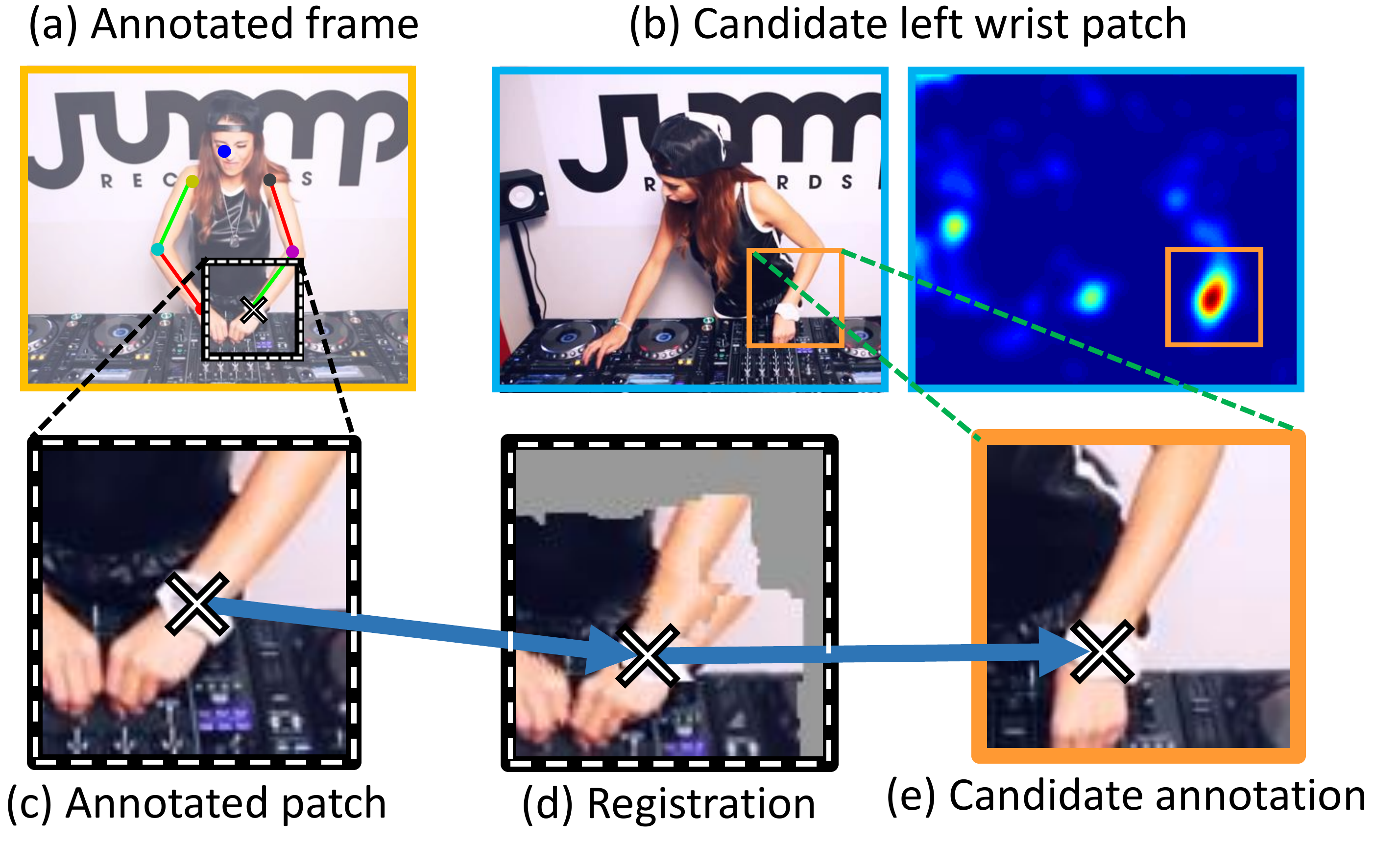}
\end{center}
\vspace{-0.4cm}
\caption{{\bf Spatial matching phase.}
(a)~Annotated frame, with the square delineating the left wrist patch;
(b)~a candidate matching RGB frame (left) and random forest
part-detector confidence map (right), square shows selected candidate
left wrist patch. 
Using SIFTflow, the annotated patch (c) is registered to the candidate
patch (e) as shown in (d). This enables annotation to be
transferred (blue arrows) to the candidate patch.}
\label{fig:siftflow}
\vspace{-5mm}
\end{figure}

\subsection{Spatial matching}
\vspace{-1mm}
In this stage, we propagate the small number of high-precision pose annotations, from the generic pose estimators, to new frames using image patch matching.  
The matching process is illustrated in Fig~\ref{fig:matching}(a-b).  
For each body joint, small image patches in the annotated frames (with the annotated body joint at their center) are matched to new image patches in other frames of the video, and the annotations are then transferred.  
Note, body part patches rather than entire poses are matched as this allows more flexibility.  
This spatial matching proceeds in three steps as follows:

\vspace{\spcpara}\noindent {\bf Candidate matching patches.}
A random forest classifier similar to~\cite{Charles13a} is trained for
each body part (e.g.\ left shoulder or left wrist) using all annotated frames for that joint.
This {\em personalized body
part detector} is applied to all frames in the video 
to discover candidate (potentially
matching) patches. The random forest classifier is trained on raw RGB
image patches using multiple window sizes,
and is able to take advantage of `opportunistic
features' such as bright colored gloves for detecting the wrists or trouser braces for detecting the shoulders.
Small windows lead to very precise location detection
and larger windows add global context. We found mixing the window
sizes improves generalization, especially when training from a small
number of initial annotations. As the forest classifier has the
ability to average out possible errors in annotation, it adds
robustness to the system and is also very fast to apply.

\vspace{\spcpara}\noindent {\bf Candidate patch verification.}
The candidate match is accepted if its HOG similarity to an original
annotated patch is above a significance threshold.  For this
verification step, an exemplar-SVM is trained to match patches with similar types of body joint configuration (i.e.\ bent elbow, or straight elbow). Configurations are found by k-means clustering RGB patches of annotated joints (typically 200 clusters per joint are used). One exemplar-SVM is trained per configuration medoid. A significance measure can then be computed as in~\cite{Gronat13}, between a candidate patch and each exemplar-SVM. Candidate patches with maximum matching significance (over all centroids) falling below a threshold are discarded (see Fig~\ref{fig:siftflow}(a-b)). 

\vspace{\spcpara}\noindent {\bf Annotation propagation and refinement.}
Annotations are then transferred to patches that match.  However, due to imperfections in the personalized detector, the candidate body joint locations may be a small offset
away from the correct location.  To rectify this, as shown in
Fig~\ref{fig:siftflow}(c), propagated annotations are refined by
registering the matched patches to the annotated patch using SIFTflow~\cite{liu11}, and transforming the annotations using this
registration.

\vspace{-1mm}
\subsection{Temporal propagation}
\vspace{-1mm}
In this stage, annotations (from initialization and spatial matching)
are further spread \emph{temporally} (as illustrated in
Fig~\ref{fig:matching}(c)).  This is achieved by computing dense
optical flow~\cite{Weinzaepfel13} for frames within a temporal window
around the annotated frames.  Starting from these annotated frames,
body joint locations are \emph{temporally propagated} forwards and
backwards along the dense optical flow tracks.  This propagation is
inspired by~\cite{Zuffi13}.  The outcome of this stage is that all
frames within the temporal window (of up to 30 frames before and
after) now have pose annotations. Some annotations may be
incorrect, however, we are able to filter these away as described
next.

\vspace{-1mm}
\subsection{Self-evaluation}
\vspace{-1mm}
In this stage, the quality of the spatially and temporally propagated
annotations is automatically evaluated, and incorrect annotations are
discarded.  To this end, we design a self-evaluation measure which,
for a given annotated frame uses temporal information together with an
occlusion-aware puppet model to detect erroneous annotation. Below we
describe these evaluators in detail:

\vspace{\spcpara}\noindent {\bf Annotation agreement.} 
When there are multiple annotations per frame, originating from different `initially annotated' frames in the video, we can use their level of agreement as a confidence measure.
This measure is formed by looking at the standard deviation of the annotation's 2D location for a given frame and joint. 
If below a threshold, a single annotation is derived from multiple annotations by selecting the 2D location with maximum annotation density (computed using a Parzen window density estimate with a Gaussian kernel). 
On average, after one iteration each joint will have at least two annotations per frame.

\vspace{\spcpara}\noindent {\bf Occlusion-aware puppet model.} 
Due to errors in both the matching and temporal propagation stages, it is possible (particularly in cases of self-occlusion) for propagated detections to drift to background content or other non-joint areas on the person. 
Subsequent iterations of the system would then reinforce these locations incorrectly.
Additionally, due to the independence assumption of the forest part detector, errors can occur due to confusion between left and right wrists. 
Both of these types of errors are alleviated by learning a puppet model for the lower arm appearance, and separate body joint occlusion detectors which check for self-occlusion at head, shoulder and elbow joints. 
A puppet model of the lower arms is used to infer likelihood of an arm given the position of lower arm joints. 
Akin to other puppet models~\cite{Buehler11, Charles11, Zuffi13}, our lower arm puppets are `pulled' around the image space according to proposed lower arm joints, and used to evaluate the underlying image content. 
In our case, the puppet is used to detect when a proposed lower arm position is incorrect or when head, shoulder and elbow joints become occluded. 

\vspace{\spcpara}\noindent {\bf Lower arm puppet construction.} 
The lower arm of a puppet is represented by a rectangle which can be oriented and scaled anisotropically (to model limb foreshortening) according to proposed elbow and wrist joint locations. 
Two linear SVMs are jointly used to classify the image content encompassed by the rectangle as `passed' or `failed', one SVM uses HOG features the other RGB values, both SVMs have to agree on a `pass' decision for the proposed lower arm to pass the evaluation. 
A separate model is trained and evaluated independently for left and right arms. 
The models are trained using initial annotation as positive examples with negative examples generated by adding random offsets to the elbow and wrist annotations. 
Further negatives are created by swapping left and right wrist locations to simulate a hand swap, similar to the method used in \cite{Charles13a}. 

\vspace{\spcpara}\noindent {\bf Occlusion detection.} 
To check for occlusion at head, shoulder and elbow joints, a square window is considered around a joint of interest. 
An SVM using HOG features and another using RGB features is applied. 
A low score from either SVM signifies an occluded joint. 
The SVMs are trained (a pair per joint) using initial un-occluded annotations, determined by considering the body part layout. 
Negative examples are generated from random offsets to these annotations. 
At run-time, if a joint is flagged as occluded we remove its track until flagged as un-occluded. 
The lower arm and occlusion detectors are retrained at each iteration using updated annotations.

\vspace{\spcpara}\noindent {\bf Discarding annotations.} 
The above self-evaluation measures are used to discard annotations that are considered `failed'. 
An annotation (per joint, per frame) is discarded if any one of the above measures falls below a threshold value or classified as `failed' by the lower arm evaluator (see appendix for details). 
The puppet model is also used to discard some initial annotations prior to subsequent stages.

\vspace{\spcpara}\noindent {\bf Correcting failed annotations.} 
Sometimes it is possible to correct lower arm joint detections `failing' the puppet evaluation method. 
This is done by randomly sampling a pair of wrist and elbow points (25 combinations in practice) around the `failed' detection and re-evaluating each pair with the lower arm model . 
If a pair of points `pass' evaluation we accept them as new annotation. 
Correcting failed annotations is beneficial as this leads to improved propagation when iterating the system.

\vspace{-1mm}
\subsection{Personalizing a ConvNet pose estimator} 
\vspace{-1mm}

\label{sec:rf}
The final distributed annotations are used to fine-tune (with back-propagation) the generic ConvNet-based pose estimator of Pfister~\etal~\cite{Pfister15}, with which we initialized our system. 
The ConvNet is fine-tuned using all annotated frames thereby personalizing the ConvNet to the input video. 
Examples of channels that adapt to detect personalized features for a particular video are shown in Fig~\ref{fig:features}.
The net is trained with a fixed learning rate of $1\times10^{-7}$ for 2,000 iterations using a batch  size of 30 frames and momentum set at 0.95. 
This pose estimator is used to predict body joints for \emph{all} frames of the video, possibly correcting any local mistakes that do persevere in the annotation process. 

\def \imgscale {0.58}
\def \imgscaletwo {0.18}
\begin{figure}
\begin{center}

\footnotesize{
\begin{tabular}{c*{3}{@{\hspace{1pt}}c}}
\includegraphics[scale=\imgscaletwo, trim = 40mm 40mm 100mm 0mm,clip]{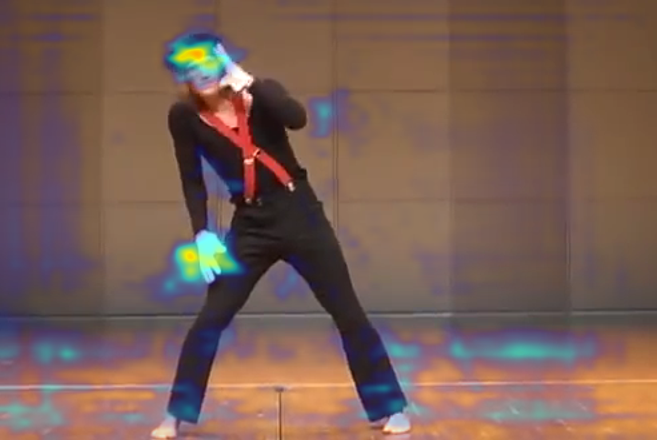}&
\includegraphics[scale=\imgscaletwo, trim = 55mm 40mm 85mm 0mm,clip]{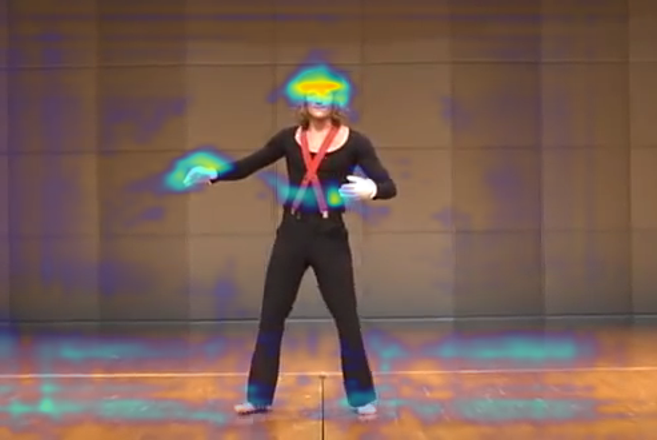}&
\hspace{0.15cm}\includegraphics[scale=\imgscale, trim = 40mm 40mm 50mm 0mm,clip]{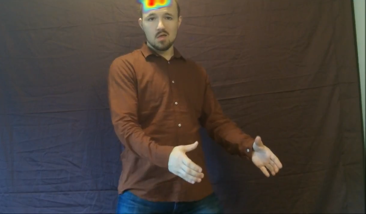}&
\includegraphics[scale=\imgscale, trim = 40mm 40mm 50mm 0mm,clip]{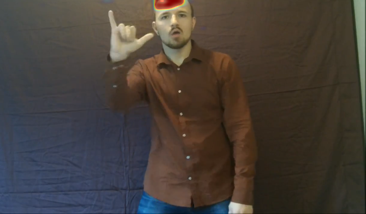}\\
\multicolumn{2}{c}{(a) Layer 2 fusion map} & \multicolumn{2}{c}{(b) Layer 3 fusion map} 
\end{tabular}}

\end{center}
\vspace{-0.2cm}
\caption{{\bf ConvNet filter response maps.} 
Heatmap responses from two different personalized ConvNets shown overlaid on example input frames. 
(a) and (b) are filter responses from layers 2 and 3, respectively, of the spatial fusion layers obtained using~\cite{Pfister15}. 
Personalized features for (a) show that the hat and glove are important, and in (b) hairline is important. 
The original generic ConvNet of the same network shows no personalized features of this type (\ie\ person specific features are added by the fine-tuning.)}
\label{fig:features}
\vspace{\spcfigcaption}
\end{figure}

\vspace{-1mm}
\section{Experiments}
\vspace{-1mm}
We first present the datasets; then evaluate gains from each stage in our method; and finally present a comparison to state of the art.
Experimental details are included in the appendix, and a demo video is online at {\small \url{https://youtu.be/YO1JF8aZ_Do}}.

\vspace{-1mm}
\subsection{Datasets and evaluation}
\vspace{-1mm}
Experiments are performed using three datasets that contain long videos suitable for personalization. 

\vspace{\spcpara}\noindent {\bf YouTube Pose.} 
This new dataset consists of 50 videos of different people from YouTube, each with a single person in the video.  
Videos range from approximately 2,000 to 20,000 frames in length. 
For each video, 100 frames were randomly selected and manually annotated (5,000 frames in total). The dataset covers a broad range of activities, e.g., dancing, stand-up comedy, how-to, sports, disk jockeys, performing arts and dancing sign language signers.

\vspace{\spcpara}\noindent {\bf YouTube Pose Subset.}
A five video subset from YouTube Pose. 
Example frames from the YouTube Pose Subset are shown in Fig~\ref{fig:poseoutput}(a), and further examples from YouTube Pose are shown in the appendix.

\vspace{\spcpara}\noindent {\bf MPII Cooking.} 
This dataset contains video sequences from~\cite{Rohrbach12} for recognizing cooking activities. 
Each video is on average approximately 20,000 frames. 
21 videos come manually annotated with upper body pose, with 1,277 testing frames. 
Each video contains a single person (person varies between videos) captured with a static camera; all sequences are shot in the same kitchen.

\vspace{\spcpara}\noindent {\bf Upper-body YouTube Dancing Pose (UYDP).}
This dataset~\cite{Shen14} consists of 20 short video clips, each containing approximately 100 consecutive frames with annotations.

\vspace{\spcpara}\noindent {\bf BBC Pose.} 
This dataset~\cite{Charles13} contains five one-hour-long videos each with different sign language signers, different clothing and sleeve length. 
Each video has 200 test frames which have been manually annotated with joint locations (1,000 test frames in total). 
Test frames were selected by the authors to contain a diverse range of poses. Additionally, we annotated the location of the nose-tip on test frames. 

\vspace{\spcpara}\noindent {\bf Evaluation measure.} 
The accuracy of the pose estimator is evaluated on ground truth frames. 
An estimated joint is deemed correctly located if it is within a set distance of $d$ pixels from the ground truth. 
Accuracy is measured as the percentage of correctly estimated joints over all test frames. 
For consistency with prior work,  we evaluate on the UYDP dataset using the Average Precision of Keypoints (APK)~\cite{Yang11} at a threshold of 0.2.

Since automatically propagated annotations may not reach all `test' frames with manual ground truth, we measure accuracy of the annotation stages by fine-tuning a ConvNet \cite{Pfister15} using all available annotations, and then evaluating this ConvNet's predictions on all test frames.  
This ensures evaluation consistency and fairness, and is a good indirect measure for annotation performance (as higher-quality training annotations should lead to improved ConvNet pose predictions).

\vspace{-1mm}
\subsection{Component evaluation}
\vspace{-1mm}
We first evaluate the stages of the method on the YouTube Pose Subset.  
Tab~\ref{tab:component} and Fig~\ref{fig:resultspropagation} show the changes in accuracy (at a threshold of 20 pixels) and coverage as the iterations progress, whereas Fig~\ref{fig:compeval} shows accuracy as the allowed distance from manual ground truth $d$ is increased.
\begin{table}
\begin{minipage}{1\textwidth}
\begin{tabular}{l*{5}{@{\hspace{3pt}}c}}
	\multicolumn{6}{c}{YouTube Pose Subset Accuracy (\%) at $d=20$ pixels} \\
	\hline
	Method & Head & Wrsts & Elbws & Shldrs & Average  \\
	\hline
Pfister \etal~\cite{Pfister15}&  74.4 & 59.0 & 70.7 & 82.7 & 71.3\\ 
Chen \& Yuille~\cite{chen14} &  89.4 & 76.5 & 83.1 & 90.7 & 84.3\\ 
Yang \& Ramanan~\cite{Yang11} &  87.2 & 43.1 & 60.6 & 82.5 & 65.7\\ 
Stage 1 - Initial&  95.2 & 61.8 & 69.4 & 86.6 & 75.8\\ 
Stage 2 - Spatial &  95.4 & 72.1 & 80.0 & 91.9 & 83.3\\ 
Stage 3 - Temporal &  96.4 & 79.7 & 82.7 & 94.7 & 87.2\\ 
Personalized ConvNet &  {\bf 97.6} & {\bf 88.6} & {\bf 84.7} & {\bf 96.5} & {\bf 91.0}\\ 
	\hline 
	\end{tabular}
\end{minipage}
	\vspace{0cm}
\caption{\small {\bf Component analysis on YouTube Pose Subset.} 
Accuracy at each stage of the method compared to baseline algorithms. 
Personalized ConvNet results are shown after 5 iterations. Corresponding curves are given in Fig~\ref{fig:compeval}.} \label{tab:component} 
\vspace{-0.3cm}
\end{table}

\noindent {\bf Accuracy:} 
Each stage leads to a significant improvement in performance across all body joints.  
Even after Stage 1 (initialization) we improve upon the generic ConvNet of Pfister~\etal~\cite{Pfister15}, demonstrating that fine-tuning with relatively few annotations (from the generic ConvNet and Yang and Ramanan arm detection) brings benefits.  
Stage 2 (spatial matching) yields further gains in accuracy. 
One of the reasons spatial matching is so beneficial is that it helps propagate annotation to frames with similar poses but different local background content -- in the YouTube videos the person moves against a static scene with some videos using a moving camera and containing shots from different angles.
Personalizing a pose estimator from annotations at this stage, therefore, introduces more invariance to background content.
Stage 3 (temporal propagation), is another mechanism for reaching unannotated frames containing different poses (more so than stage 2) from previous stages. 
Thus at this stage, we begin increasing the variation of poses that can be recognized. 
Again, this leads to an increase in performance. 

The main causes of failure for Stage 3 are heavy self occlusion and optical flow errors,
causing propagated annotation to drift to background content.  
In some cases annotations from different frames in the video can drift to the same background location, and (incorrectly) `pass' the annotation agreement measure of Stage 4.  
However, the occlusion-aware puppet model effectively detects and removes these errors, permitting subsequent iterations to progress without propagating errors.  
This is evident from the increase in accuracy when iterating the stages, most notably for the wrist joints which reach near 90\% after five iterations.

\def \imgscale {0.29}
\begin{figure*}
\begin{center}
\begin{tabular}{c*{4}{@{\hspace{2pt}}c}}
\hspace{-0.26cm}\includegraphics[scale=\imgscale]{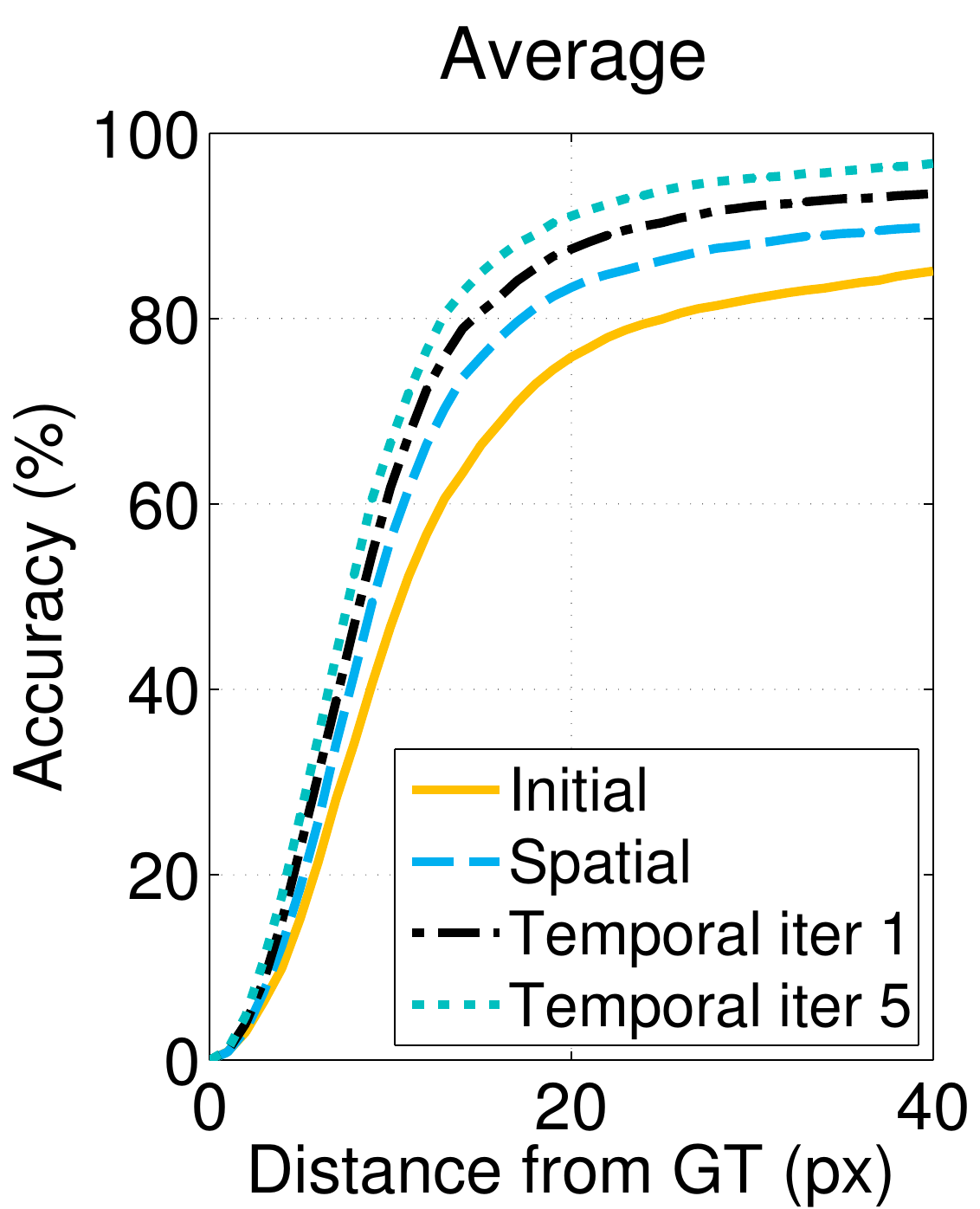}&
\includegraphics[scale=\imgscale]{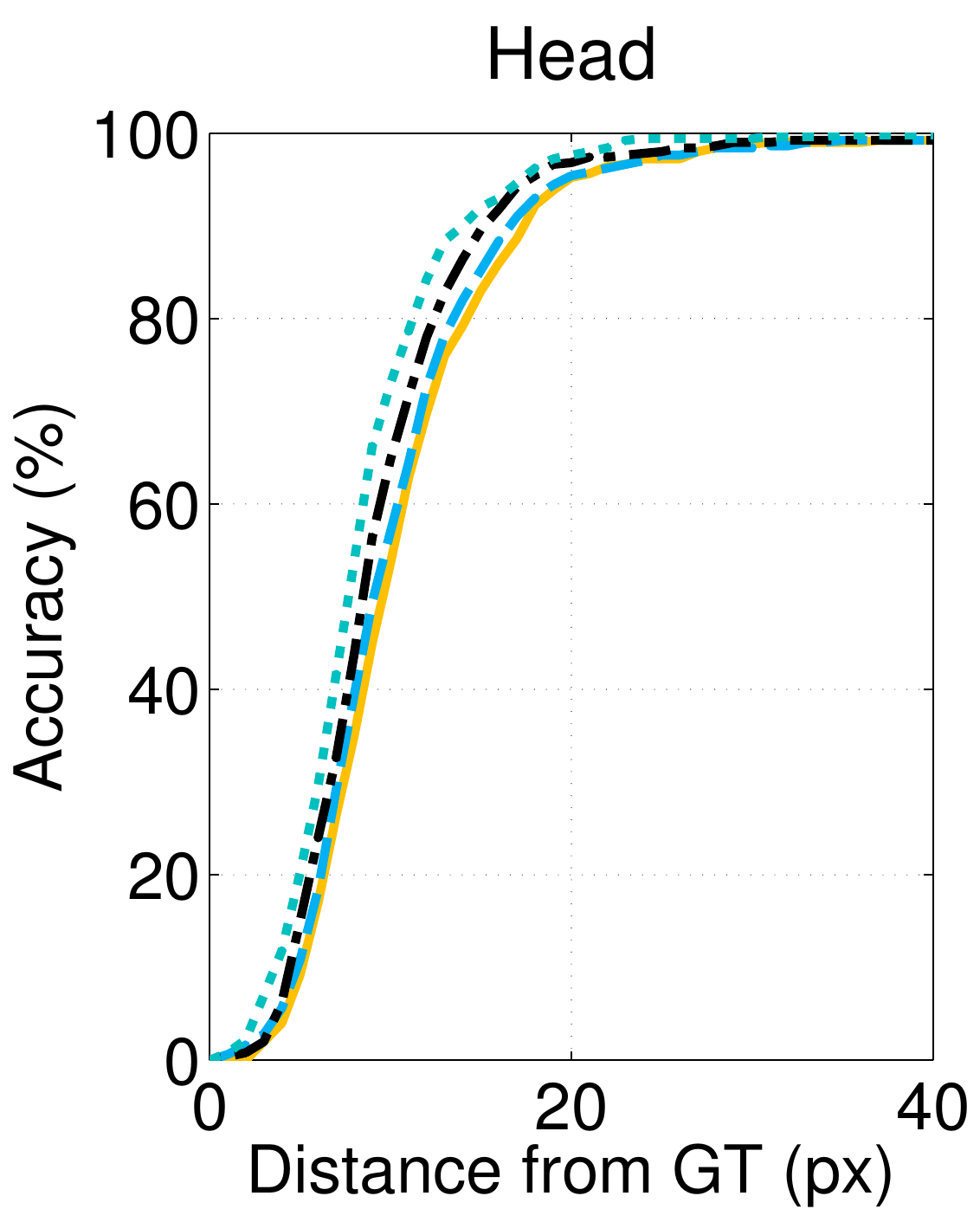} &
\includegraphics[scale=\imgscale]{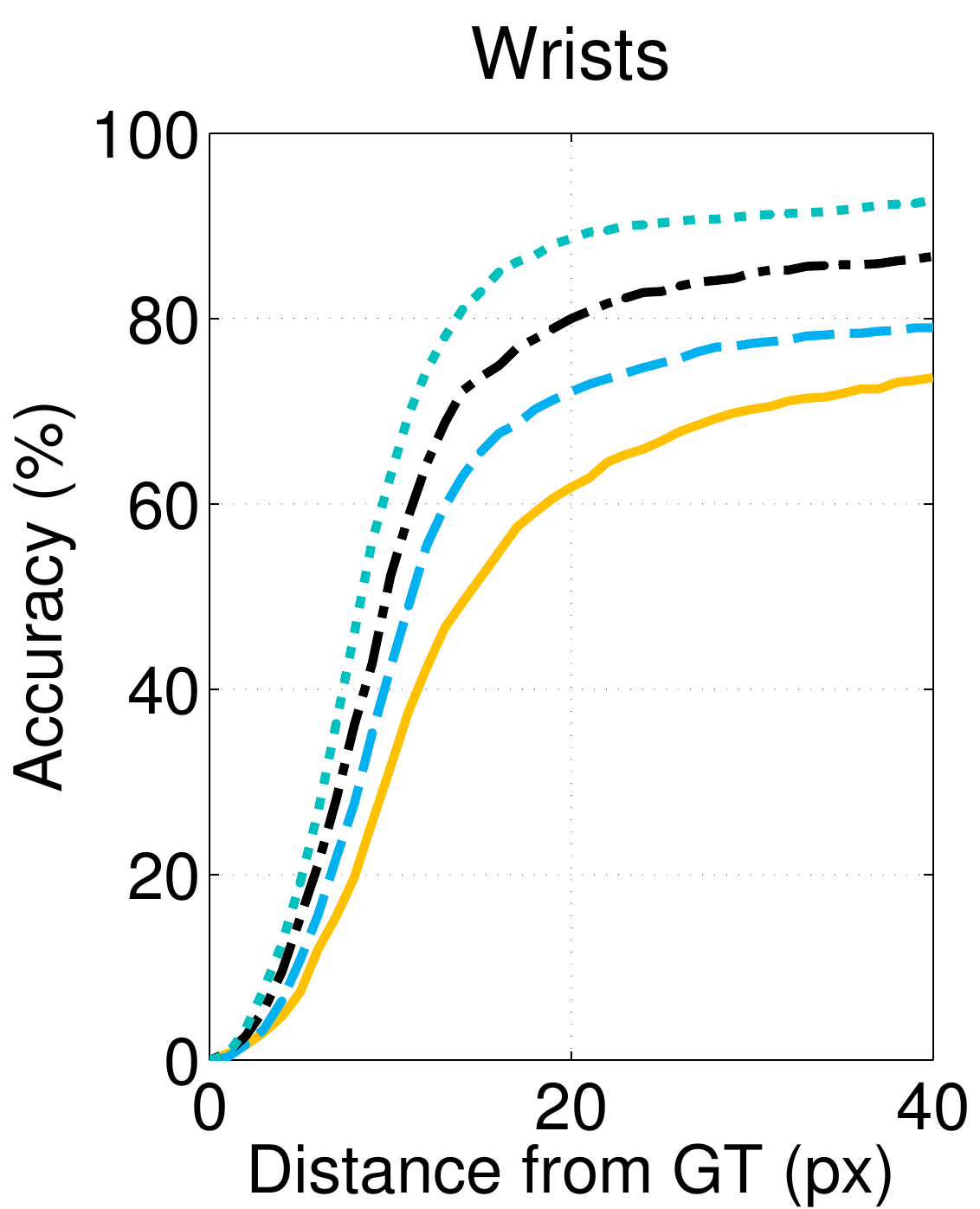} &
\includegraphics[scale=\imgscale]{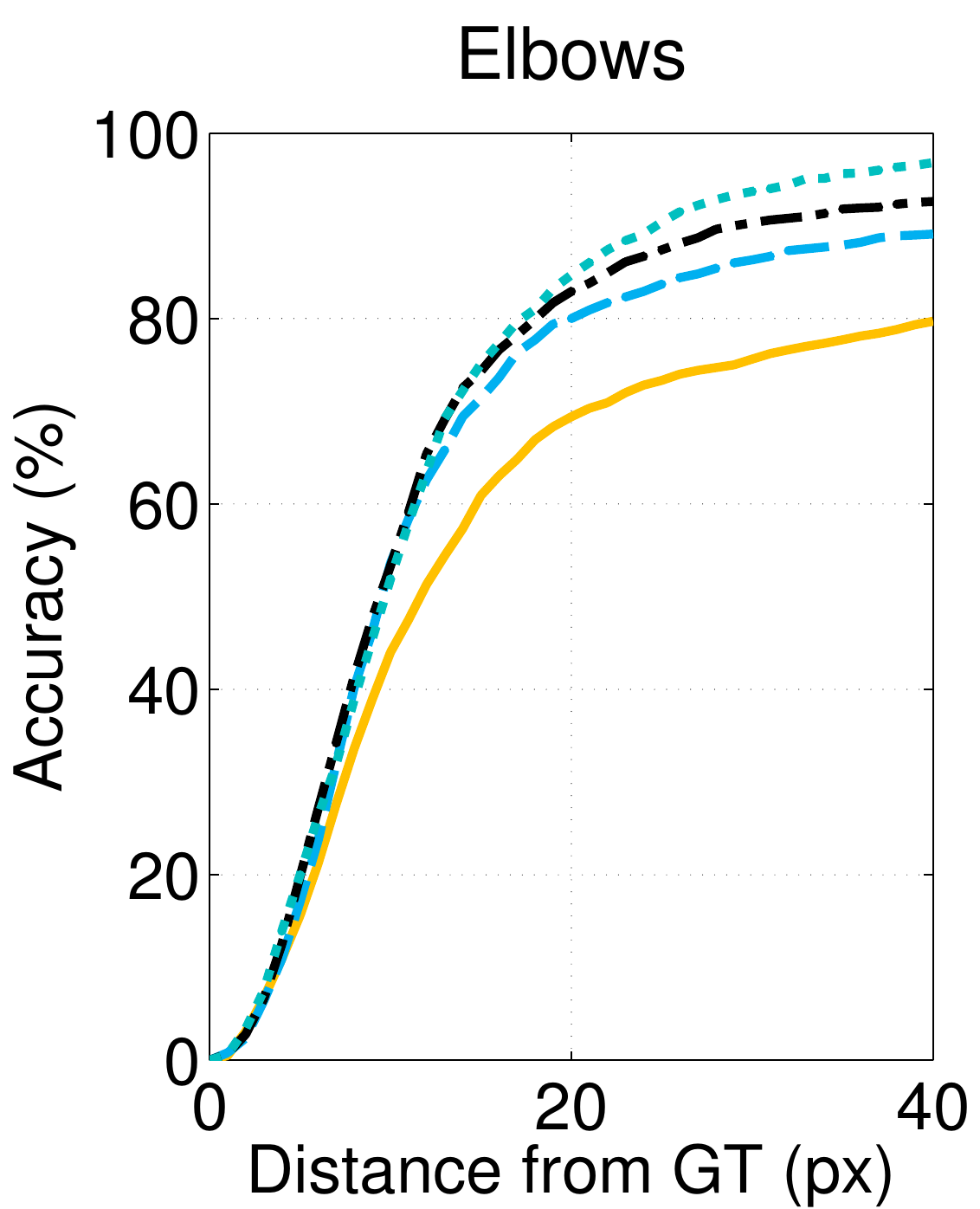}&
\includegraphics[scale=\imgscale]{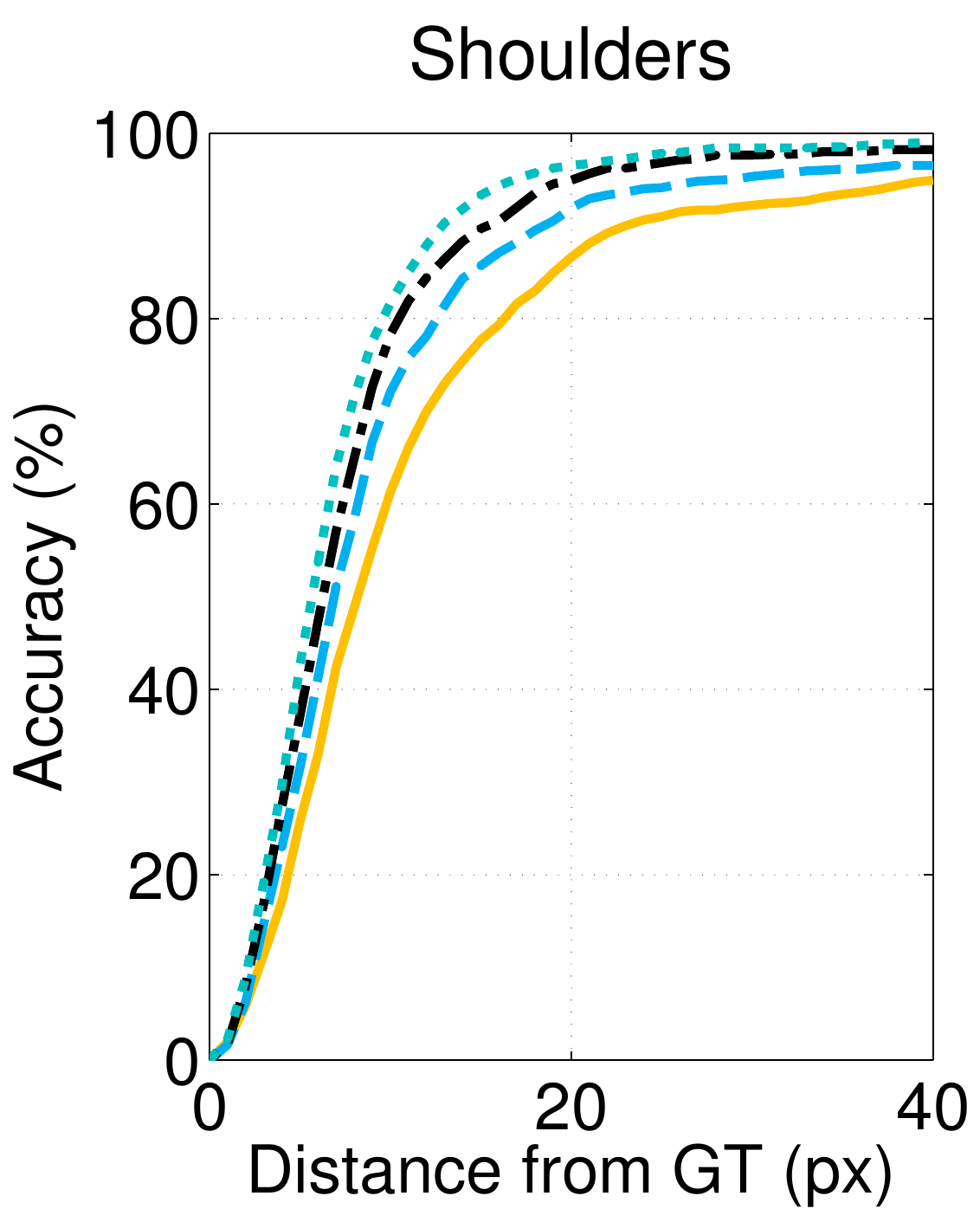}
\end{tabular}
\end{center}
\vspace{-0.2cm}
\caption{{\bf Component evaluation on the YouTube Pose Subset dataset.}
The graphs show the improvement from each stage of the algorithm. 
Notice how each stage leads to a very significant increase in accuracy. 
Accuracy is shown (averaged over left \& right body parts) as the allowed distance from ground truth is increased.}
\label{fig:compeval}
\end{figure*}

\noindent {\bf Coverage:} 
The number of initially annotated frames across each video varies greatly and is highly dependent on the types of pose being performed and camera angle used. 
More initial annotations are obtained for videos where people have their hands down, such as the disc jokey sequences.
For videos containing a high pose variation there is a greater increase in coverage over the iterations.

Fig~\ref{fig:resultspropagation} shows the percentage of video frames (on Youtube Pose Subset) that have been annotated after each iteration. 
A typical video of 10,000 frames, starting with 500 initially annotated, can rapidly increase annotation coverage in just one iteration. 
Stage 2 (spatial matching) normally doubles the annotations. 
The biggest boost in coverage comes from stages 3 and 4 (temporal propagation followed by checking with self-evaluation), with annotated frames rapidly increasing to over 60\% (6,000 frames) of the video. 
Gains in coverage after temporal propagation are dependent upon the size of the temporal window (a larger temporal window improves coverage but can result in decreased accuracy (before self-evaluation) as errors in optical flow compound). 
A temporal window of 30 frames is selected as the trade-off between coverage and accuracy. 
Subsequent iterations result in close to 100\% of the frames being annotated for head and shoulder joints with wrist and elbow annotations covering 85\% (8,500 frames) of the video. 
Relative gains in coverage decrease at each iteration because the system is tackling increasingly more difficult-to-detect poses.

\vspace{\spcpara}\noindent {\bf Timings:} 
Timings for training are amortized over the application costs per frame for a typical 10k frame video, processed on a single core Intel Xeon 2.60GHz. 
Stage 1 (the most expensive stage) takes ${\raise.19ex\hbox{$\scriptstyle\sim$}}15$ seconds per-frame ($s/f$) for initialization as multiple arm-models have to be applied. 
Stage 2 performs both model training and matching in ${\raise.19ex\hbox{$\scriptstyle\sim$}}8s/f$. 
Stage 3 computes in  ${\raise.19ex\hbox{$\scriptstyle\sim$}}8s/f$ and self-evaluation in ${\raise.19ex\hbox{$\scriptstyle\sim$}}4s/f$. 
Further iterations are quicker as less of the video requires spatial matching.
	
\begin{table}
\begin{minipage}{1\textwidth}
\begin{tabular}{l*{5}{@{\hspace{3pt}}c}}
	\multicolumn{6}{c}{YouTube Pose Accuracy (\%) at $d=20$ pixels} \\
	\hline
	Method & Head & Wrsts & Elbws & Shldrs & Average  \\
	\hline
Pfister \etal~\cite{Pfister15} &  89.3 & 64.2 & 74.6 & 85.8 & 76.9\\ 
Chen \& Yuille~\cite{chen14} &  85.7 & 78.8 & 83.5 & 87.3 & 83.5\\ 
Yang \& Ramanan~\cite{Yang11} &  89.9 & 38.5 & 58.3 & 85.3 & 64.9\\ 
Cherian \etal~\cite{Cherian14} &  - & 54.3 & 66.9 & 84.7 & -\\ 
Personalized ConvNet &  {\bf 95.4} & {\bf 86.1} & {\bf 86.8} & {\bf 93.9} & {\bf 89.9}\\ 
\hline
	\multicolumn{6}{c}{MPII Cooking Accuracy (\%) at $d=20$ pixels}\\
	\hline
Pfister \etal~\cite{Pfister15} &  53.4 & 79.0 & 70.4 & 57.0 & 66.6\\ 
Chen \& Yuille~\cite{chen14} &  62.3 & 75.5 & 73.8 & 72.7 & 72.3\\ 
Yang \& Ramanan~\cite{Yang11} &  46.7 & 37.5 & 43.7 & 46.1 & 43.0\\ 
Rohrbach \etal~\cite{Rohrbach12} &  80.5 & 66.2 & 67.1 & 72.2 & 70.2\\ 
Personalized ConvNet &  {\bf 86.7} & {\bf 85.8} & {\bf 80.4} & {\bf 76.3} & {\bf 81.7}\\ 
	\hline
\noalign{\vskip 0.0cm} 
	\multicolumn{6}{c}{UYDP (\%) at APK=0.2}\\
	\hline
Pfister~\etal~\cite{Pfister15} &  78.7 & - & 35.2 & 63.3 & -\\ 
Chen \& Yuille~\cite{chen14} &  86.3 & - & 46.8 & 80.3 & - \\ 
Yang \& Ramanan~\cite{Yang11} &  81.7 & -& 17.6 & 66.5 & -\\ 
Shen~\etal~\cite{Shen14} &  90.9 & - & 33.3 &  83.5 & -\\ 
Personalized ConvNet &  {\bf 91.7} & - & {\bf 57.6} & {\bf 83.8} & -\\ 
	\hline
\noalign{\vskip 0.0cm} 
	\multicolumn{6}{c}{BBC Pose Accuracy (\%) at $d=6$ pixels}\\
	\hline
Pfister \etal~\cite{Pfister15} &  97.1 & 78.6 & 88.2 & 83.0 & 85.3\\ 
Chen \& Yuille~\cite{chen14} &  65.9 & 47.9 & 66.5 & 76.8 & 64.1\\ 
Yang \& Ramanan~\cite{Yang11} &  91.6 & 27.6 & 66.0 & 81.0 & 63.0\\ 
Charles \etal~\cite{Charles13} &  98.2 & 59.9 & 85.3 & 88.6 & 80.8\\ 
Personalized ConvNet &  {\bf 99.5} & {\bf 93.5} & {\bf 95.5} & {\bf 95.9} & {\bf 95.6}\\ 
	\hline
\end{tabular}
\end{minipage}
	\vspace{0cm}
\caption{\small {\bf Evaluation of accuracy over the four datasets.} 
Accuracy is the percentage of correctly estimated body joints within a distance $d$ pixels from ground truth (wrist width approx 15 pixels on average on YouTube Pose and MPII cooking, and 8 pixels on BBC pose).
Results are averaged over all videos with ground truth from each dataset. 
Note, for Cherian \etal~~\cite{Cherian14}, head estimates are not comparable with other methods; and
for UYPD there is a problem with the evaluation script for wrists.
}  \label{tab:quantity} 
\end{table}

\def \imgscalea {0.22}
\def \imgscaleb {0.20}
\def \imgscalec {0.398}
\def \imgscaled {0.285}
\def \imgscalee {0.34}
\def \imgscalef {0.228}
\begin{figure*}[ht!]
\begin{center}
\footnotesize{
\begin{tabular}{c*{6}{@{\hspace{2pt}}c}}
\includegraphics[scale=\imgscalea, trim = 00mm 0mm 20mm 0mm,clip]{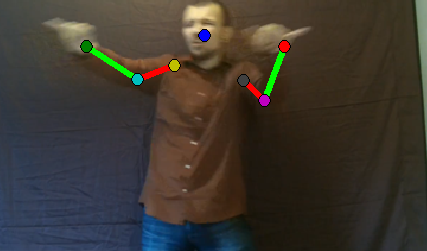}&
\includegraphics[scale=\imgscalea, trim = 00mm 0mm 40mm 0mm,clip]{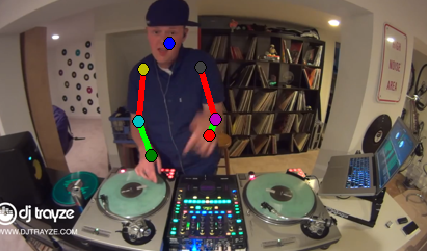}&
\includegraphics[scale=\imgscaleb]{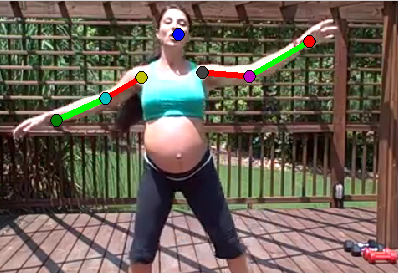}&
\includegraphics[scale=\imgscalec, trim = 55mm 30mm 60mm 10mm,clip]{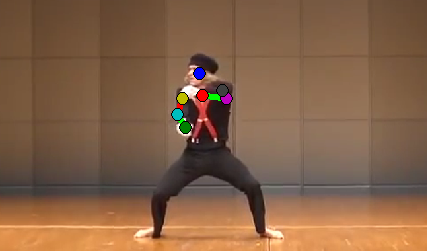}&
\includegraphics[scale=\imgscalef, trim = 45mm 2mm 5mm 2mm,clip]{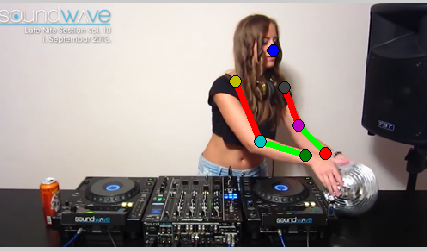}&
\hspace{0.4cm}\includegraphics[scale=\imgscalee, trim = 95mm 2mm 0mm 30mm,clip]{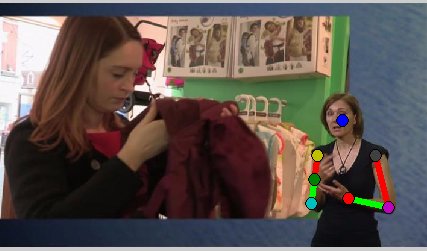} &
\hspace{0.4cm}\includegraphics[scale=\imgscaled, trim = 35mm 11mm 45mm 10mm,clip]{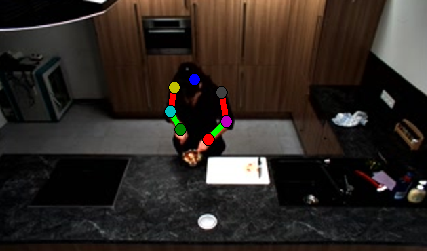}\\
\multicolumn{5}{c}{(a) YouTube Pose} &\hspace{0.4cm}(b) BBC Pose & \hspace{0.4cm}(c) MPII Cooking 
\end{tabular}}
\end{center}
\vspace{-0.2cm}
\caption{{\bf Example pose estimates} on frames from three datasets (a) YouTube Pose , (b) BBC Pose and (c) MPII Cooking. 
Note the variety of poses, clothing, and body shapes in the YouTube videos.}
\label{fig:poseoutput}
\vspace{-3mm}
\end{figure*}

\vspace{-1mm}
\subsection{Comparison to baselines and state of the art}
\vspace{-1mm}

As baselines we compare against two ConvNet-based estimators from Chen \& Yuille~\cite{chen14} and Pfister~\etal~\cite{Pfister15}, and the deformable parts-based model by Yang \& Ramanan~\cite{Yang11}. 
All baseline pose estimators are trained for upper-body pose detection on the FLIC dataset~\cite{Modec13}. 
Additionally,  on YouTube Pose we compare against Cherian~\etal~\cite{Cherian14} (trained on FLIC); and to Charles~\etal~\cite{Charles13}, Rohrbach~\etal~\cite{Rohrbach12} and Shen~\etal~\cite{Shen14} on BBC Pose, MPII Cooking and UYDP, respectively. On BBC Pose, head location accuracy is evaluated for Charles~\etal~\cite{Charles13} using \emph{head center of mass} ground truth (as this is how the model is trained), all other models are evaluated against \emph{nose-tip}.

Results are given in Tab~\ref{tab:quantity} and Fig~\ref{fig:resultsyoutube}.  
The results from the baselines indicate that MPII is the most challenging dataset of the three. 
As is evident, personalization achieves a huge improvement over both the baselines and state of the art. 
For example, obtaining 86.1\% accuracy for wrist detection on YouTube Pose and an astonishing 93.5\% accuracy for wrist detection on BBC Pose -- significantly increasing over the state of the art results of 59.9\% of~\cite{Charles13} and 78.6\% of~\cite{Pfister15}. The boost in performance of the generic ConvNet estimator by fine-tuning using personalization, that was noted on the YouTube dataset, is also repeated here on the BBC and MPII datasets. 
For example, increasing average prediction accuracy from 66.6\% to 81.7\% on MPII. 
Personalization leverages many frames in long videos; yet even for short videos such as UYDP we see an increase in accuracy and perform particularly well for the elbow joints.

In comparing stages of the algorithm on YouTube Pose Subset (Tab~\ref{tab:component} and Fig~\ref{fig:resultspropagation}), by Stage 2 (spatial matching) the head and shoulder accuracy already exceeds all baselines. 
By Stage 3 (temporal propagation), the system outperforms all baselines across all body joints. 
As mentioned above, spatial matching helps propagate annotations to frames with similar poses but different local background content. 
This occurs frequently in the BBC Pose dataset since signers are overlaid on a moving background in broadcasts.

\def \imgscale {0.29}
\def \imgscaletwo {0.3}
\begin{figure*}
\begin{center}
\begin{tabular}{c*{6}{@{\hspace{2pt}}c}}
\hspace{-0.25cm}\raisebox{10.8mm}{\includegraphics[scale=\imgscaletwo, trim = 0mm 122mm 15mm 0mm,clip]{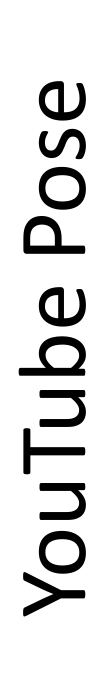}}&
\includegraphics[scale=\imgscale]{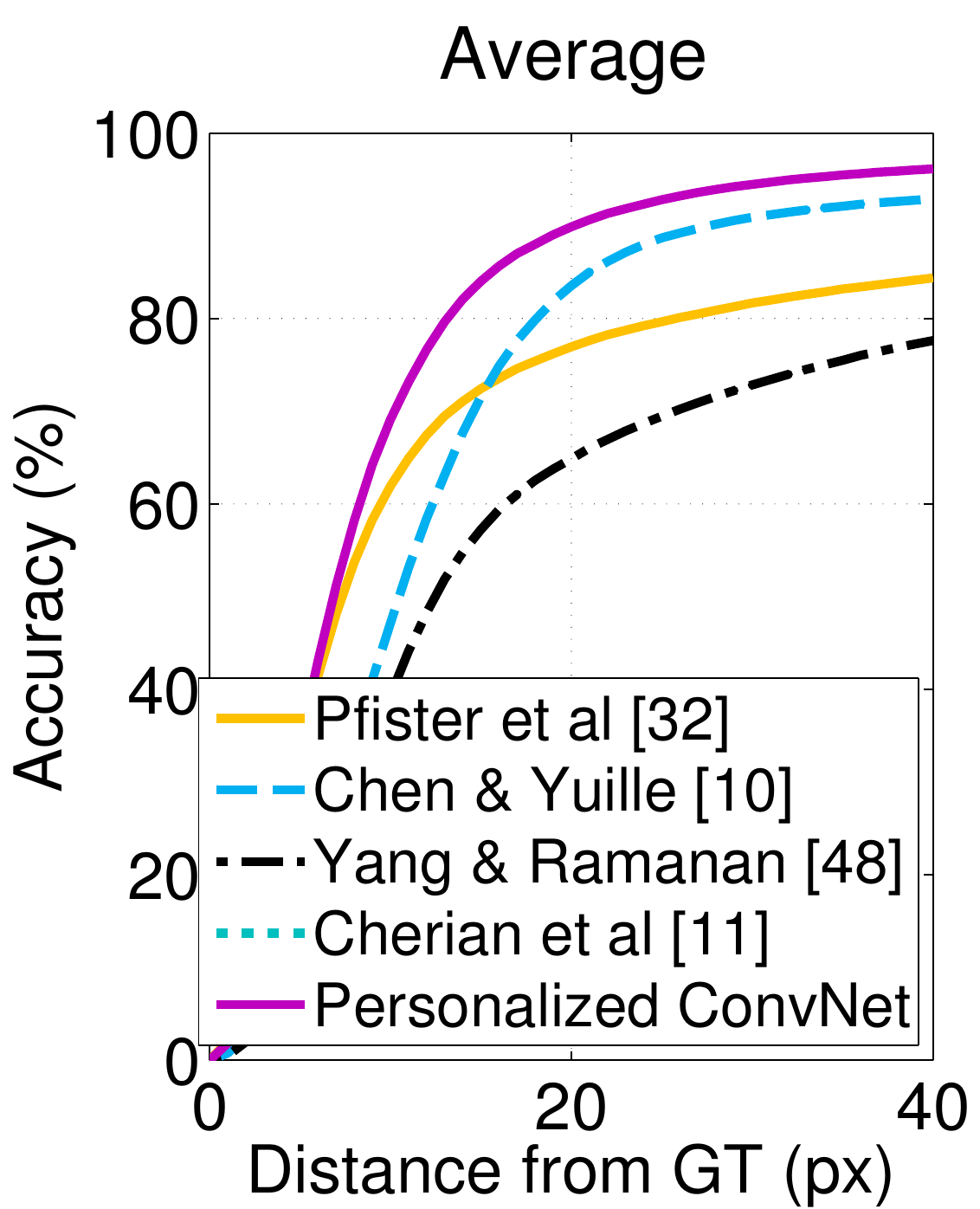}&
\includegraphics[scale=\imgscale, trim = 10mm 0mm 0mm 0mm,clip]{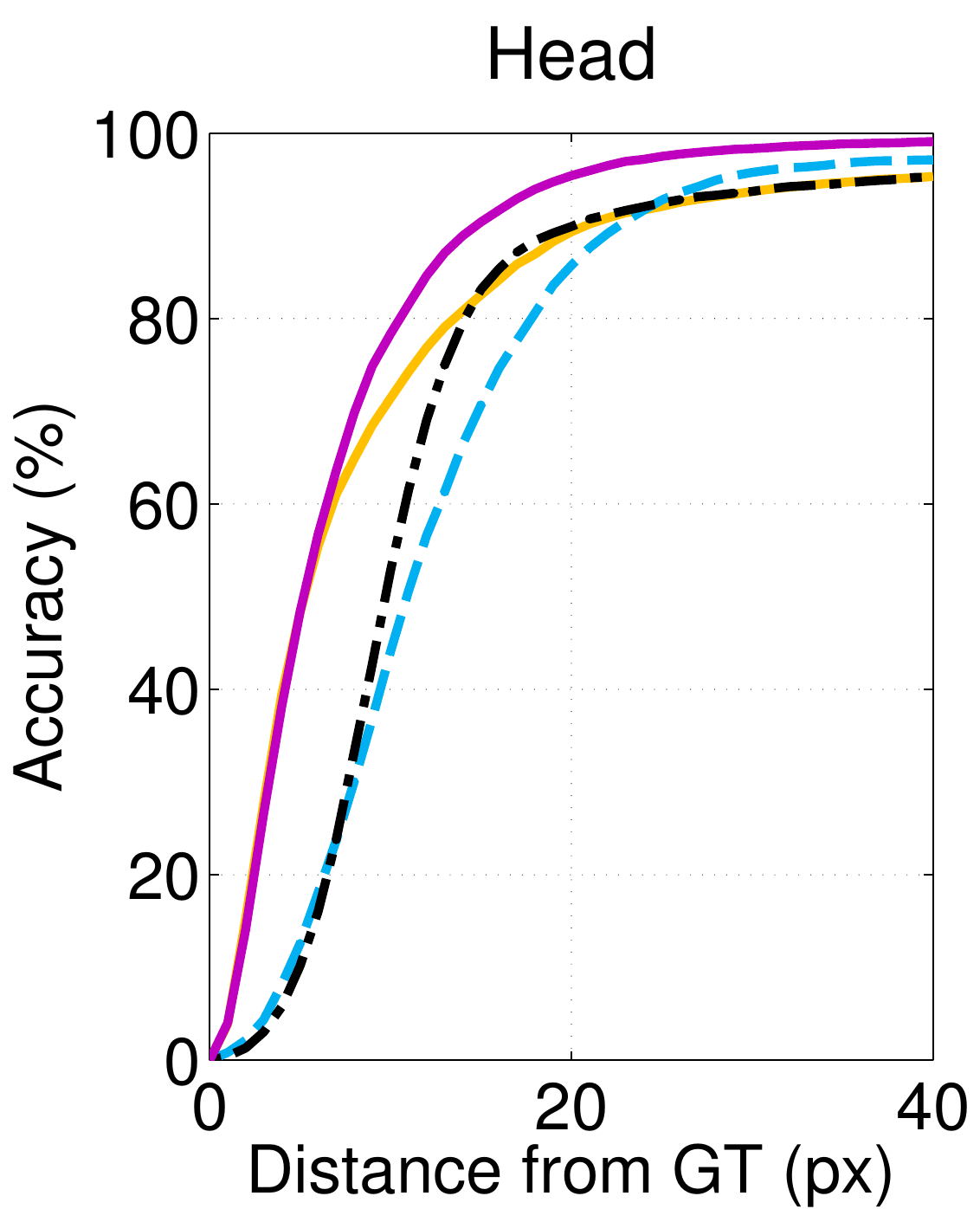}&
\includegraphics[scale=\imgscale, trim = 10mm 0mm 0mm 0mm,clip]{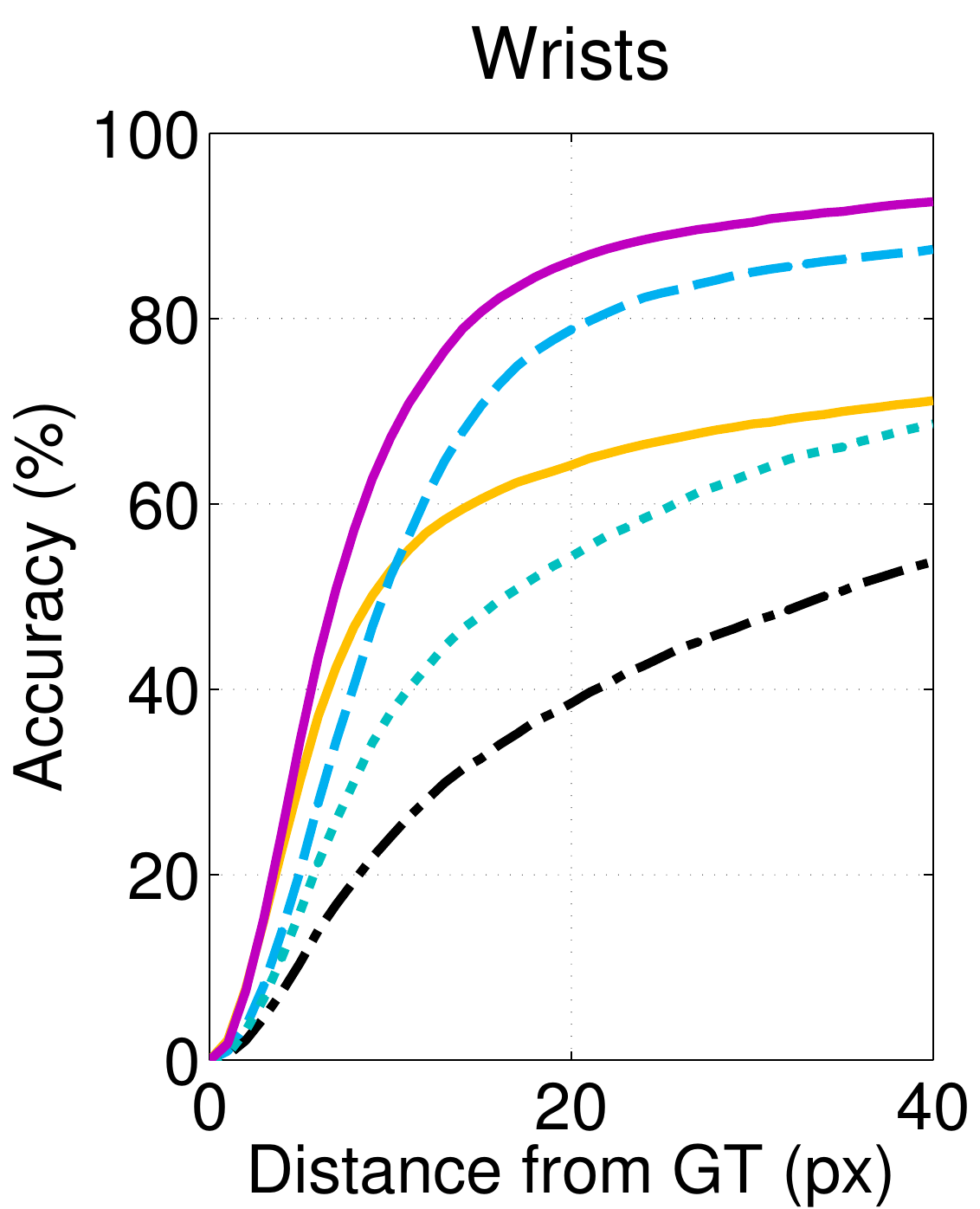}&
\includegraphics[scale=\imgscale, trim = 10mm 0mm 0mm 0mm,clip]{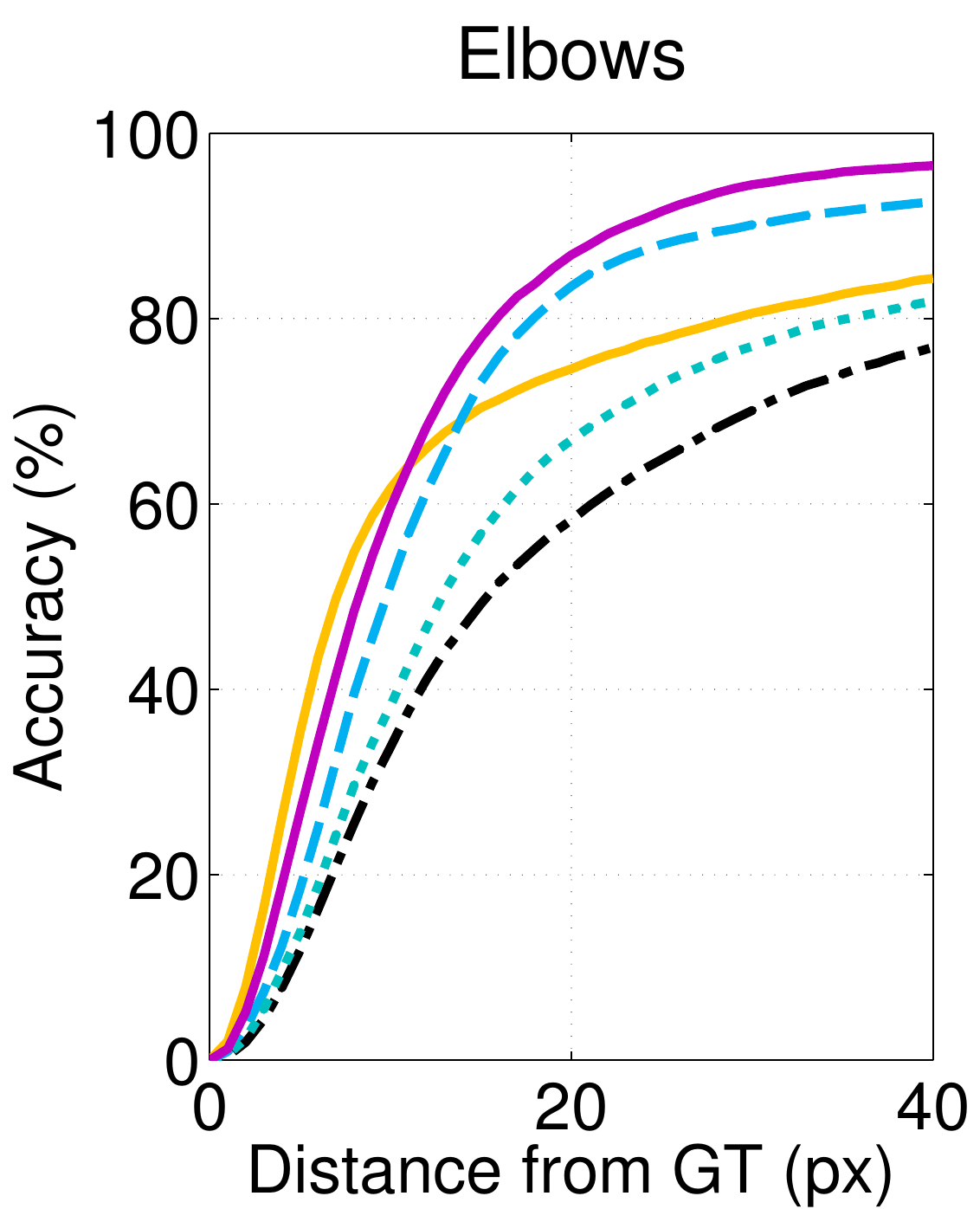}&
\includegraphics[scale=\imgscale, trim = 10mm 0mm 0mm 0mm,clip]{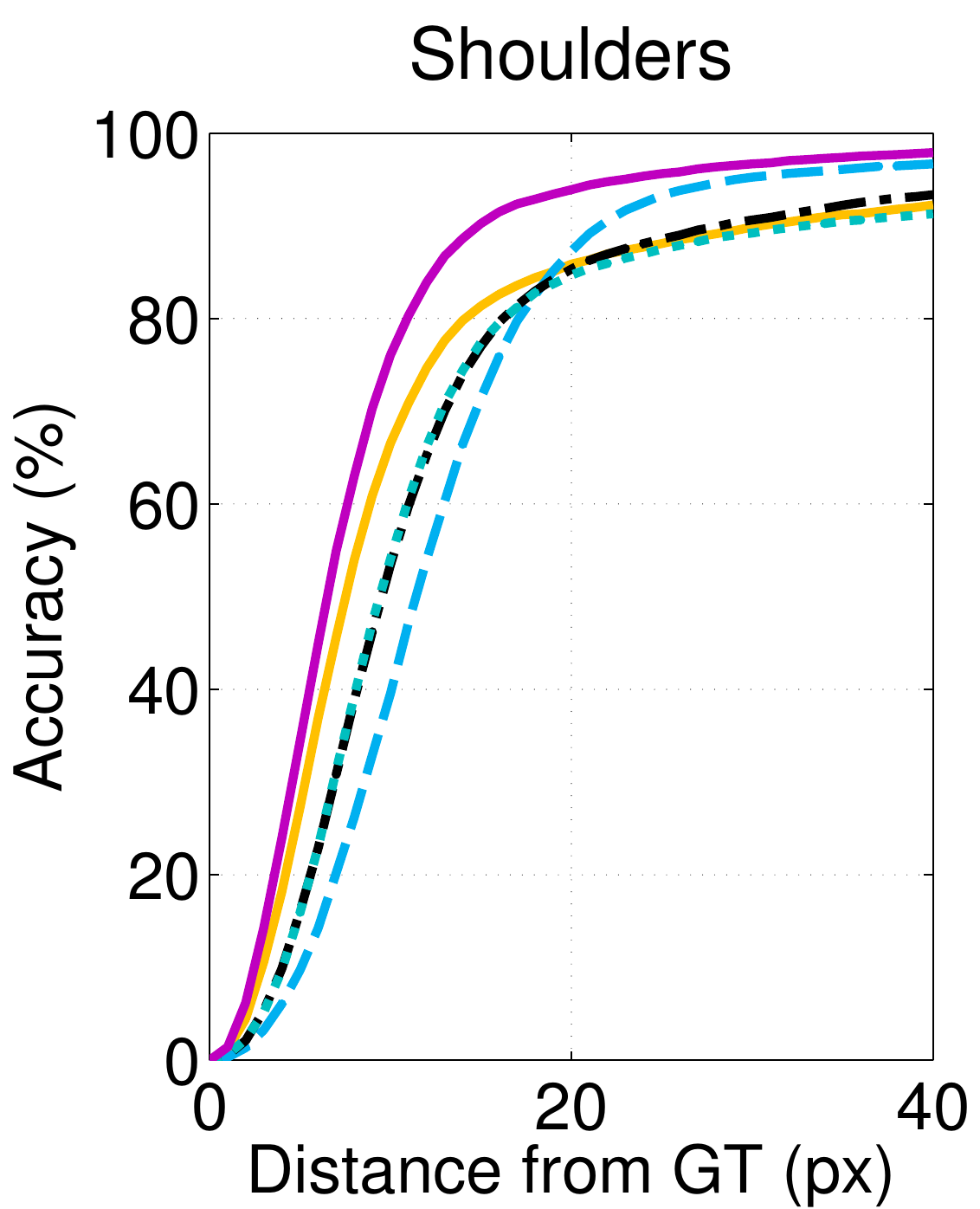} 
 \\
\hspace{-0.25cm}\raisebox{11.3mm}{\includegraphics[scale=\imgscaletwo, trim = 0mm 0mm 15mm 140mm,clip]{graph_titles.pdf}}&
\includegraphics[scale=\imgscale]{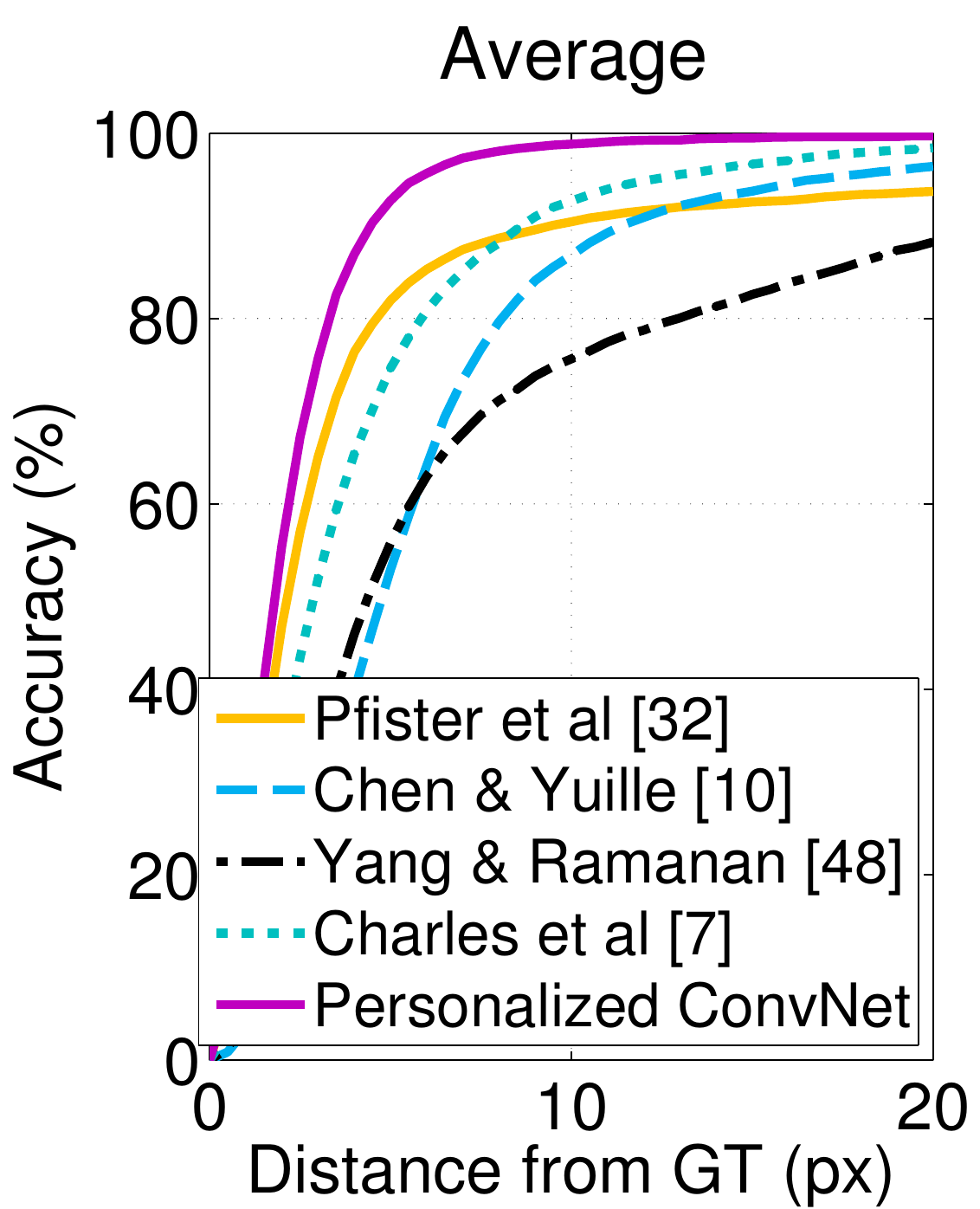}&
\includegraphics[scale=\imgscale, trim = 10mm 0mm 0mm 0mm,clip]{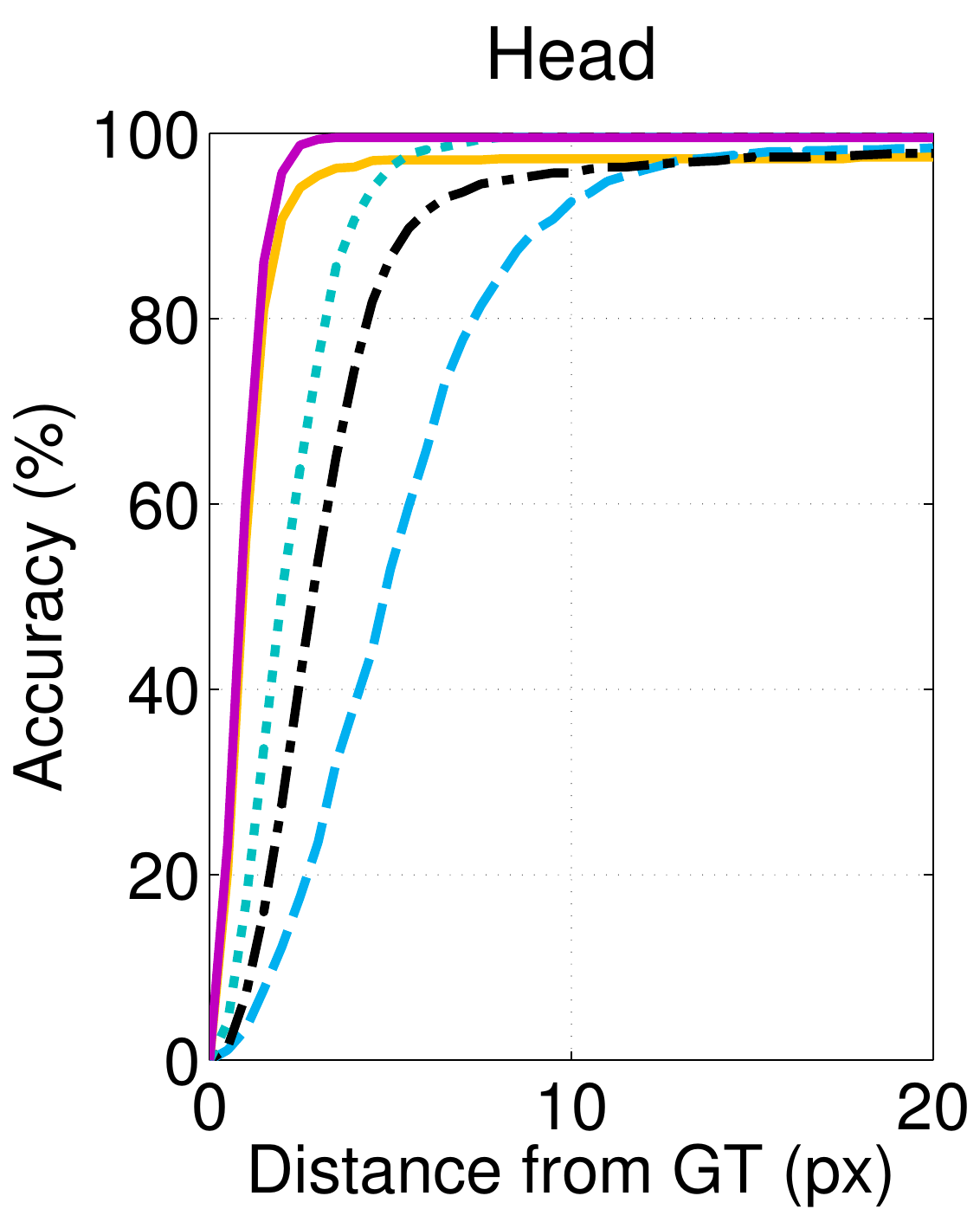} &
\includegraphics[scale=\imgscale, trim = 10mm 0mm 0mm 0mm,clip]{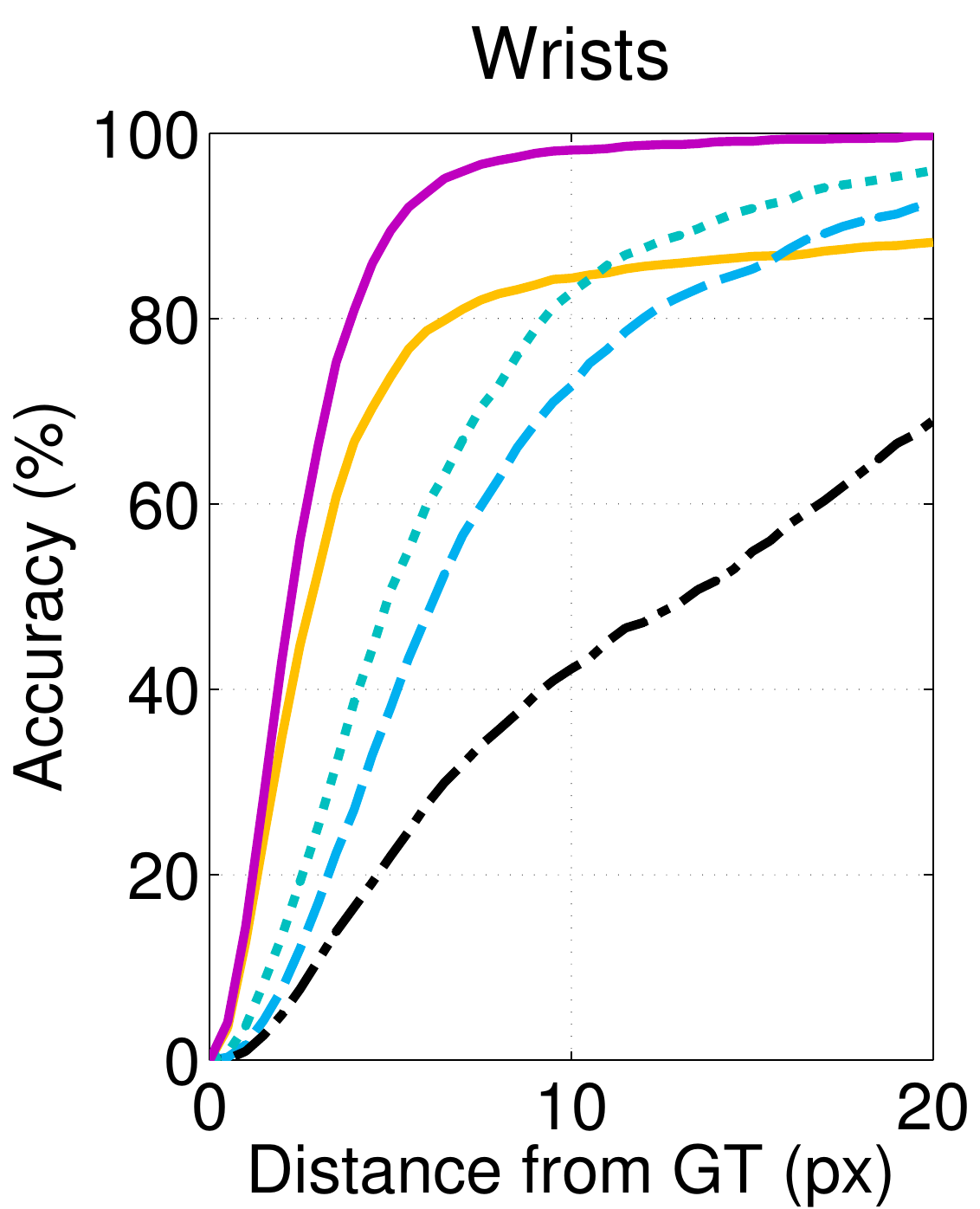} &
\includegraphics[scale=\imgscale, trim = 10mm 0mm 0mm 0mm,clip]{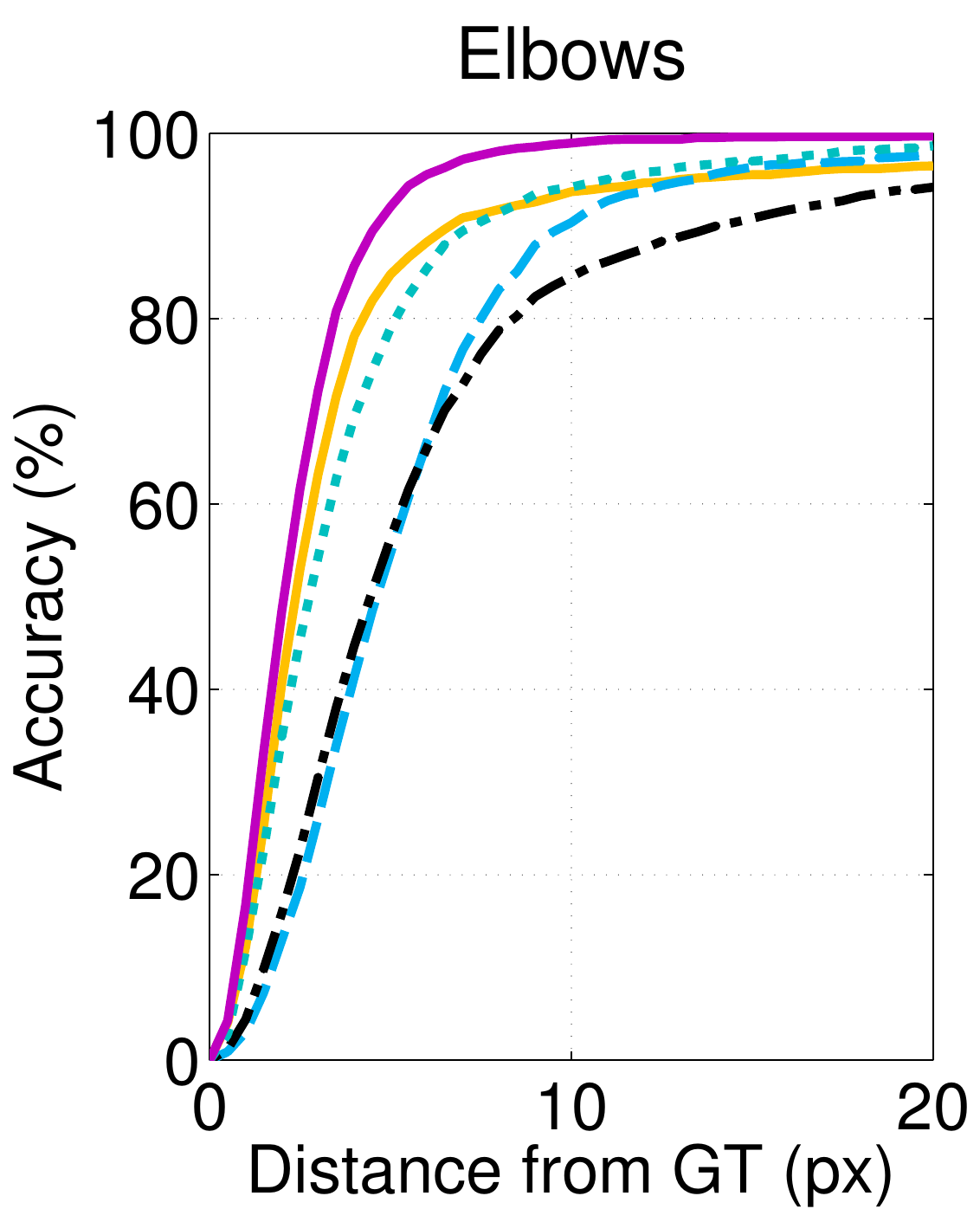} &
\includegraphics[scale=\imgscale, trim = 10mm 0mm 0mm 0mm,clip]{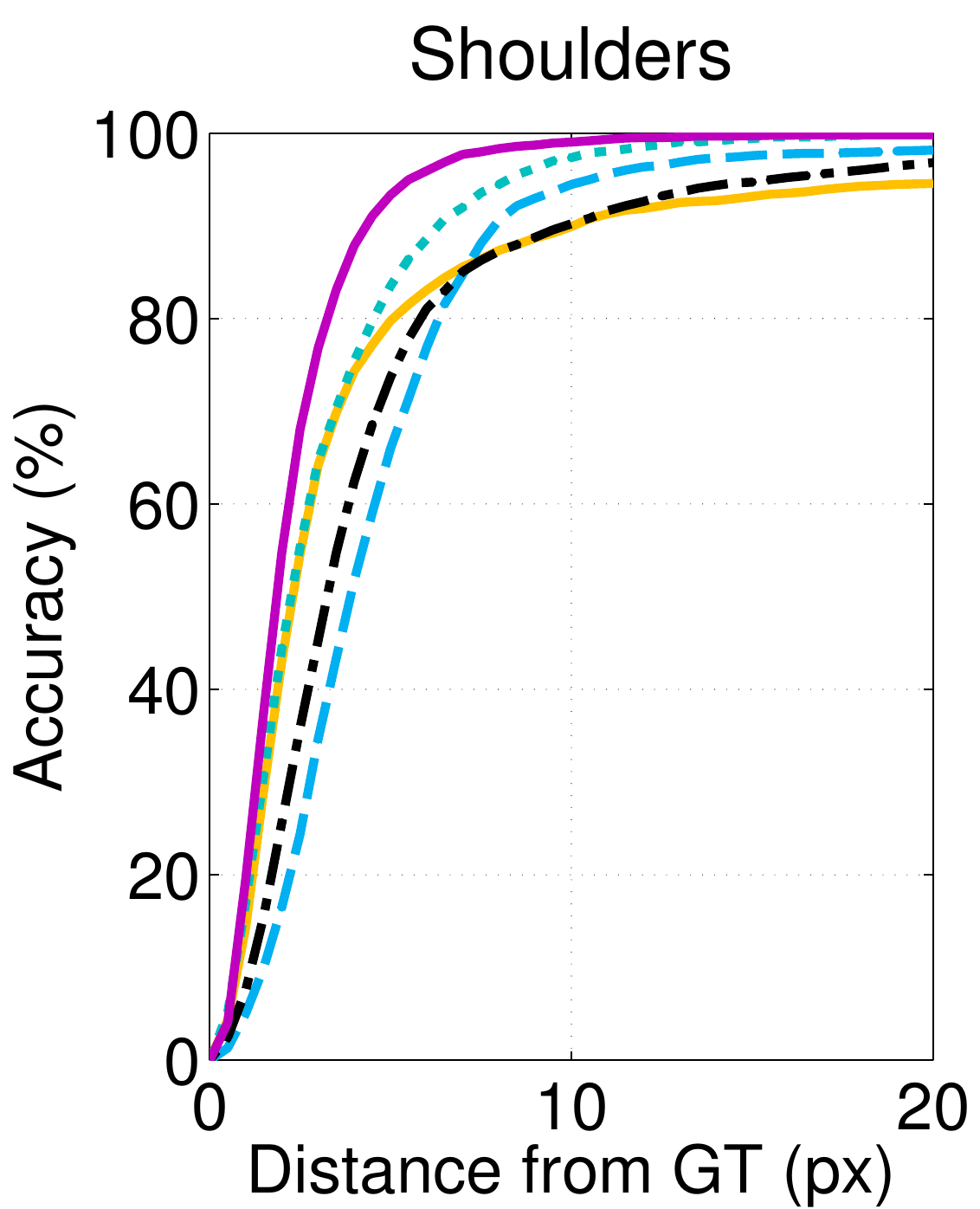}\\
\hspace{-0.25cm}\raisebox{10.5mm}{\includegraphics[scale=\imgscaletwo, trim = 0mm 60mm 15mm 70mm,clip]{graph_titles.pdf}}&
\includegraphics[scale=\imgscale]{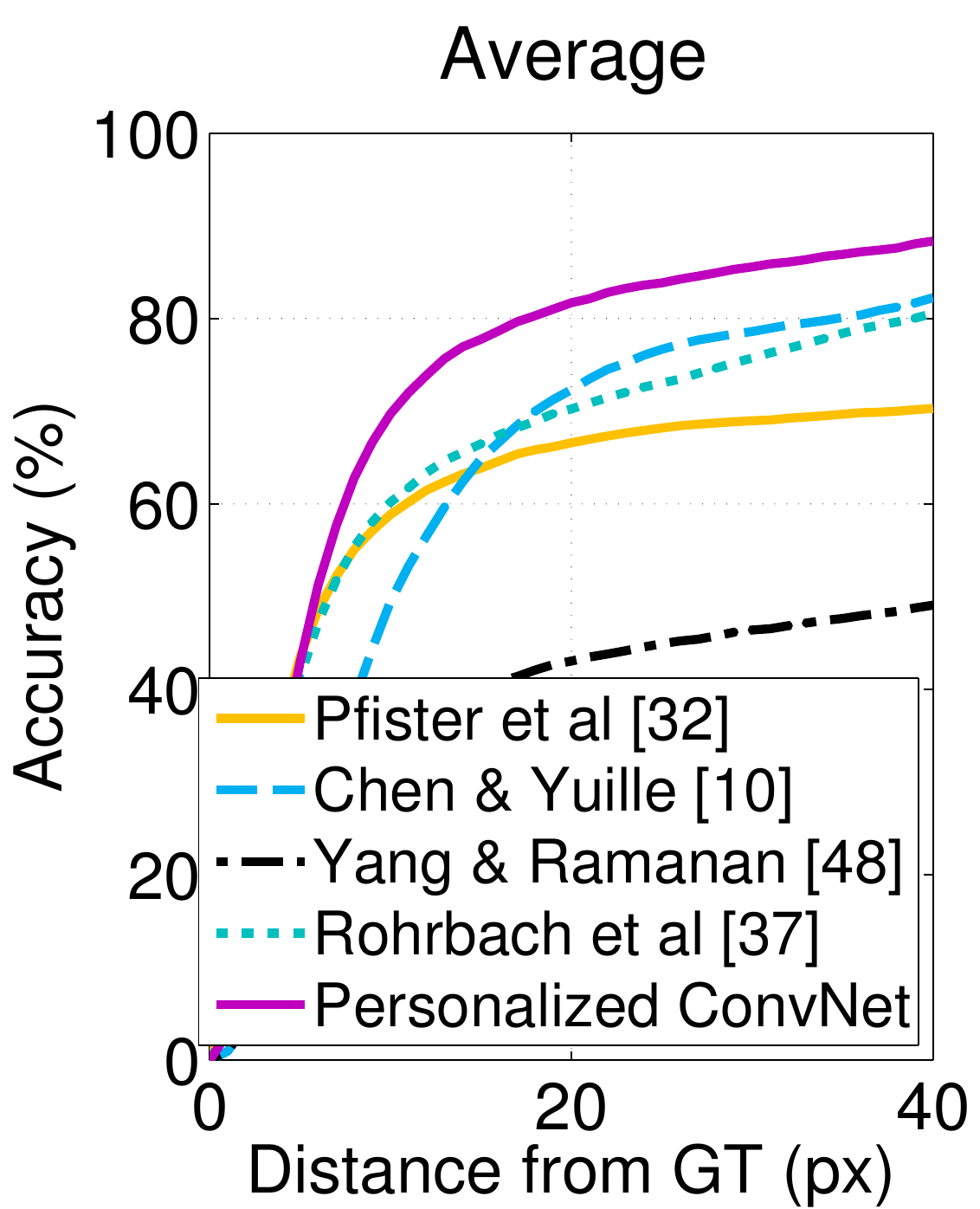}&
\includegraphics[scale=\imgscale, trim = 10mm 0mm 0mm 0mm,clip]{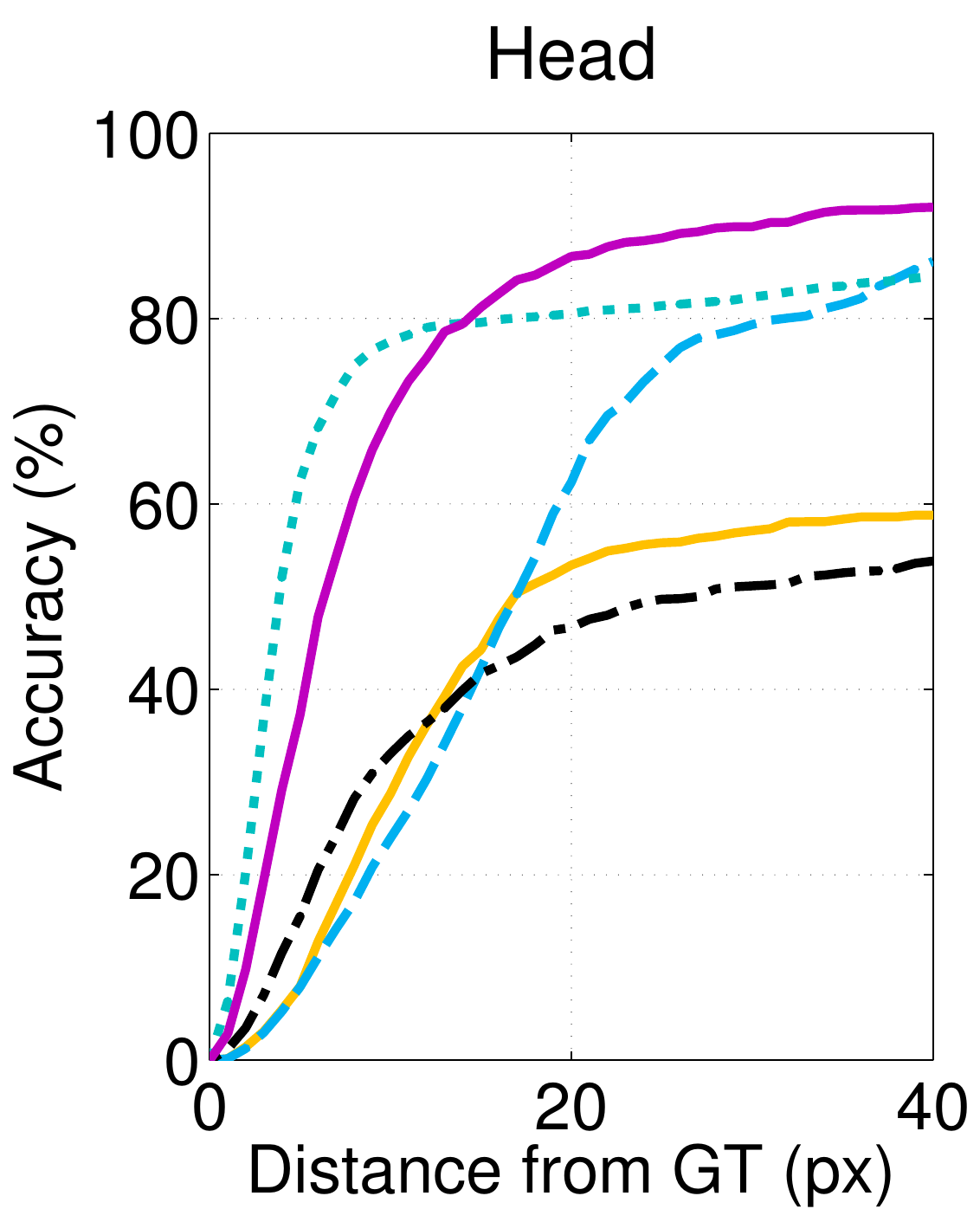} &
\includegraphics[scale=\imgscale, trim = 10mm 0mm 0mm 0mm,clip]{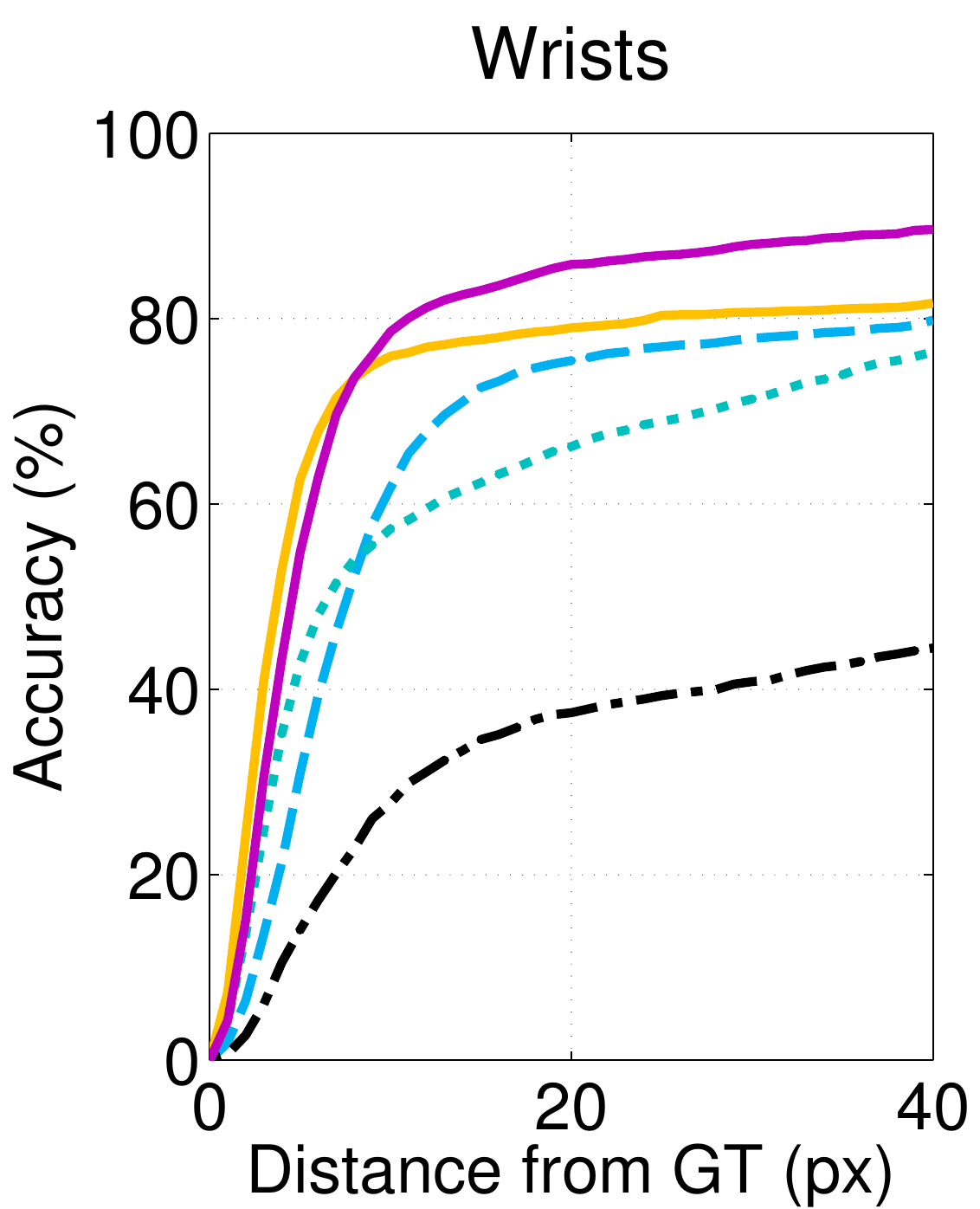} &
\includegraphics[scale=\imgscale, trim = 10mm 0mm 0mm 0mm,clip]{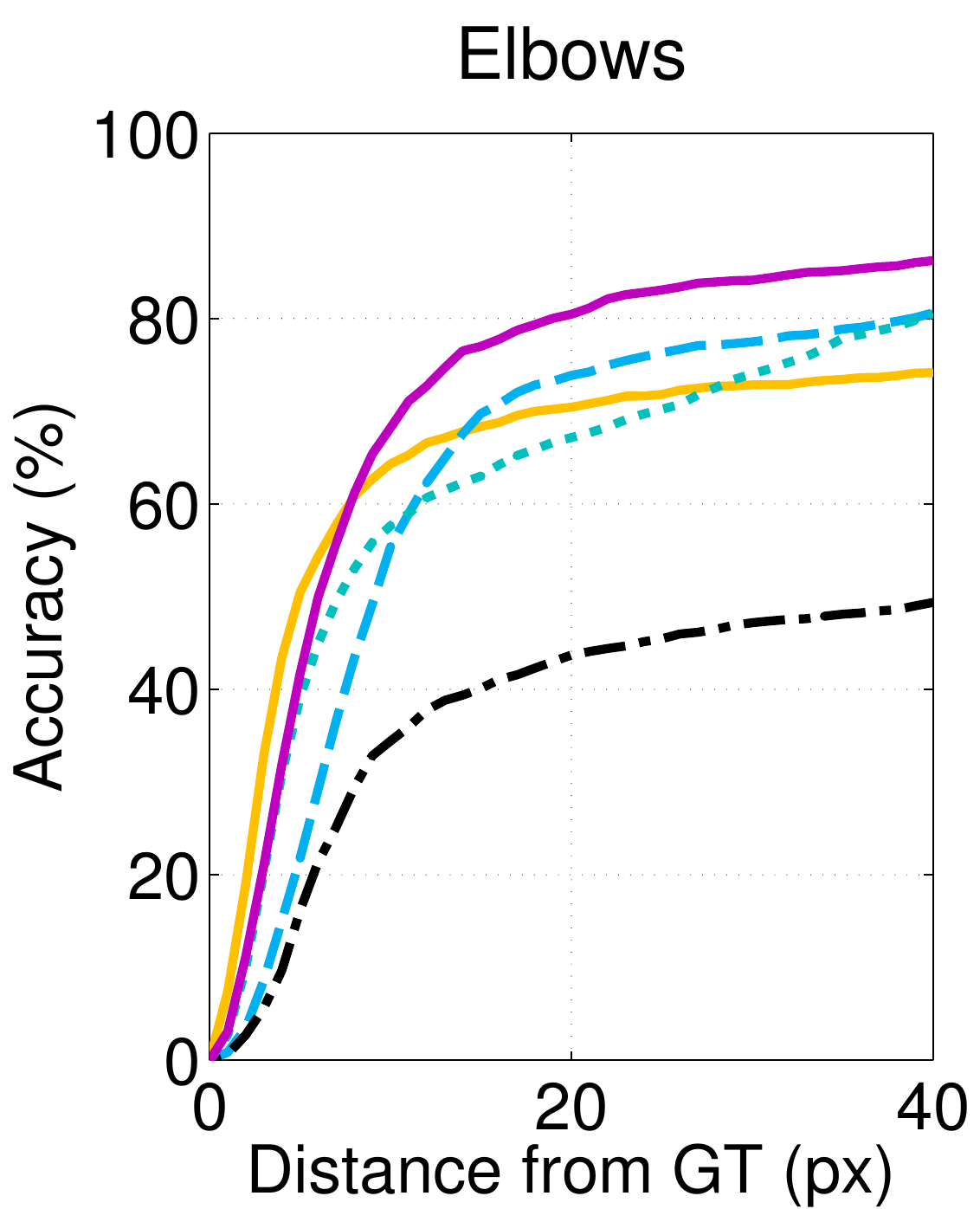} &
\includegraphics[scale=\imgscale, trim = 10mm 0mm 0mm 0mm,clip]{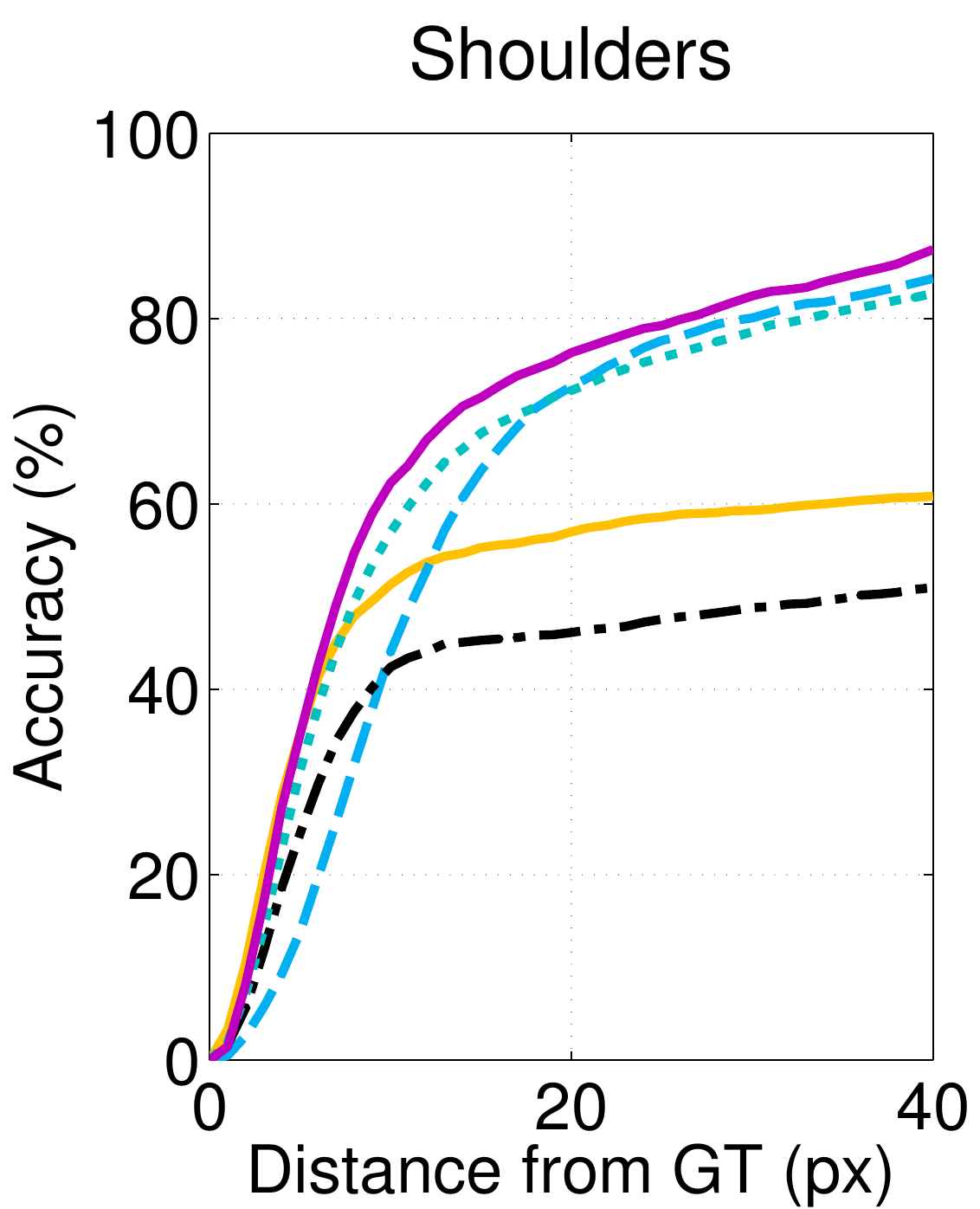}  
\end{tabular}
\end{center}
\vspace{-4mm}
\caption{{\bf Comparison to the state of the art.}
Accuracy of pose estimation evaluated on three datasets. 
Accuracy is averaged over left and right body parts and shown as allowed distance from manual ground truth as $d$ is increased.}
\label{fig:resultsyoutube}
\vspace{1mm}
\vspace{\spcfigcaption}

\end{figure*}

\vspace{-1mm}
\subsection{Boosting a generic ConvNet for other videos}
\vspace{-1mm}
The previous section showed the boost in performance through
personalization on a target video.  There remains the question of
whether there is an additional benefit for generic pose estimation: if
the ConvNet is fine-tuned on personalized annotations over many
videos, does this improve pose estimation performance when applied to
other videos and datasets?

This question is answered by attempting to boost performance of the
generic ConvNet model by supplementing the training data with
automatically annotated frames from the YouTube dataset (leaving out
the YouTube Pose Subset for testing).  For this we use a model that is
pre-trained on the FLIC training set, and fine-tune using the
full FLIC training set together with an equal number of annotated
frames sampled from the YouTube videos.

The performance of the pre-trained and fine-tuned models is compared
on the FLIC, YouTube and MPII Cooking test frames. There is a performance boost in
all cases: for FLIC an increase in wrist \& elbow accuracy of 4\% \& 6\%
respectively (at 0.1 normalized distance); for YouTube an increase
in wrist \& elbow accuracy of 8\% \& 5\% respectively (at $d=20$); and for MPII Cooking an increase of 5\% \& 8\% respectively (at $d=20$),
demonstrating the benefit of using additional automatically annotated
training material.

\vspace{-1mm}
\section{Summary and extensions}
\vspace{-1mm}
We have proposed a semi-supervised-like method for personalizing video
pose estimation and have shown that this significantly improves
performance compared to a non-personalized, `generic', pose estimator,
and beats the state of the art by a large margin on four challenging
video pose estimation datasets.  

The method can be used to boost the pose estimation performance on any
long video sequence containing the same person, and we have also shown
that the annotations generated by this personalization can be used to
improve the performance of a ConvNet estimator for other videos.

It is straightforward to extend the method to estimate full body pose
by adding the extra joints and limbs to the initialisation and the
puppet model.  The method can also be extended to deal with multiple
people in a video and occlusion, given a suitable ConvNet model.  For
example, using the generated occlusion-aware annotations to train an
occlusion-aware ConvNet pose estimator.  One alternative formulation
is to train from additional synthetic data as in~\cite{Park15}, with
the data generated using the puppet model. Given the recent success
with training ConvNets on synthetic data, this would certainly be
worth investigating.

\vspace{-5mm}
\paragraph{Acknowledgments.} 
Financial support was provided by the EPSRC grants EP/I012001/1 and EP/I01229X/1.

{\small
\bibliographystyle{ieee}
\bibliography{./bib/shortstrings,mybib,mybib2,./bib/vgg_other,./bib/vgg_local}
}

\newpage
\begin{figure*}[th!]
\begin{minipage}{1.05\textwidth}
\begin{center}
\section*{\LARGE APPENDIX}
\vspace{0.5cm}
\section*{Sub-component evaluations}
Graphs showing personalization sub-component evaluations.\end{center}
\end{minipage}
\end{figure*}

\def \imgscale {0.3}
\begin{figure*}[th!]
\begin{minipage}{1.05\textwidth}
\begin{center}
\begin{tabular}{c*{5}{@{\hspace{2pt}}c}}
\hspace{-0.26cm}\includegraphics[scale=\imgscale]{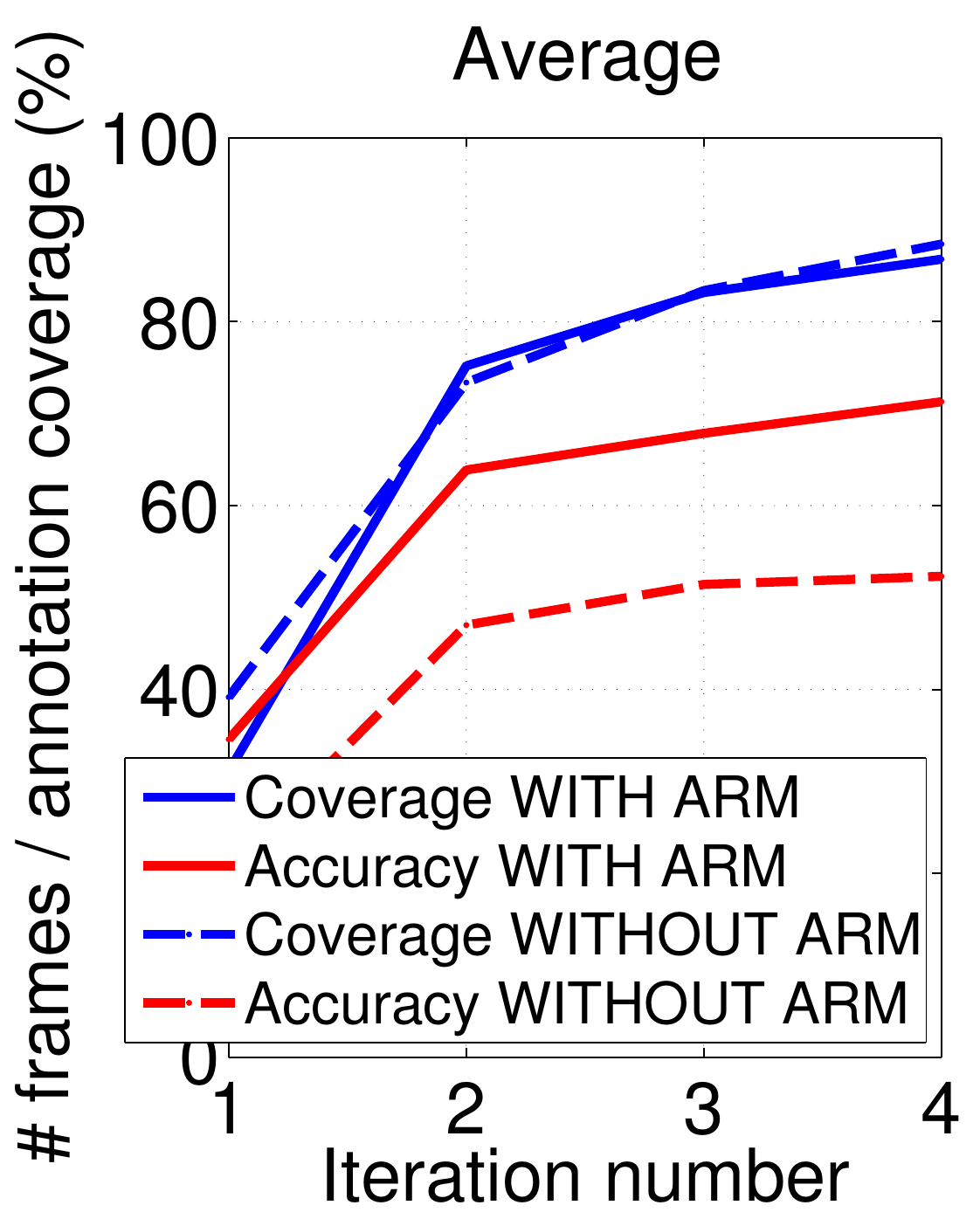}&
\includegraphics[scale=\imgscale]{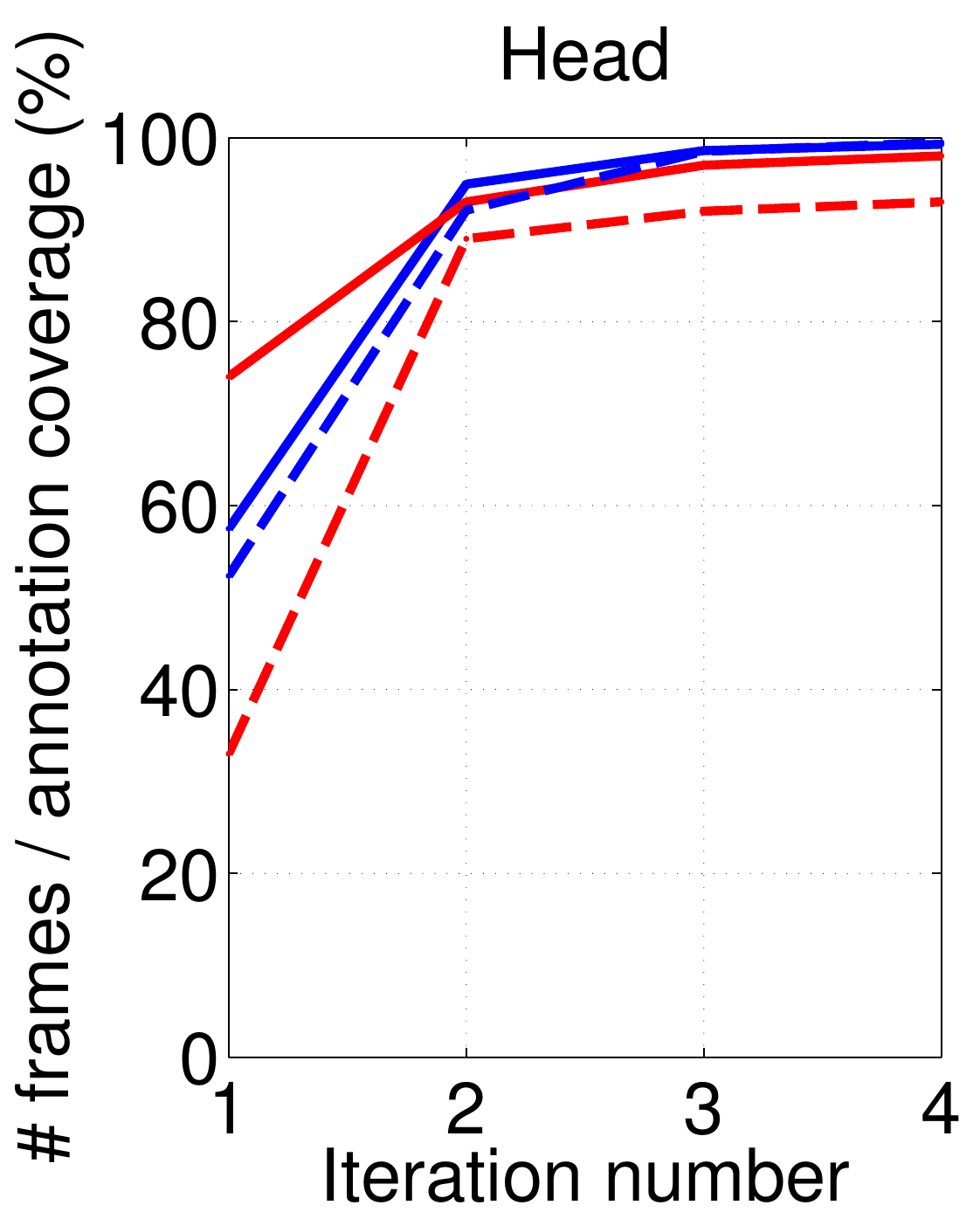} &
\includegraphics[scale=\imgscale]{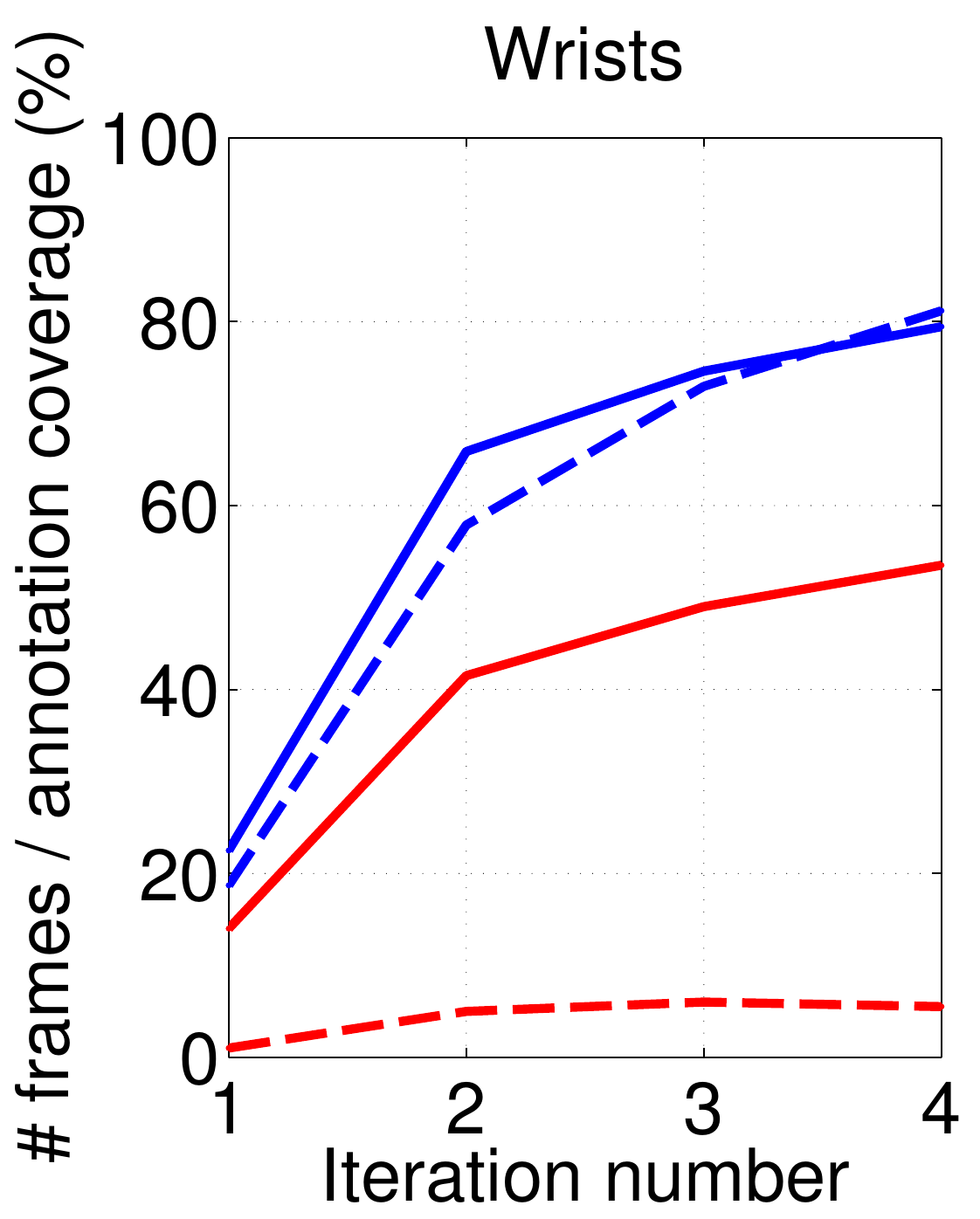} &
\includegraphics[scale=\imgscale]{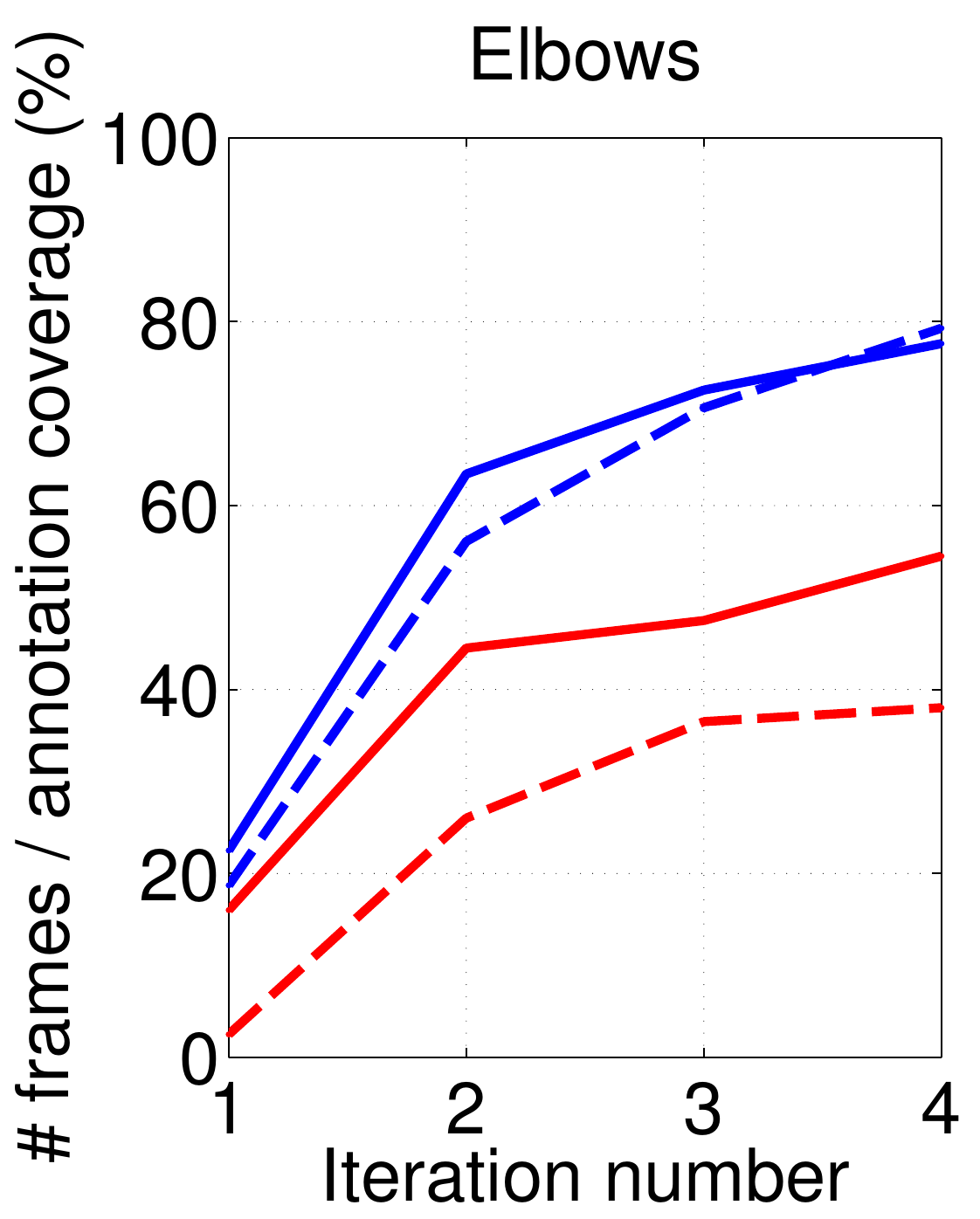}&
\includegraphics[scale=\imgscale]{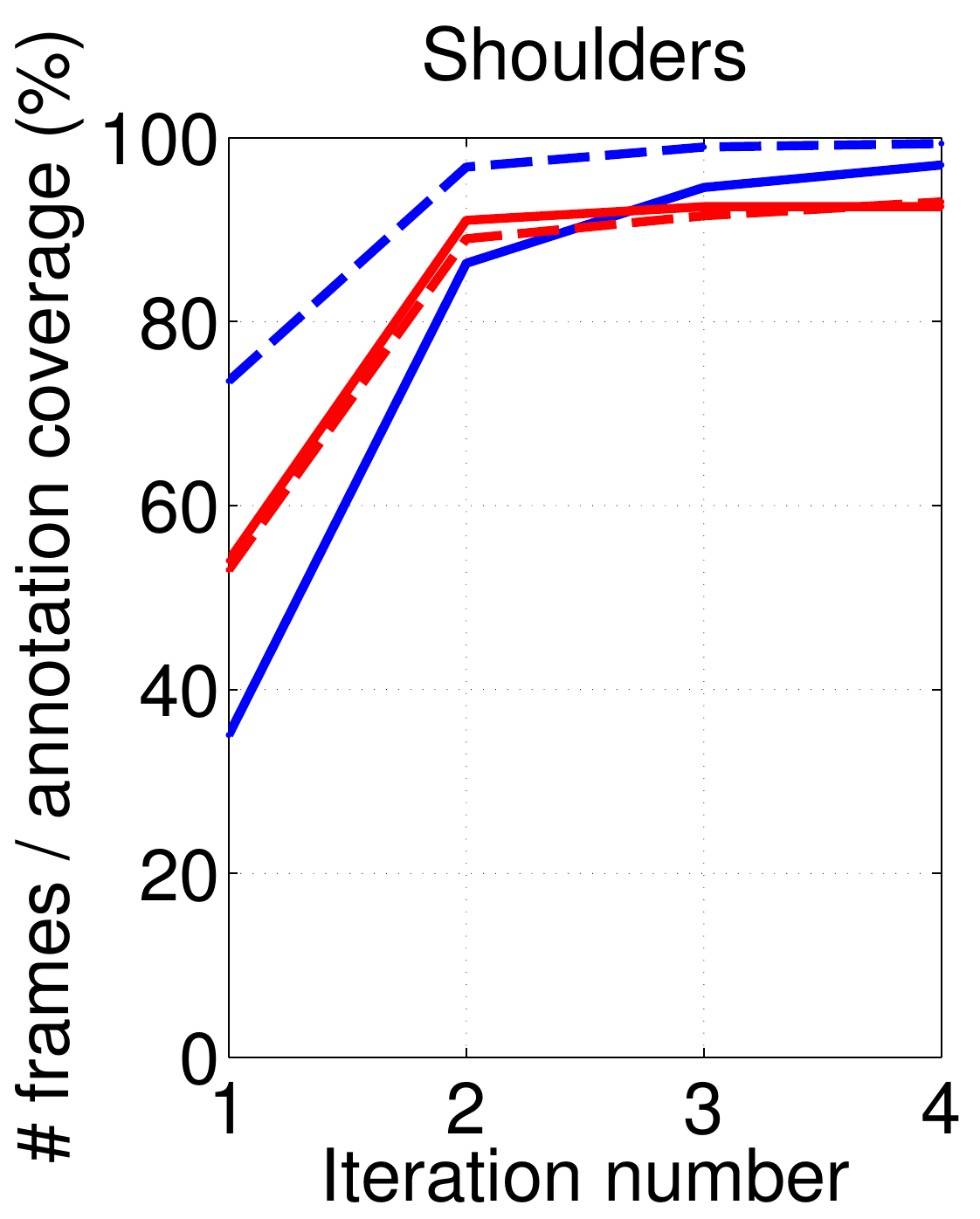}

\end{tabular}
\end{center}
\vspace{-0.2cm}
\caption{{\bf Coverage and accuracy under different initialization methods on YouTube Pose Subset.}
The graphs show the improvement from each stage of our algorithm under two different initialization methods. 
The first is initialized using the ConvNet~[32] and the separate arm detectors (WITH ARM), the second uses only the ConvNet to initialize (WITHOUT ARM).
Accuracy curves are produced by training a random forest body part detector (as described in the main paper) from current annotations, and evaluating it on all ground truth frames from YouTube Pose Subset.}
\label{fig:compeval}\end{minipage}
\vspace{-0.5cm}
\end{figure*}

\def \imgscale {0.3}
\begin{figure*}[th!]
\begin{minipage}{1.05\textwidth}
\begin{center}
\begin{tabular}{c*{5}{@{\hspace{2pt}}c}}
\hspace{-0.26cm}\includegraphics[scale=\imgscale]{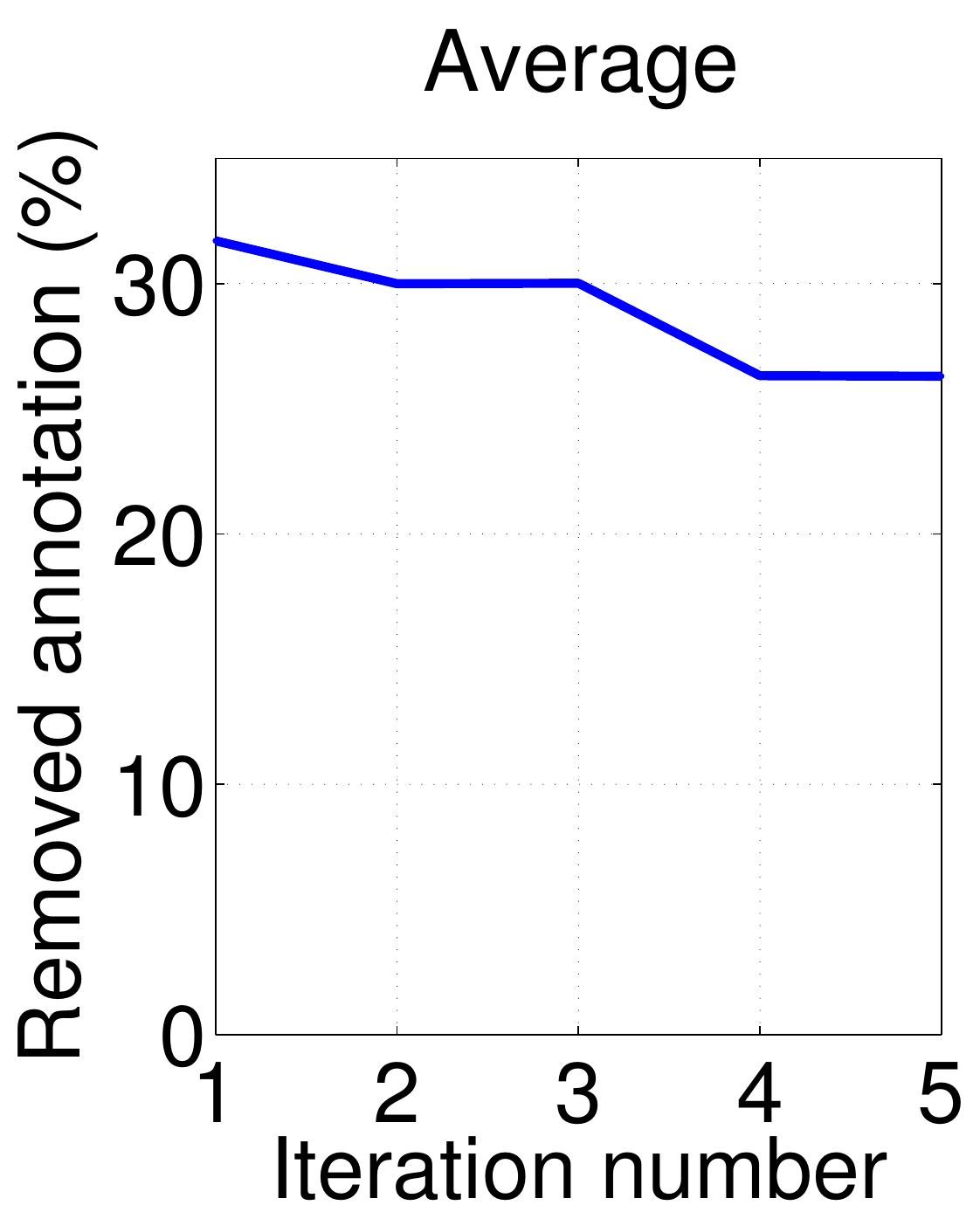}&
\includegraphics[scale=\imgscale]{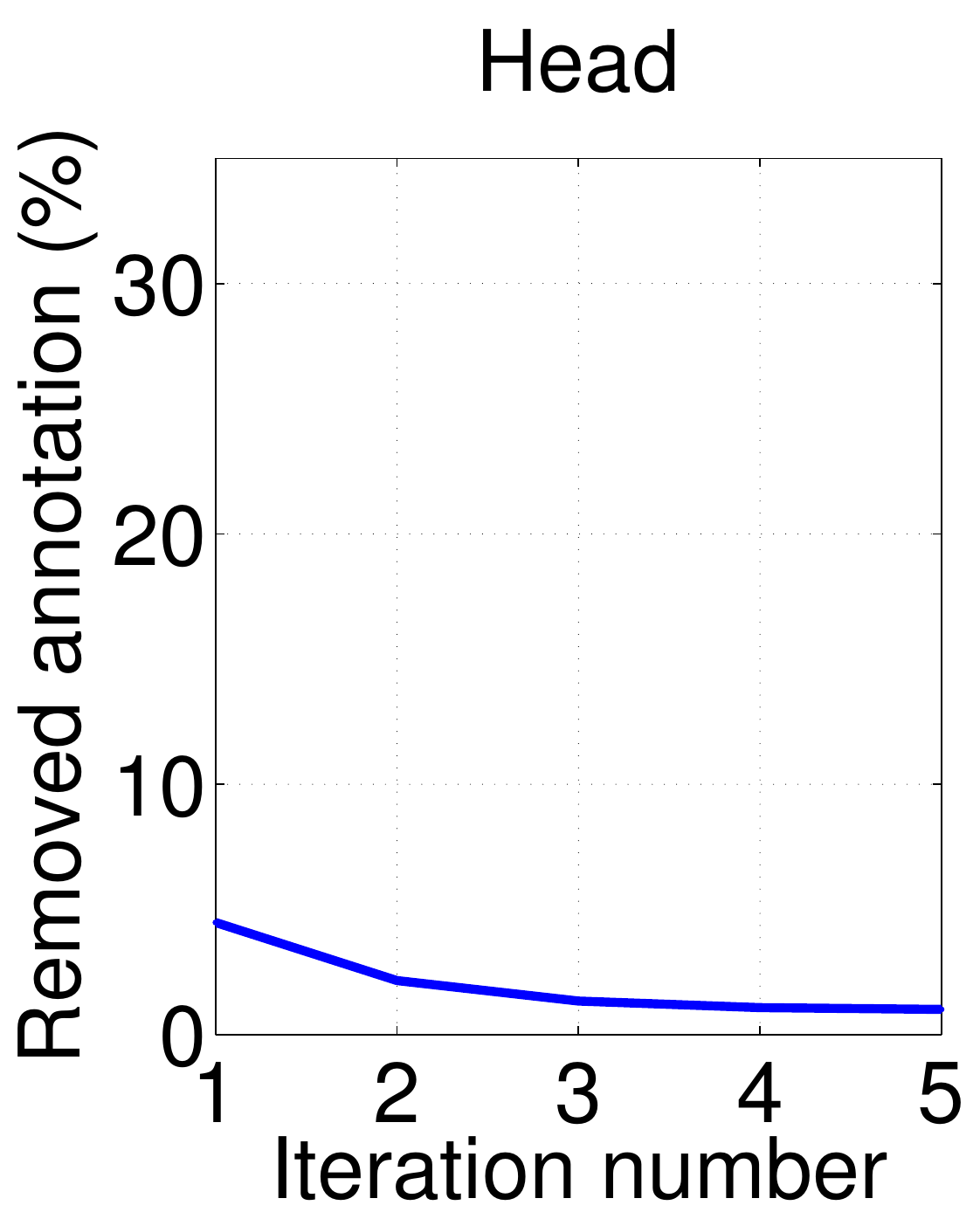} &
\includegraphics[scale=\imgscale]{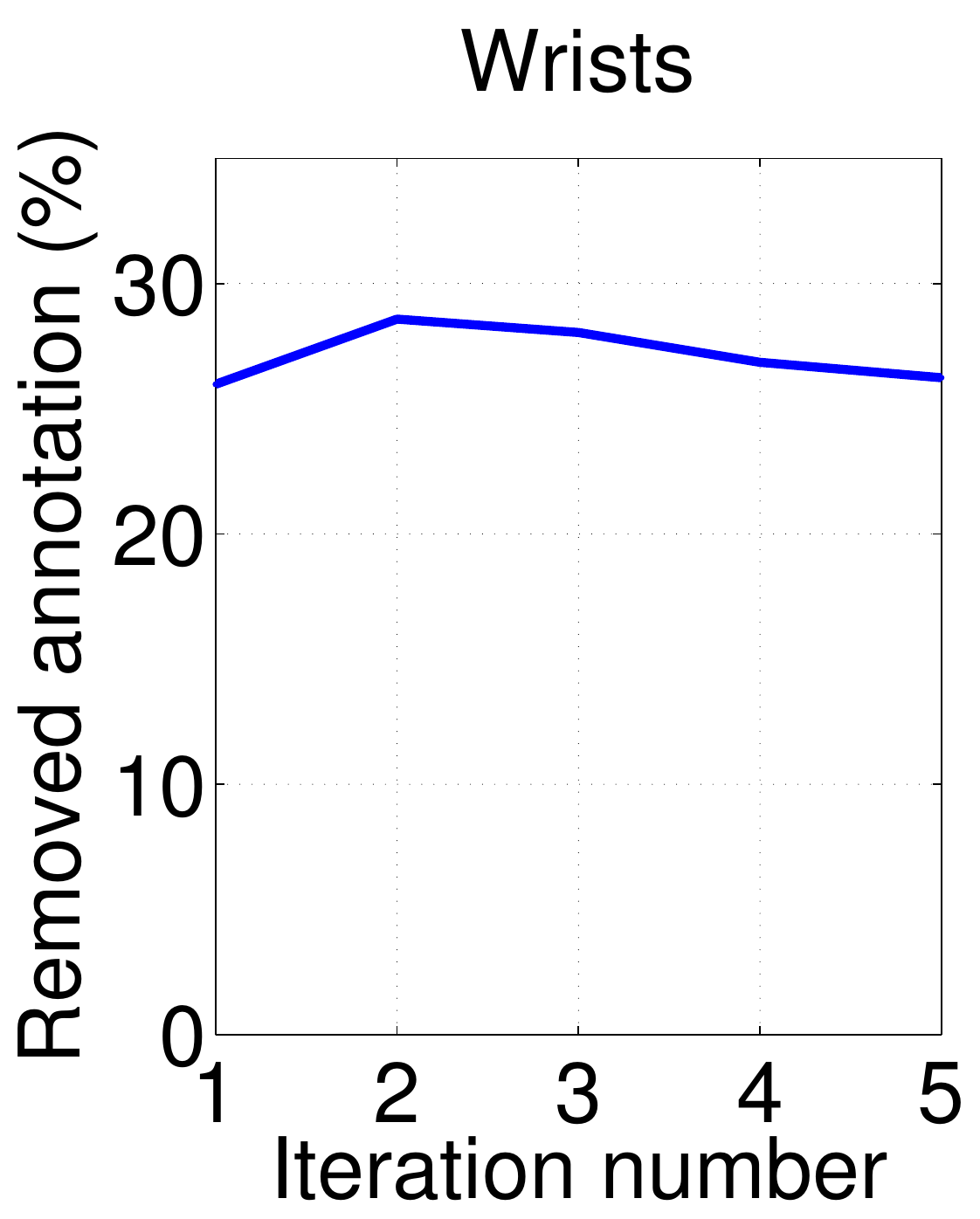} &
\includegraphics[scale=\imgscale]{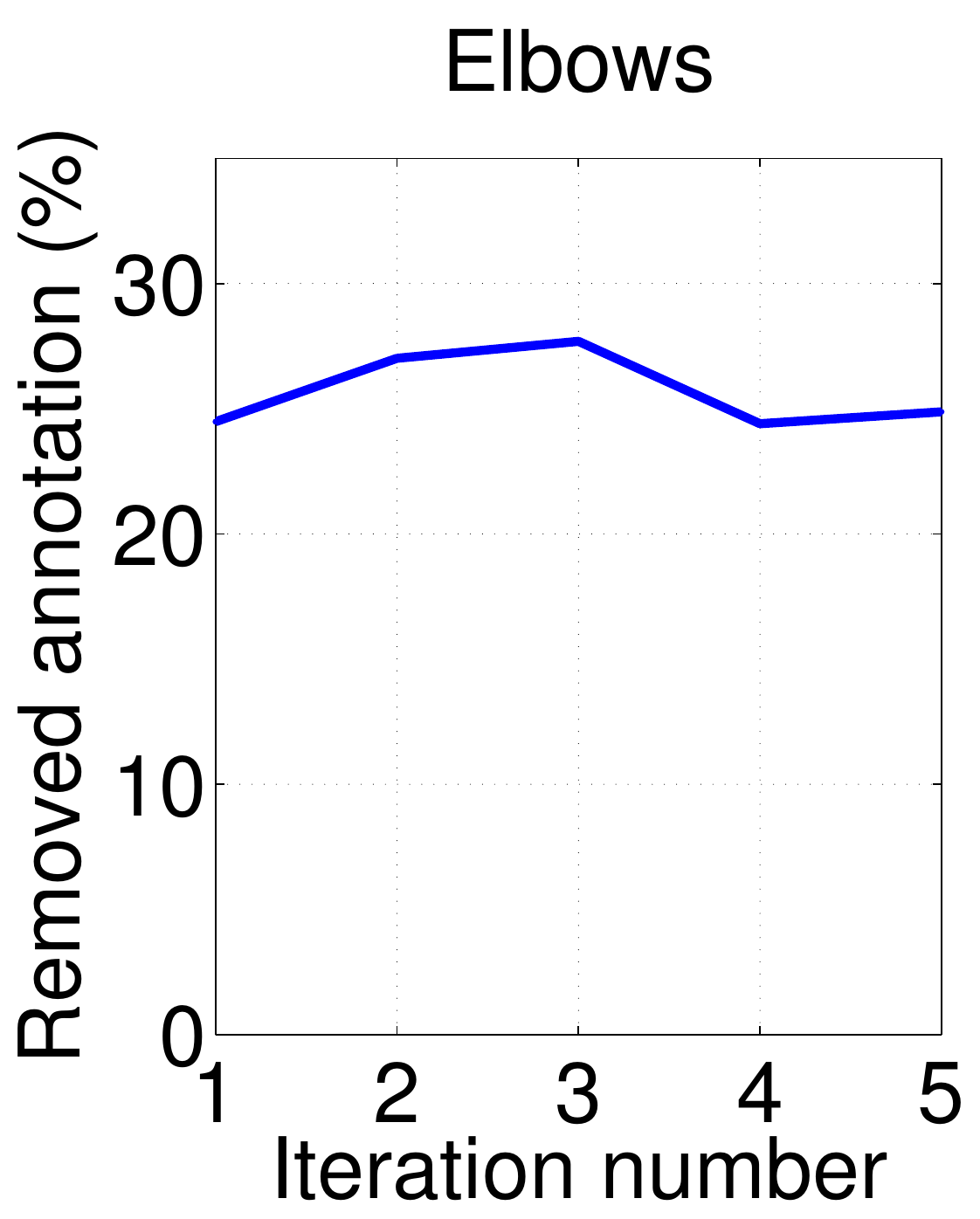}&
\includegraphics[scale=\imgscale]{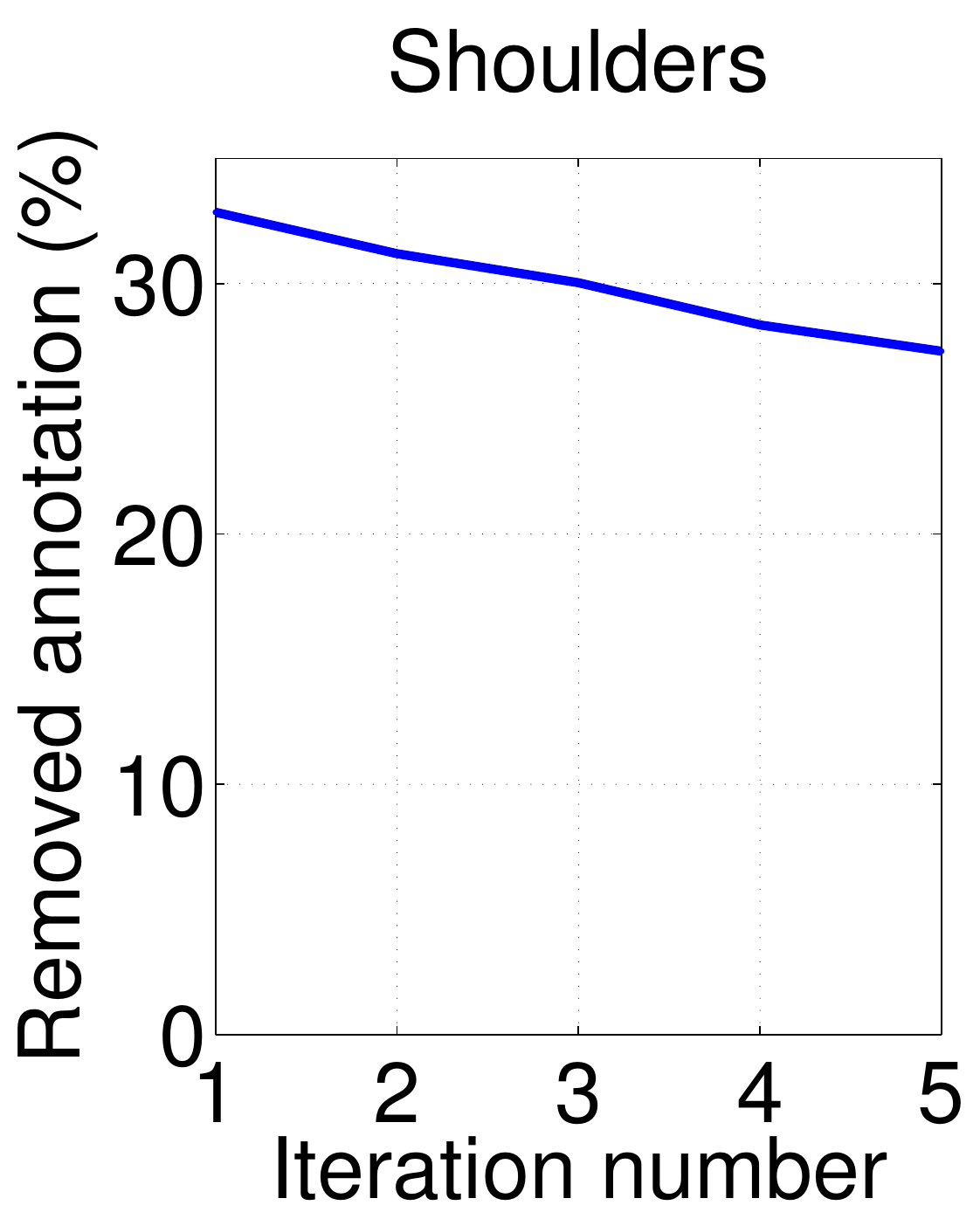}

\end{tabular}
\end{center}
\vspace{-0.2cm}
\caption{{\bf Puppet evaluator response on YouTube Pose Subset.}
For each iteration of our system, the number of per body joint annotations removed by our puppet evaluator are counted. 
These are expressed as a percentage of the total per body joint annotations prior to applying the puppet evaluator, but after removing some annotations with our annotation agreement measure. 
The puppet evaluator is shown to remove additional annotations which pass the agreement measure. 
The graphs demonstrate, on average, a reduction in removed annotation as our system iterates.}
\label{fig:compeval}
\end{minipage}
\end{figure*}

\begin{figure*}
\begin{minipage}{1.05\textwidth}
\begin{center}
\section*{Boosting a generic ConvNet on FLIC}
\end{center}
\end{minipage}
\end{figure*}

\def \imgscale {0.3}
\begin{figure*}[th!]
\begin{minipage}{1.05\textwidth}
\begin{center}
\begin{tabular}{c*{5}{@{\hspace{2pt}}c}}
\hspace{-0.26cm}\includegraphics[scale=\imgscale]{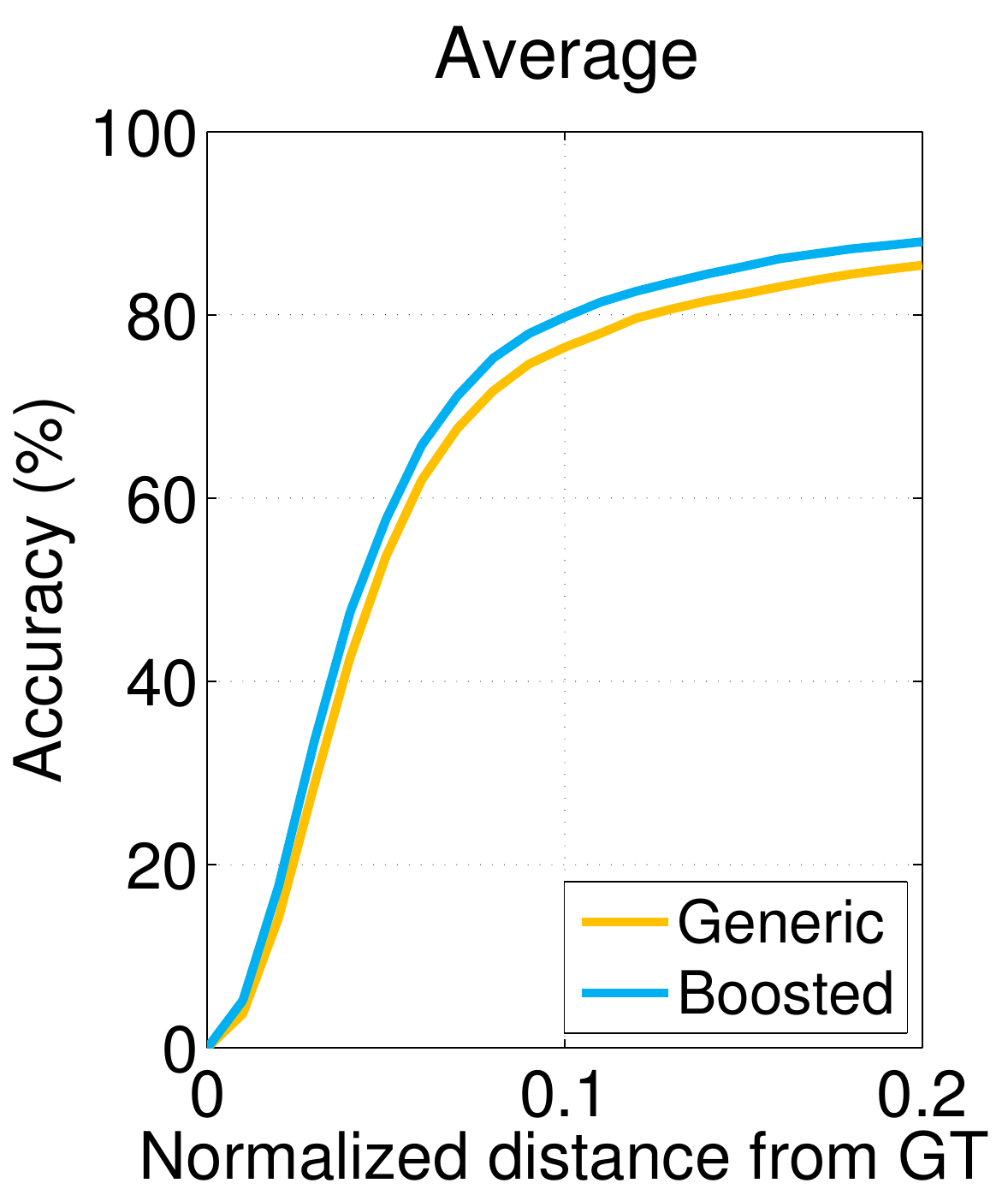}&
\includegraphics[scale=\imgscale]{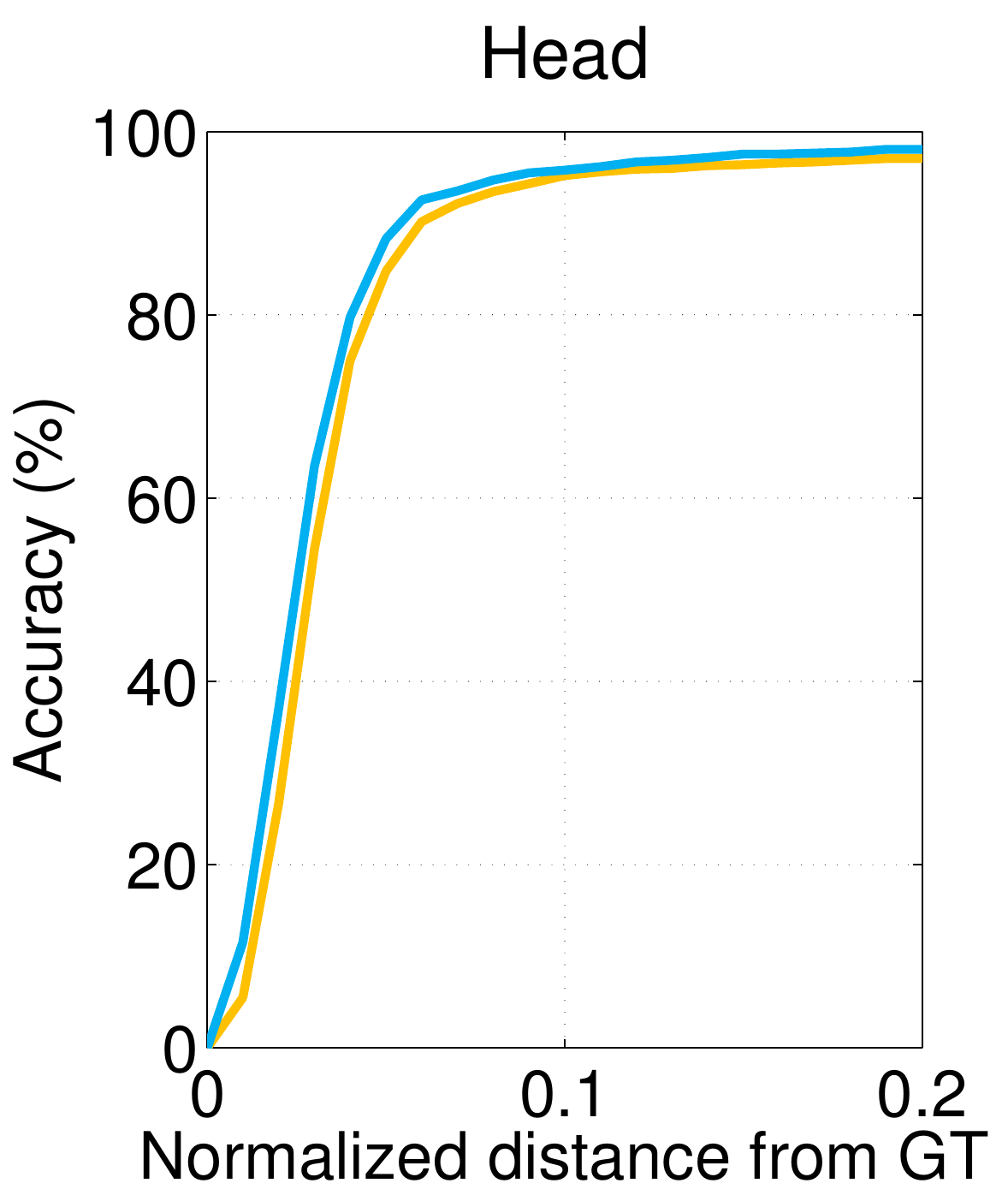} &
\includegraphics[scale=\imgscale]{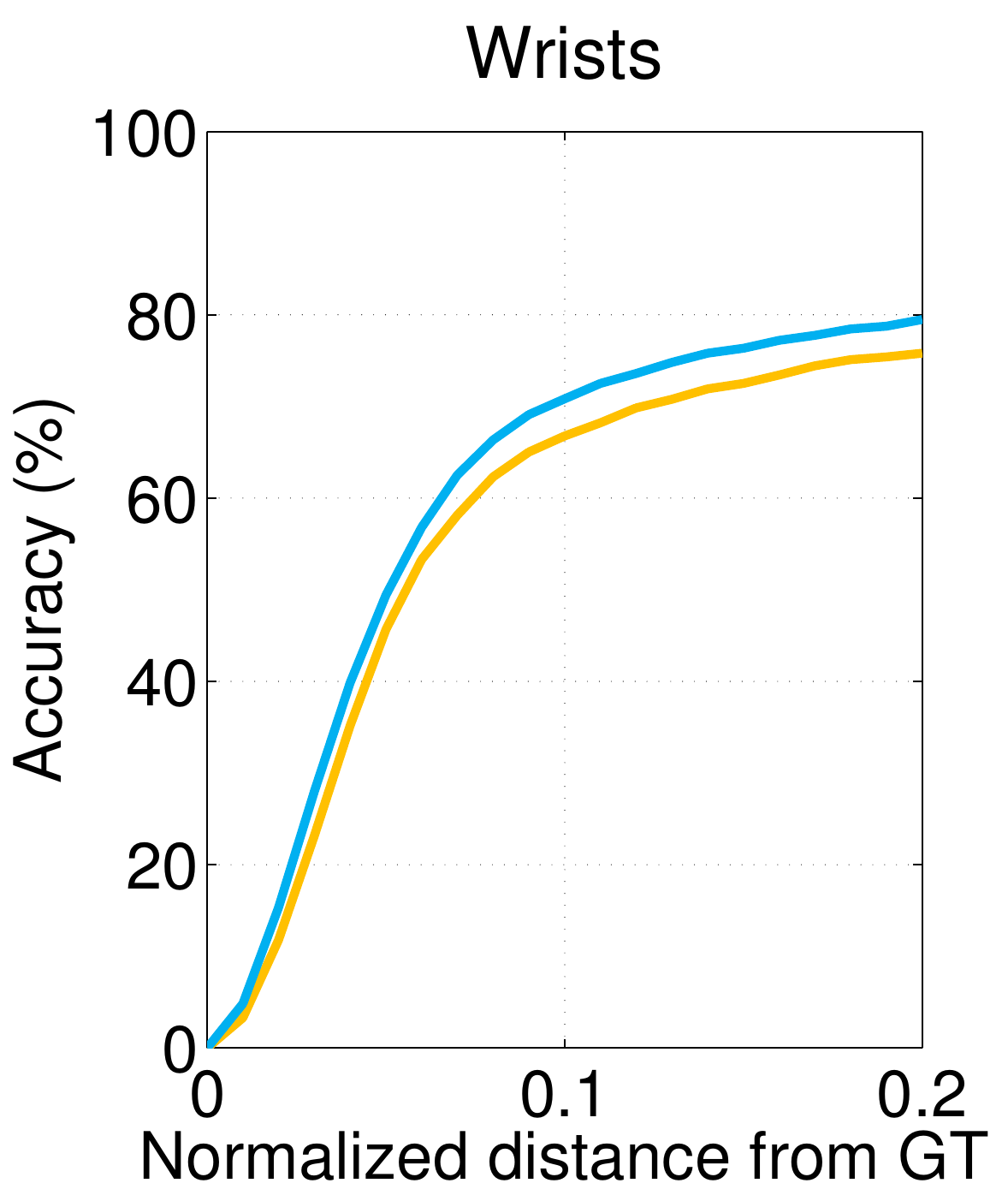} &
\includegraphics[scale=\imgscale]{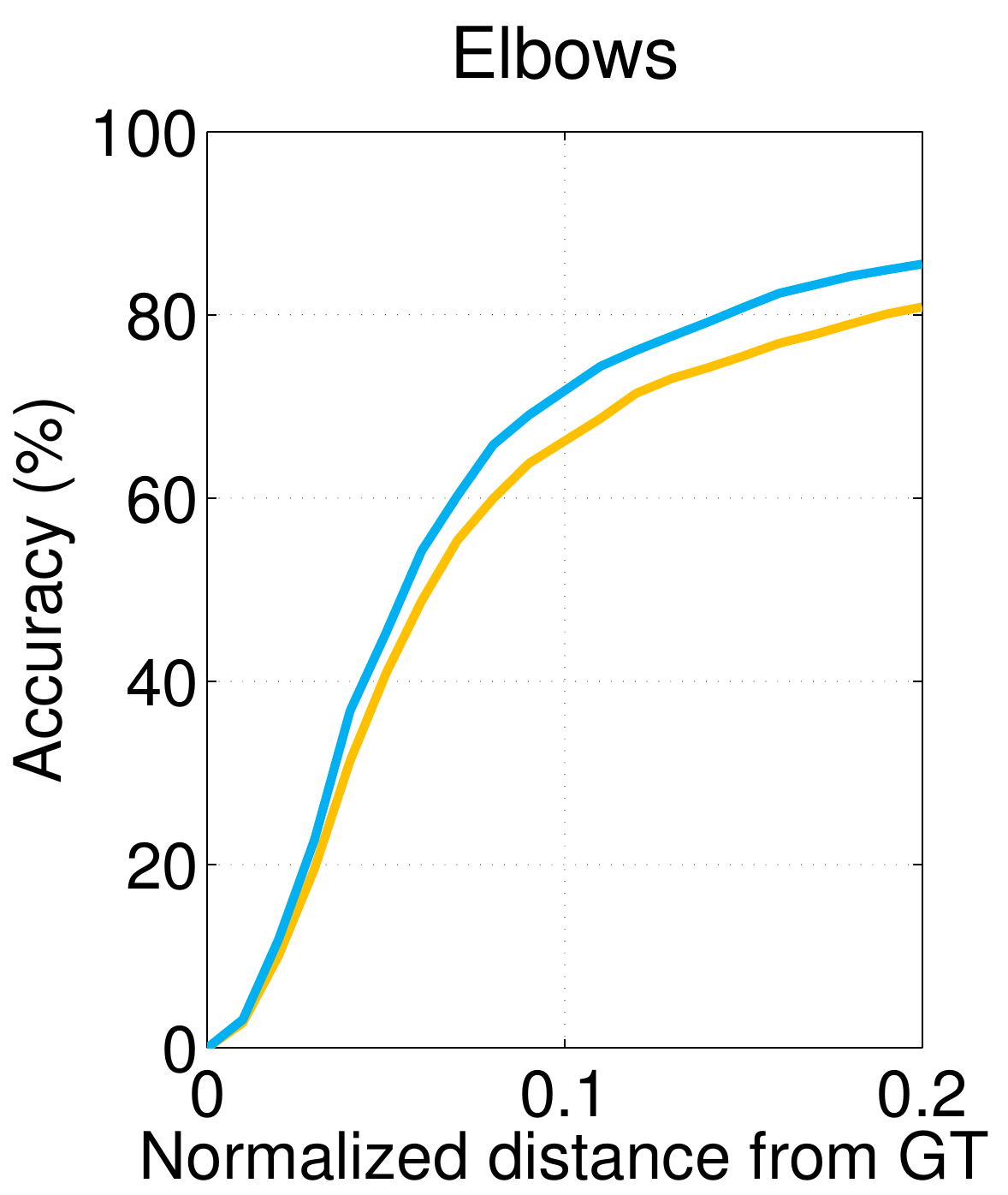}&
\includegraphics[scale=\imgscale]{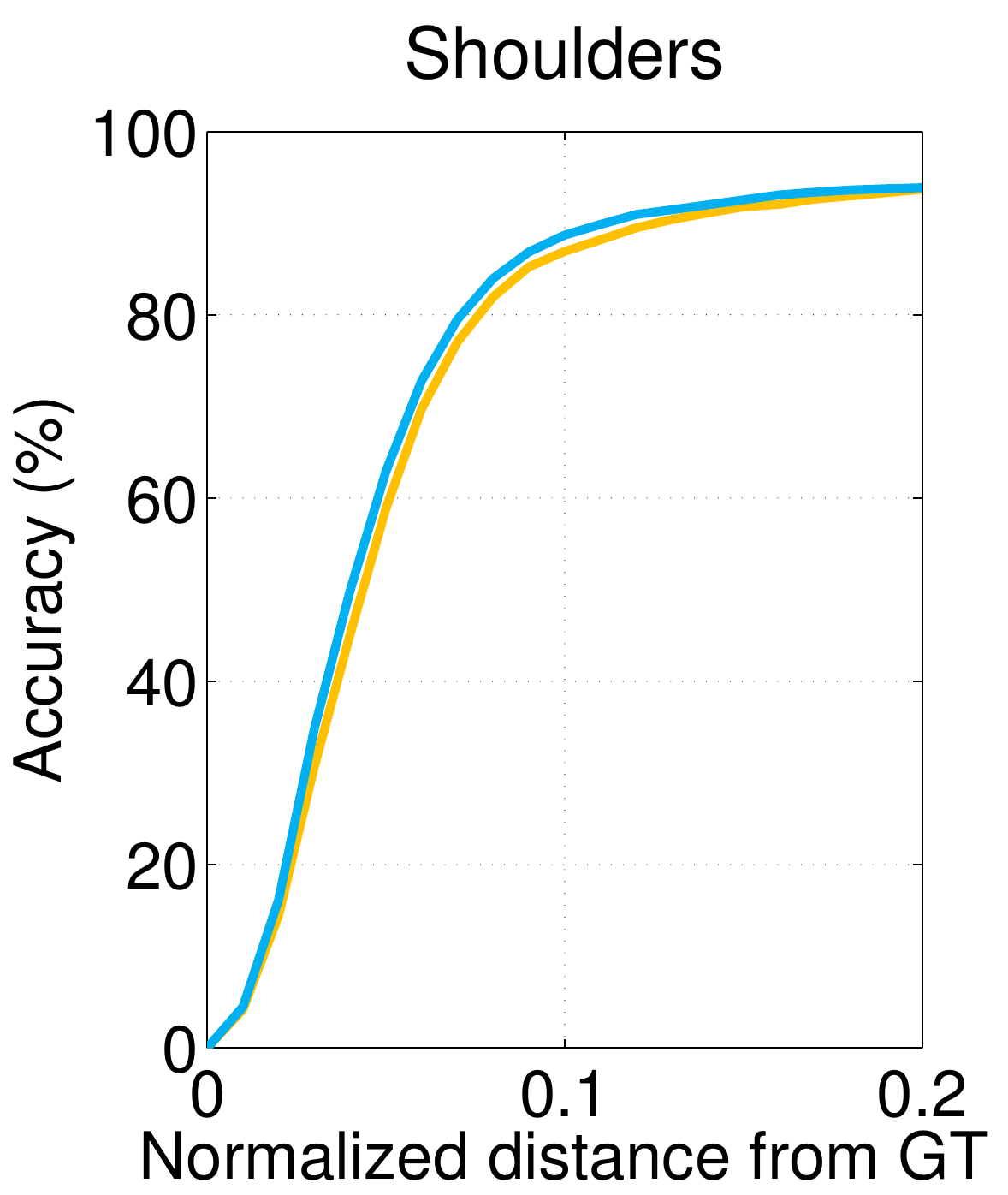}
\end{tabular}
\end{center}
\vspace{-0.2cm}
\caption{{\bf Improvement when training with automatically annotated videos.} 
Performance of a generic ConvNet pose estimator~[32] trained on the FLIC training set (Generic) is compared against a boosted version produced by fine-tuning with 20 additional automatically annotated YouTube Pose videos (Boosted). 
Comparison is performed on the FLIC test set. 
An improvement in performance is observed for all body joints, particularly for the elbows.
}
\label{fig:flicboost}
\end{minipage}
\end{figure*}

\begin{figure*}
\begin{minipage}{1.05\textwidth}
\begin{center}
\section*{Experimental details}
\end{center}
\vspace{\spcpara}\noindent {\bf ConvNet.} 
We use the publicly available ConvNet of Pfister~\etal~\cite{Pfister15} both for initialization and for fine-tuning. 
The available model is pre-trained on the FLIC dataset~\cite{Sapp13} using the 3987 training video frames.
Empirically, for our initialization, we found using body joint estimates with 80\% confidence or above produce very good precision.

\vspace{\spcpara}\noindent {\bf Arm model training.}
The second initialization method is for arm pose estimates using the generic pose estimator of Yang and Ramanan~\cite{Yang11}. 
15 arm pose models are trained on the MPII human pose dataset (by clustering all arm poses into 15 clusters using $k$-means, and retaining the nearest 150 poses to the cluster centroid for training), making it possible to detect up to 225 different poses. 
Note, there is no overlap between the MPII human pose dataset used for training and the MPII cooking dataset~\cite{Rohrbach12} used for testing.
Each model is trained to have high precision (at least 90\% detection accuracy) by setting their confidence threshold so as not to fire on pose clusters that they weren't trained on. 
We use the LSP extended dataset to learn these thresholds.

\vspace{\spcpara}\noindent {\bf Parameters.}
After temporal propagation, annotations are only retained if temporal agreement of overlapping annotation is below 20 pixels {\em and} overlapping annotation stems from at least three different frames. 
All videos are scaled to contain a person with width between the shoulders of approximately 100 pixels.

\vspace{\spcpara}\noindent {\bf Joint offsets.}
There exists consistent body joint offsets between manual ground truth annotations on FLIC and those on BBC Pose or MPII Cooking. 
Therefore, to ensure a fair comparison between all models, pose estimates from those models trained/initialized from FLIC are adjusted by these offsets.

\vspace{\spcpara}\noindent {\bf Personalized ConvNet average accuracy on training annotation.} 
Here we report average training error (using all automatically generated annotations as ground truth) of the personalized ConvNet on BBC Pose (98\%), YouTube Pose Subset (91\%) and MPII Cooking (90\%). Interestingly, after training, we found the personalized ConvNet predictions have higher precision than the generated annotation.
\end{minipage}
\vspace{50cm}
\end{figure*}

\newpage

\begin{figure*}[th!]
\begin{minipage}{1.05\textwidth}

\begin{center}
\section*{YouTube Pose dataset pose tracking output}
Example video frames and pose tracking output for the YouTube Pose dataset.
\end{center}
\end{minipage}
\vspace{-1cm}
\end{figure*}

\def \imgscale {0.52}
\begin{figure*}[h]
\vspace{1cm}
\begin{minipage}{1.05\textwidth}
\hspace{-5mm}\begin{tabular}{c*{6}{@{\hspace{2pt}}c}}
\includegraphics[scale=\imgscale]{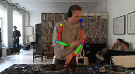} & \includegraphics[scale=\imgscale]{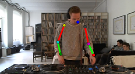} & \includegraphics[scale=\imgscale]{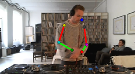} & \includegraphics[scale=\imgscale]{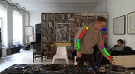} & \includegraphics[scale=\imgscale]{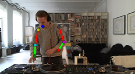} & \includegraphics[scale=\imgscale]{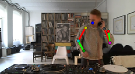} & \includegraphics[scale=\imgscale]{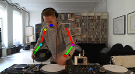} \\
 \includegraphics[scale=\imgscale]{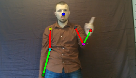} & \includegraphics[scale=\imgscale]{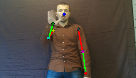} & \includegraphics[scale=\imgscale]{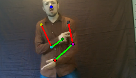} & \includegraphics[scale=\imgscale]{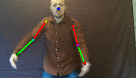} & \includegraphics[scale=\imgscale]{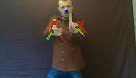} & \includegraphics[scale=\imgscale]{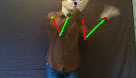} & \includegraphics[scale=\imgscale]{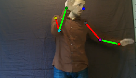} \\
 \includegraphics[scale=\imgscale]{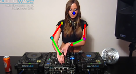} & \includegraphics[scale=\imgscale]{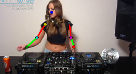} & \includegraphics[scale=\imgscale]{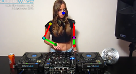} & \includegraphics[scale=\imgscale]{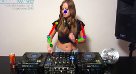} & \includegraphics[scale=\imgscale]{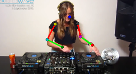} & \includegraphics[scale=\imgscale]{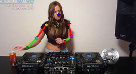} & \includegraphics[scale=\imgscale]{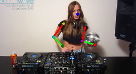} \\
 \includegraphics[scale=\imgscale]{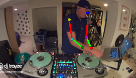} & \includegraphics[scale=\imgscale]{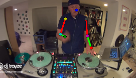} & \includegraphics[scale=\imgscale]{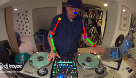} & \includegraphics[scale=\imgscale]{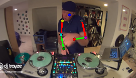} & \includegraphics[scale=\imgscale]{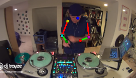} & \includegraphics[scale=\imgscale]{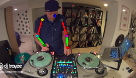} & \includegraphics[scale=\imgscale]{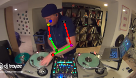} \\
 \includegraphics[scale=\imgscale]{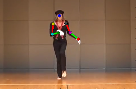} & \includegraphics[scale=\imgscale]{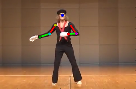} & \includegraphics[scale=\imgscale]{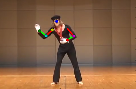} & \includegraphics[scale=\imgscale]{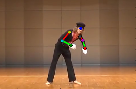} & \includegraphics[scale=\imgscale]{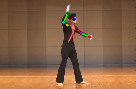} & \includegraphics[scale=\imgscale]{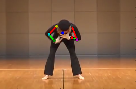} & \includegraphics[scale=\imgscale]{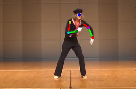} \\
 \includegraphics[scale=\imgscale]{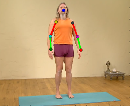} & \includegraphics[scale=\imgscale]{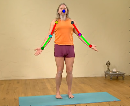} & \includegraphics[scale=\imgscale]{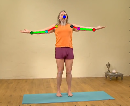} & \includegraphics[scale=\imgscale]{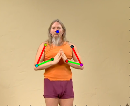} & \includegraphics[scale=\imgscale]{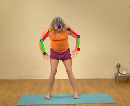} & \includegraphics[scale=\imgscale]{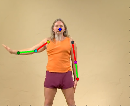} & \includegraphics[scale=\imgscale]{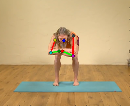} \\
 \includegraphics[scale=\imgscale]{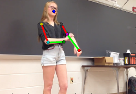} & \includegraphics[scale=\imgscale]{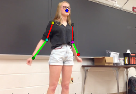} & \includegraphics[scale=\imgscale]{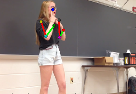} & \includegraphics[scale=\imgscale]{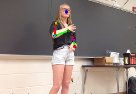} & \includegraphics[scale=\imgscale]{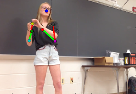} & \includegraphics[scale=\imgscale]{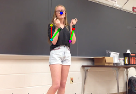} & \includegraphics[scale=\imgscale]{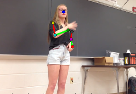} \\
 \includegraphics[scale=\imgscale]{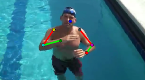} & \includegraphics[scale=\imgscale]{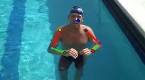} & \includegraphics[scale=\imgscale]{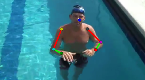} & \includegraphics[scale=\imgscale]{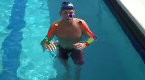} & \includegraphics[scale=\imgscale]{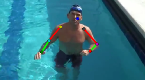} & \includegraphics[scale=\imgscale]{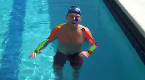} & \includegraphics[scale=\imgscale]{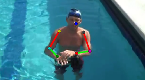} \\
 \includegraphics[scale=\imgscale]{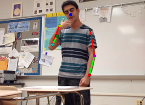} & \includegraphics[scale=\imgscale]{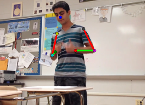} & \includegraphics[scale=\imgscale]{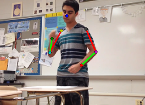} & \includegraphics[scale=\imgscale]{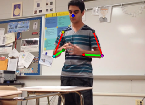} & \includegraphics[scale=\imgscale]{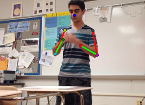} & \includegraphics[scale=\imgscale]{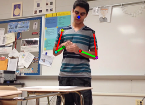} & \includegraphics[scale=\imgscale]{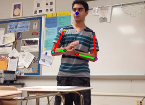} \\
 \includegraphics[scale=\imgscale]{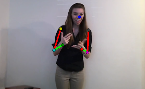} & \includegraphics[scale=\imgscale]{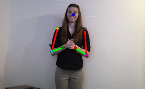} & \includegraphics[scale=\imgscale]{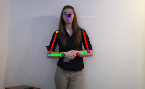} & \includegraphics[scale=\imgscale]{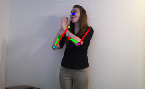} & \includegraphics[scale=\imgscale]{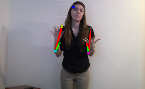} & \includegraphics[scale=\imgscale]{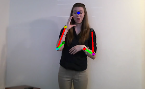} & \includegraphics[scale=\imgscale]{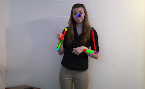} \\
 \includegraphics[scale=\imgscale]{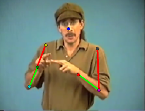} & \includegraphics[scale=\imgscale]{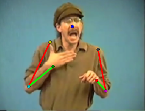} & \includegraphics[scale=\imgscale]{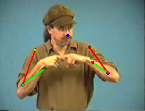} & \includegraphics[scale=\imgscale]{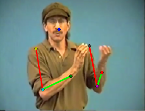} & \includegraphics[scale=\imgscale]{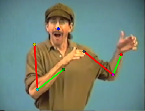} & \includegraphics[scale=\imgscale]{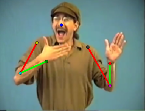} & \includegraphics[scale=\imgscale]{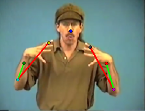} \\
 \includegraphics[scale=\imgscale]{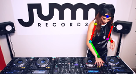} & \includegraphics[scale=\imgscale]{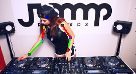} & \includegraphics[scale=\imgscale]{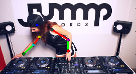} & \includegraphics[scale=\imgscale]{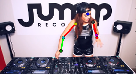} & \includegraphics[scale=\imgscale]{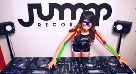} & \includegraphics[scale=\imgscale]{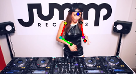} & \includegraphics[scale=\imgscale]{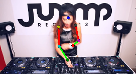}

\end{tabular}
\caption{{\bf YouTube Pose dataset and pose estimates.}
Example frames from videos in the YouTube Pose dataset are shown in each row along with pose estimates (as stick figures) from the personalized ConvNet. 
Note the variety of poses, clothing, backgrounds and camera angles.}
\end{minipage}
\vspace{-1cm}
\end{figure*}
\newpage

\begin{figure*}
\begin{minipage}{1.05\textwidth}
\hspace{-5mm}\begin{tabular}{c*{6}{@{\hspace{2pt}}c}}
\includegraphics[scale=\imgscale]{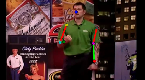} & \includegraphics[scale=\imgscale]{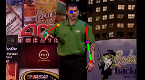} & \includegraphics[scale=\imgscale]{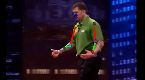} & \includegraphics[scale=\imgscale]{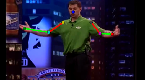} & \includegraphics[scale=\imgscale]{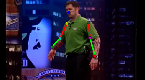} & \includegraphics[scale=\imgscale]{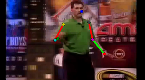} & \includegraphics[scale=\imgscale]{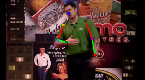} \\
 \includegraphics[scale=\imgscale]{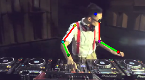} & \includegraphics[scale=\imgscale]{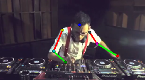} & \includegraphics[scale=\imgscale]{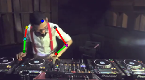} & \includegraphics[scale=\imgscale]{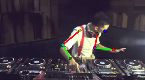} & \includegraphics[scale=\imgscale]{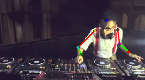} & \includegraphics[scale=\imgscale]{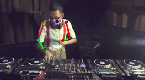} & \includegraphics[scale=\imgscale]{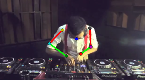} \\
 \includegraphics[scale=\imgscale]{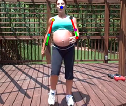} & \includegraphics[scale=\imgscale]{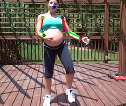} & \includegraphics[scale=\imgscale]{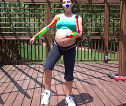} & \includegraphics[scale=\imgscale]{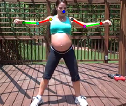} & \includegraphics[scale=\imgscale]{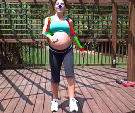} & \includegraphics[scale=\imgscale]{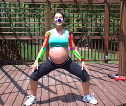} & \includegraphics[scale=\imgscale]{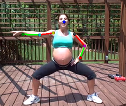}

\end{tabular}
\caption{{\bf More YouTube Pose dataset and pose estimates.}
Example frames from videos in the YouTube Pose dataset are shown in each row along with pose estimates (as stick figures) from the personalized ConvNet.}
\end{minipage}
\end{figure*}

\def \imgscale {0.6}

\begin{figure*}
\begin{minipage}{1.05\textwidth}
\begin{centering}

 \includegraphics[scale=\imgscale]{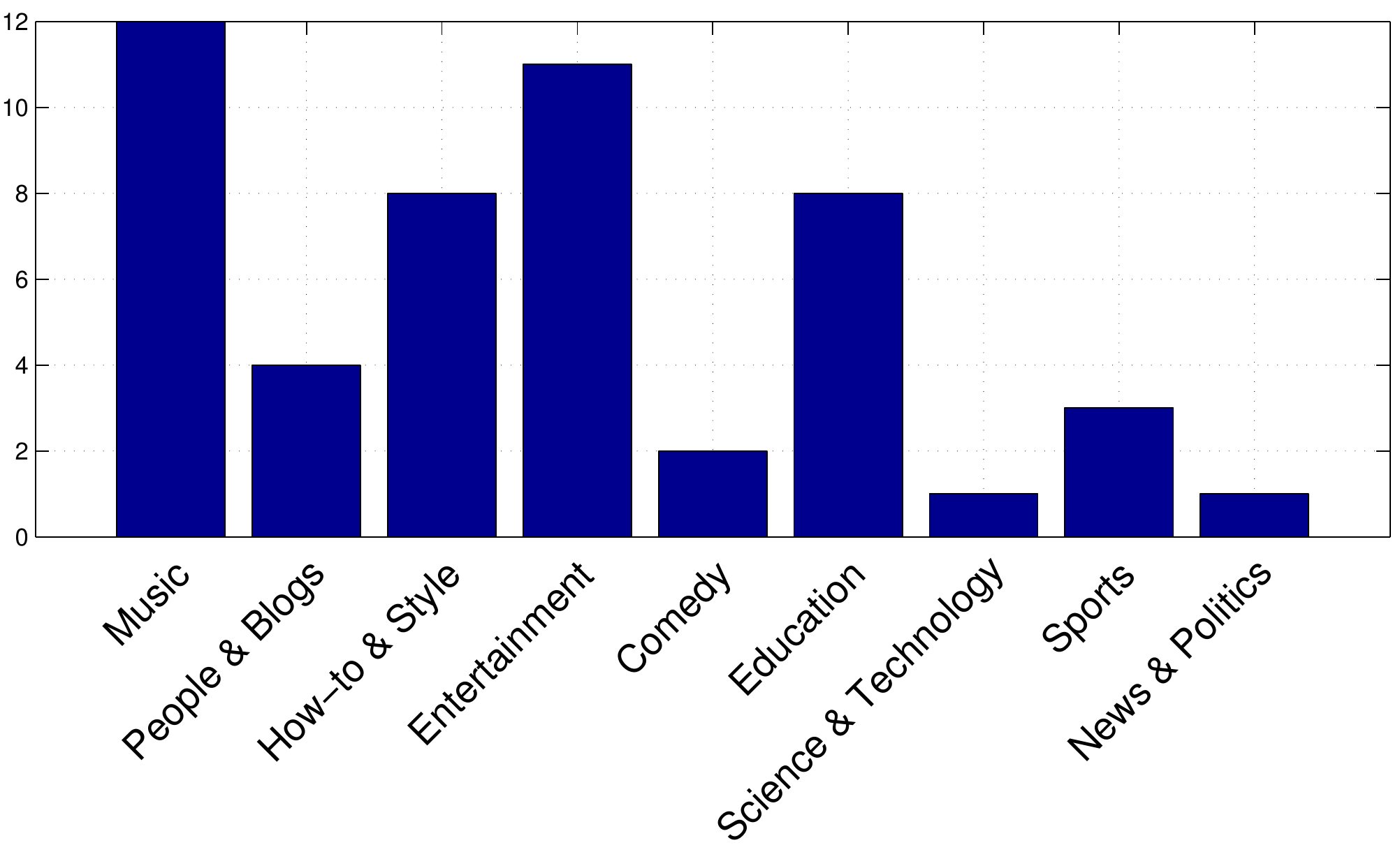}

 \caption{{\bf YouTube Pose category distribution.} 
 Distribution of video categories in the 50 video YouTube Pose dataset.}
\end{centering}
\end{minipage}
\end{figure*}

\def \imgscale {0.4}

\begin{figure*}
\begin{minipage}{1.05\textwidth}
\begin{centering}
\begin{tabular}{c*{3}{@{\hspace{2pt}}c}}
 \includegraphics[scale=\imgscale]{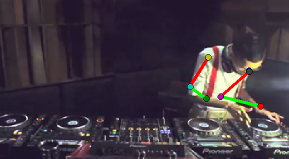} & \includegraphics[scale=\imgscale]{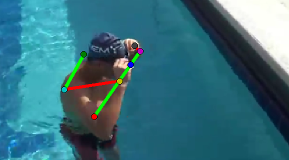} & \includegraphics[scale=\imgscale]{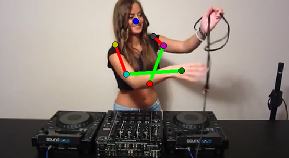} & \includegraphics[scale=\imgscale]{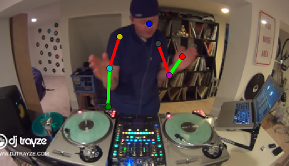} \\
 \includegraphics[scale=\imgscale]{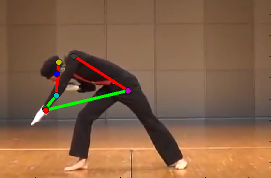} & \includegraphics[scale=\imgscale]{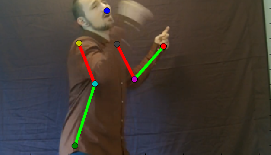} & \includegraphics[scale=\imgscale]{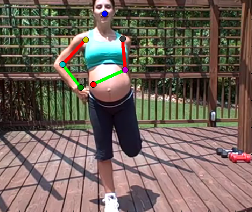} & \includegraphics[scale=\imgscale]{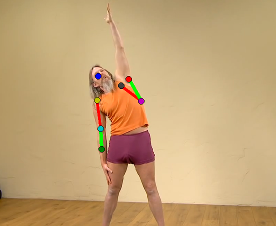}
\end{tabular}
\end{centering}
\caption{{\bf Failure cases.} 
Example frames with erroneous pose estimates from personalized ConvNets. 
There are two main causes of failure: 
(i)~heavy occlusion (including self-occlusion), and 
(ii)~poses which our automated annotation system could not propagate, due to either optical flow error or very few initial annotations. }
\end{minipage}
\end{figure*}

\end{document}